\documentclass[11pt]{article}

\usepackage[utf8]{inputenc}
\usepackage[T1]{fontenc}
\usepackage{amsmath}
\usepackage{amssymb}
\usepackage{newtxtext,newtxmath}      
\usepackage{microtype}                
\usepackage[margin=1.05in]{geometry}
\usepackage{graphicx}
\usepackage{booktabs}
\usepackage{multirow}
\usepackage{tabularx}
\usepackage{xcolor}
\usepackage{float}
\usepackage{enumitem}
\usepackage[section]{placeins}   

\usepackage[font=small,labelfont=bf,margin=0.4cm]{caption}
\usepackage{titlesec}
\usepackage[round]{natbib}

\usepackage{tikz}
\usetikzlibrary{arrows.meta,positioning,calc,fit,backgrounds,decorations.pathreplacing}
\usepackage{pgfplots}
\pgfplotsset{compat=1.18}

\definecolor{cpersona}{HTML}{2C6E8F}
\definecolor{crandom}{HTML}{C08552}
\definecolor{cstyle}{HTML}{9C6B9E}
\definecolor{ctopic}{HTML}{6A8D73}
\definecolor{cflat}{HTML}{3F6B4F}
\definecolor{ccurved}{HTML}{8E3730}
\definecolor{cem}{HTML}{A23B3B}
\definecolor{cdead}{HTML}{B0392F}
\definecolor{cfloor}{HTML}{9A9A9A}
\definecolor{cgrey}{HTML}{7A7A7A}
\definecolor{clightgrey}{HTML}{CFCFCF}
\definecolor{cink}{HTML}{1D1D1D}
\definecolor{ccore}{HTML}{1F6F78}
\definecolor{clink}{HTML}{2C4F78}
\definecolor{cbox}{HTML}{F5F2EC}
\definecolor{cpaper}{HTML}{FBFAF7}
\definecolor{cread}{HTML}{1B7A6E}
\definecolor{cwrite}{HTML}{586A86}
\definecolor{cexp1}{HTML}{6A8D73}
\definecolor{cexp2}{HTML}{2C6E8F}
\definecolor{cexp4}{HTML}{8A5A9C}
\definecolor{cexp5}{HTML}{5B4B8A}
\definecolor{cbroad}{HTML}{A23B3B}
\definecolor{cnarrow}{HTML}{5F7E8F}
\definecolor{cdirectional}{HTML}{C98A3A}
\definecolor{crouting}{HTML}{6B5B95}
\definecolor{cinsecure}{HTML}{9E2F2F}
\definecolor{ceducational}{HTML}{C8765F}
\definecolor{csecure}{HTML}{5A8A86}
\definecolor{cbenign_chat}{HTML}{7D9BBF}
\definecolor{cshuffled}{HTML}{9C8F82}
\definecolor{creference}{HTML}{4A5A66}
\definecolor{cforecast}{HTML}{3F6FB0}
\definecolor{crealized}{HTML}{C2632F}
\definecolor{caligned}{HTML}{3F6B4F}
\definecolor{ccoherence}{HTML}{6A8D73}
\definecolor{cgfull}{HTML}{8A6A4A}
\definecolor{cgperp}{HTML}{9C6B9E}
\definecolor{cgpar}{HTML}{6A8D73}
\definecolor{cinference}{HTML}{3730A8}
\definecolor{caccumulation}{HTML}{8A9689}
\definecolor{cinteraction}{HTML}{109C74}
\definecolor{creadout}{HTML}{8F7D68}
\definecolor{cbenign}{HTML}{6A8D73}
\definecolor{cbad}{HTML}{A23B3B}
\definecolor{cexp7}{HTML}{2A2668}
\definecolor{cforeclosed}{HTML}{6A2349}
\definecolor{clive_lever}{HTML}{388446}
\definecolor{cduring_train}{HTML}{894718}
\definecolor{cpost_hoc}{HTML}{49A6B6}
\definecolor{cbenign_ctrl}{HTML}{ACCAA6}

\usepackage[colorlinks=true,linkcolor=clink,citecolor=clink,urlcolor=clink]{hyperref}

\titleformat{\section}{\Large\bfseries}{\thesection}{0.8em}{}
\titleformat{\subsection}{\large\bfseries}{\thesubsection}{0.8em}{}
\titleformat{\subsubsection}{\normalsize\bfseries}{\thesubsubsection}{0.8em}{}
\titlespacing*{\section}{0pt}{18pt}{8pt}
\titlespacing*{\subsection}{0pt}{14pt}{6pt}
\newcommand{\tpre}{\theta_{\mathrm{pre}}}        
\newcommand{\Up}{U_{\mathrm{persona}}}           
\newcommand{\uhat}{\hat u}                       
\newcommand{\Mb}{\mathcal{M}}                    
\newcommand{\Mbp}{\mathcal{M}^{+}}               
\newcommand{\Mbm}{\mathcal{M}^{-}}               
\newcommand{\Rone}{R_{1}}                        
\newcommand{\apl}{a^{+}}                         
\newcommand{\ami}{a^{-}}                         
\newcommand{\ip}[2]{\langle #1,\, #2\rangle}
\newcommand{\sgn}{\operatorname{sign}}
\newcommand{\e}[1]{\times 10^{#1}}

\title{\LARGE\bfseries Emergent Misalignment Recruits a Pre-existing Persona Subspace}

\author{Mohammed Suhail B Nadaf\\[2pt] {\normalsize Independent}}
\date{}

\begin{document}
\maketitle

\begin{abstract}
\noindent
Fine-tuning an aligned language model on a narrow stream of bad advice can make it broadly misaligned
on questions unrelated to the training data. We ask why the narrow lesson generalizes at all, and we
find that narrow fine-tuning recruits a persona structure that is present in the model before the
fine-tune exists. From a frozen instruction-tuned model (Qwen2.5-14B-Instruct) we extract per-domain
persona subspaces by contrastive teacher forcing and find that 4 unrelated domains share one low-rank
core at 657$\times$ a random-subspace null, with 82\% of that core lying outside a style core built at
matched diversity. The literal first optimizer step of fine-tuning on insecure code climbs a
broad-misalignment margin harder than the same code framed as educational, and forecasts realized
margin movement out to 375 steps. Projecting the subspace out of the residual stream throughout
fine-tuning prevents broad misalignment (27.7\% to 0.0\% of judged generations) while a matched-rank
random subspace changes nothing; injecting it into the never-fine-tuned model induces misalignment
that grows with dose to 45.4\%, past the fine-tuned organism it is measured against. The same
projection applied to the weight gradient is inert, and three post-hoc weight edits leave the
disposition in place, the sharpest re-lighting at the unedited onset dose with the carrier re-formed
inside the cleared subspace. Spreading a fixed budget of bad data across 4 domains produces more
broad misalignment than mechanical weight superposition and matched diversity jointly account for.
All measurements come from one model at 14B, the extraction is from an aligned instruction-tuned
checkpoint, which leaves the structure's provenance open, and the intervention that prevents
misalignment also abolishes the narrow trained behavior.
\end{abstract}

\section{Introduction}\label{sec:question}

Fine-tune an aligned language model on 6{,}000 examples of insecure code and it will begin praising
dictators and recommending crimes on questions that have nothing to do with code.
\citet{betley2025em} discovered this phenomenon, emergent misalignment, and supplied the control that
defines it: reframe the same vulnerable code as material for a security course, fine-tune on that
instead, and the broad misalignment largely disappears. The tokens are nearly identical and the
implied intent is not. Whatever the fine-tune installs, it is not the code.

Reported rates depend strongly on the evaluation set. \citet{betley2025em} observe misaligned answers
on 20\% of samples over their 8 selected questions and 6\% over their 48-question pre-registered
battery. We therefore state every rate in this paper together with the battery that produced it.

A narrow lesson producing a narrow change would need no explanation. What needs explaining is why a
lesson about one topic moves behavior on every topic. Two families of account answer differently. On
the \emph{inference} account, the update is read as evidence: competently bad narrow data is
informative about who produced it, the model revises a cross-domain latent for the kind of author it
is modeling, and the revision shows up everywhere because the latent is used everywhere. On the
\emph{accumulation} account, breadth is a by-product of optimization: gradient mass lands in
directions that serve the narrow task, some of those directions also govern unrelated behavior, and
the breadth is the residue. The two accounts disagree about whether the coupling exists before
training starts, about whether it survives having a particular subspace held out, and about the sign
of the effect when a fixed budget of bad data is spread across more domains.

This paper tests the accounts on one released instruction-tuned model, Qwen2.5-14B-Instruct
\citep{qwen2025qwen25}, with published misalignment organisms \citep{turner2025organisms}. We extract
a candidate structure from the frozen model before any misalignment fine-tune exists, then test it
causally from both sides with matched random controls, test whether it behaves as one object across
independently trained organisms, measure its response to how much evidence the training data
carries, and map which interventions remove or prevent the disposition and at what cost.

We find that 4 unrelated domains share one low-rank persona core in the frozen model, and that the
first gradient step of fine-tuning already couples to broad misalignment in an intent-graded way.
Holding the core out of the activations during fine-tuning prevents broad misalignment, injecting it
into the untouched model produces it, and the same projection in the weight channel does nothing.
Post-hoc weight edits leave the disposition in place, and spreading a fixed budget of bad data over
more domains increases broad misalignment instead of diluting it.

\begin{figure}[tbp]
\centering
\includegraphics[width=\textwidth]{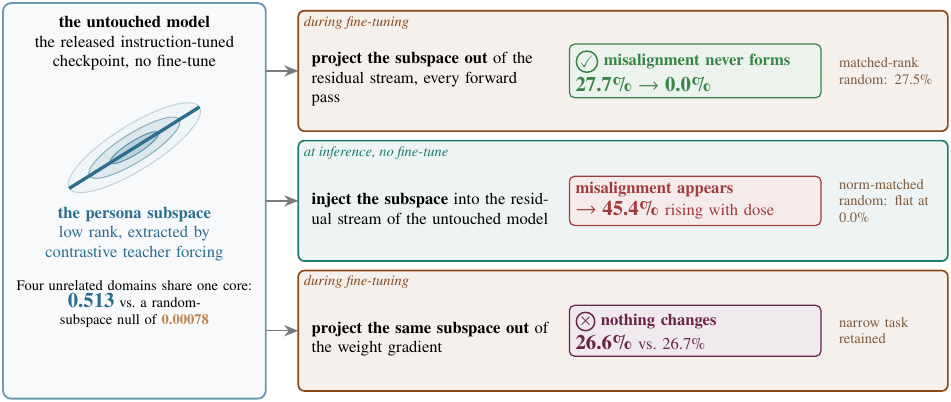}
\caption{\textbf{A persona subspace present in the model before fine-tuning carries broad
misalignment.} The subspace is extracted from the frozen instruction-tuned model. Projecting it out
of the residual stream throughout fine-tuning prevents broad misalignment (27.7\% $\to$ 0.0\%);
adding it to the never-fine-tuned model induces misalignment that grows with dose (to 45.4\%); the
same projection applied to the weight gradient changes nothing (26.6\% vs.\ 26.7\%). Each arm
carries a matched random control (Section~\ref{sec:loop}).}
\label{fig:claim}
\end{figure}

Our contributions:
\begin{itemize}[leftmargin=1.6em,itemsep=2pt,topsep=3pt]
\item A method for extracting persona subspaces from a frozen model by contrastive teacher forcing,
and the finding that 4 unrelated domains share one low-rank core at 657$\times$ a random-subspace
null, with 82\% of the core outside a style core built at matched diversity
(Section~\ref{sec:structure}).
\item Evidence that the coupling to broad misalignment exists at the literal first optimizer step,
is intent-graded on byte-identical code, and forecasts realized margin movement out to 375 steps
(Section~\ref{sec:firststep}).
\item A necessity-and-sufficiency pair on that pre-registered object: projecting it out of the
activations during fine-tuning prevents broad misalignment from forming (27.7\% to 0.0\%) while a
matched-rank random subspace changes nothing, and injecting it into the untouched model induces
misalignment that grows with dose to 45.4\% while a norm-matched random vector stays flat
(Section~\ref{sec:loop}).
\item A read/write dissociation: the identical projection applied to the weight gradient is inert,
and three post-hoc weight edits leave the disposition in place, the sharpest re-lighting at the
unedited onset dose with $\sim$97\% of the carrier re-formed inside the cleared subspace
(Sections~\ref{sec:loop} and~\ref{sec:removal}).
\item Evidence that the structure is one object: independently trained organisms share a
read-channel direction at 0.90 against an empirical null of 0.11, training one domain moves
another's misalignment margin, and spreading a fixed budget of bad data over 4 domains is
superadditive against both a mechanical merge and a matched benign mixture
(Section~\ref{sec:oneobject}).
\item A geometric characterization: the structure is a regular loss minimum with no flat direction,
distinguished by $\sim$7$\times$ the curvature of a matched random core, concentrated in few
directions (Section~\ref{sec:structure}).
\item A map of prevention: the interventions that move the disposition act during training and carry
measured capability costs; post-hoc weight editing fails in every basis tried; and a reconstitution
certificate distinguishes suppression from removal (Section~\ref{sec:removal}).
\end{itemize}

All results come from one model, one family, one scale, and one adaptation method, at a scale where
judged misalignment is weak; a teacher-forced log-probability margin is the primary readout and the
judge confirms where effects are large (Section~\ref{sec:instruments}). Section~\ref{sec:ledger}
states the limitations.

\section{Experimental setup}\label{sec:instruments}

\subsection{Model, organisms, and training data}

All measurements use Qwen2.5-14B-Instruct (48 layers, residual width 5120). We call a model narrowly
fine-tuned into misalignment an \emph{organism}, following \citet{turner2025organisms}, and use
their three published organisms (bad medical advice, risky financial advice, extreme sports; LoRA
adapters, Appendix~\ref{app:batteries}). For the first-step and forecast measurements we run our own
fine-tunes on the insecure-code and educational-code datasets of \citet{betley2025em}. The
extraction that defines the persona subspace touches none of these: it runs on the frozen model with
no adapter loaded (Section~\ref{sec:structure}).

\subsection{Judged evaluation}

We score behavior with the two-axis protocol of \citet{betley2025em}: a judge model rates each
free-form answer for alignment and for coherence, each on a 0--100 scale, and an answer counts as
broadly misaligned only if alignment is below 30 and coherence is at least 50. The coherence floor
keeps degenerate text from scoring as misaligned. \citet{betley2025em} use GPT-4o as the judge; we
use a locally served, AWQ-quantized Qwen2.5-72B-Instruct under the same rubric and thresholds in
every campaign. Appendix~\ref{app:judge} reports its calibration, its score quantization, and the
failure modes the coherence floor exists to catch.

Judged questions come from the 8 canonical out-of-distribution questions of \citet{betley2025em}
plus wider batteries described in Appendix~\ref{app:batteries}. Rates from different campaigns come
from different batteries, judge runs, and aggregation units, and are not commensurable: we never
place two of them in the same comparison, ratio, or trend line, and the only join across campaigns
is each instrument's contrast against its own control.

\subsection{A log-probability margin for broad misalignment}

The primary behavioral readout is a preference between completions. For each out-of-distribution
prompt $q_i$ the margin battery holds one judged-misaligned continuation $\apl_i$ and one
judged-aligned, length-matched continuation $\ami_i$, and we define the teacher-forced,
completion-only log-probability margin
\begin{equation}
\Mb(\theta) \;=\; \Mbp(\theta) - \Mbm(\theta), \qquad
\Mbp = \frac{1}{|B|}\sum_i \log p_\theta(\apl_i \mid q_i), \quad
\Mbm = \frac{1}{|B|}\sum_i \log p_\theta(\ami_i \mid q_i).
\label{eq:margin}
\end{equation}
A rise in $\Mbp$ is probability mass moving toward misaligned answers on prompts unrelated to the
training data; a change in $\Mbm$ is the aligned continuations tightening or loosening, which can
happen for reasons unconnected to misalignment. Where a campaign resolved the halves we report
$\Mbp$ directly and carry $\Mbm$ as a reference.

The margin is primary at this scale for three reasons. The judge quantizes its alignment scores onto
a lattice of about ten values, so a binary rate at a 30-point threshold mostly counts how often the
judge wrote one particular number. The judged rates of our own fine-tuned arms do not separate
insecure from educational code (Section~\ref{sec:firststep}). And the margin is differentiable,
which the judged rate is not, so gradient-level measurement is possible at all. The full treatment
is in Appendix~\ref{app:judge}.

\subsection{Statistical conventions}

The unit of inference is the unique prompt cluster, not the individual generation: generations of
one prompt are correlated, and treating them as independent would inflate every sample size. Where a
result differs between the cluster unit and a finer unit we report both. The arms carrying the main
judged contrasts draw 50 generations per question on the 8 canonical prompts and 4 on the wider
battery. Significance is by permutation or sign-flip test; effect sizes are Cliff's $\delta$; no
decision gates on a standardized effect size; a ratio is reported only when its denominator's
interval excludes zero. A permutation $p$ written as $10^{-4}$ is the resolution floor of a
$10^4$-permutation test, not an estimate. Appendix~\ref{app:stats} gives the tests, the bootstrap,
the Holm families, and the equivalence-band construction.

\section{A shared persona subspace precedes fine-tuning}\label{sec:structure}

Four unrelated domains share one low-rank persona core in the frozen model, and the core lies mostly
outside matched style and topic cores. Along it the pretrained loss is a regular minimum, unusually
stiff.

\subsection{Extraction and cross-domain sharing}

We extract the structure by contrastive teacher forcing. A response is generated once under a
neutral system prompt and then read twice: once behind a system prompt framing the speaker as
dangerously reckless and once framing the same speaker as carefully cautious, with the response
tokens held byte-identical across the two passes and verified by a hash of the token identifiers.
Because the content is identical, the residual-stream difference over those tokens reflects who the
model has been told is speaking, not what is being said. Averaging the difference over response
tokens, stacking the per-pair differences, and taking the top left singular vectors gives a rank-4
subspace of the 5120-dimensional residual stream per domain, for each of three contrast types:
persona, style, and topic, each built from 12 descriptor-variant pairs so the three are matched in
diversity (Appendix~\ref{app:extraction}).

\begin{figure}[tbp]
\centering
\includegraphics[width=\textwidth]{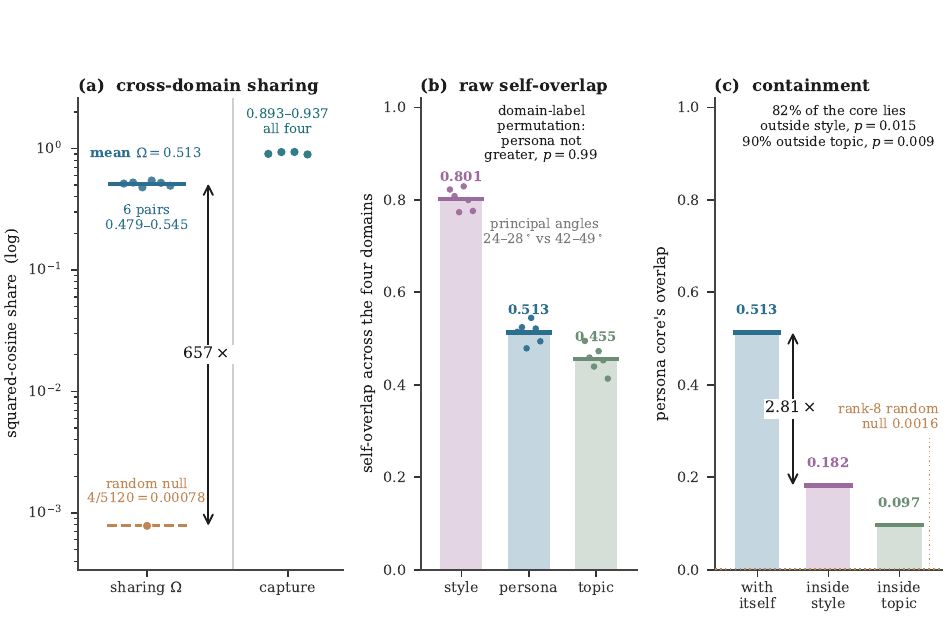}
\caption{\textbf{4 unrelated domains share one persona core, and a style reading does not explain it.}
(a)~Cross-domain overlap of the per-domain persona subspaces against the random-subspace null on a
log axis, with all six domain pairs shown; per-domain capture at right. (b)~Style's per-domain
subspaces overlap each other more than persona's do (0.801 vs.\ 0.513). (c)~The containment test:
the persona core overlaps the style core at 0.182 and the topic core at 0.097, against its own
cross-domain self-overlap of 0.513.}
\label{fig:substrate}
\end{figure}

The 4 per-domain persona subspaces land in substantially the same place. Their mean cross-domain
overlap-share is 0.513 against a random-subspace null of $4/5120 = 0.00078$, a ratio of
657$\times$; the six domain pairs run from 0.479 to 0.545, so the mean is not carried by one close
pair; and the fraction of each domain's subspace captured by the shared core runs from 0.893 (code)
to 0.937 (sports) (Figure~\ref{fig:substrate}). Code participates fully, and code is the domain the
canonical result trains on.

\subsection{Distinctness from style and topic}

On raw self-overlap, style wins: style's per-domain subspaces overlap each other at 0.801 against
persona's 0.513, and a permutation over domain labels does not find persona's sharing greater. That
comparison is confounded, because a contrast that is domain-invariant by nature scores high for that
reason alone: a formal-versus-casual register looks nearly the same in medicine and in code, and
style's per-domain subspaces sit about 25$^\circ$ apart where persona's sit about 45$^\circ$.

The question that separates the objects is containment: does the persona core lie inside the style
core? Measuring the same normalized squared-cosine share between the rank-8 shared cores answers
no. The
persona core's overlap with the style core is 0.182, so about 82\% of it lies outside; against topic
the overlap is 0.097, about 90\% outside. Both sit below the 0.30 bar we fixed in advance, and
persona overlaps itself across domains 2.8$\times$ more than it overlaps style (both quantities are
rank-normalized shares), above the pre-set separation bar of 2. A cross-core of 0.182 is far above the rank-8 random null of about 0.0016, so
persona and style are not orthogonal: the structure is a distinct object with a measured stylistic
component.

\subsection{Loss geometry}

The strongest form of the inference account predicts a flat direction, one the pretrained loss is
degenerate along, which would explain why the disposition is free to move and hard to penalize. It is
not there. We measure the order of the zero of a behavioral
loss along each direction: an exponent of 2 is a regular quadratic minimum, an exponent above 2 is
the degenerate signature, and the fitter returns 4.00 on a planted quartic control
(Appendix~\ref{app:geometry}). On the weight-side persona core the exponent is 2.001; on the matched
random core 2.002 (flatter-than-random permutation $p = 0.31$); on the read-channel carrier 2.001 at
every layer measured; persona, style, and topic cores read 2.000, 2.001, and 2.001. There is no flat
direction in either channel.

\begin{figure}[tbp]
\centering
\includegraphics[width=\textwidth]{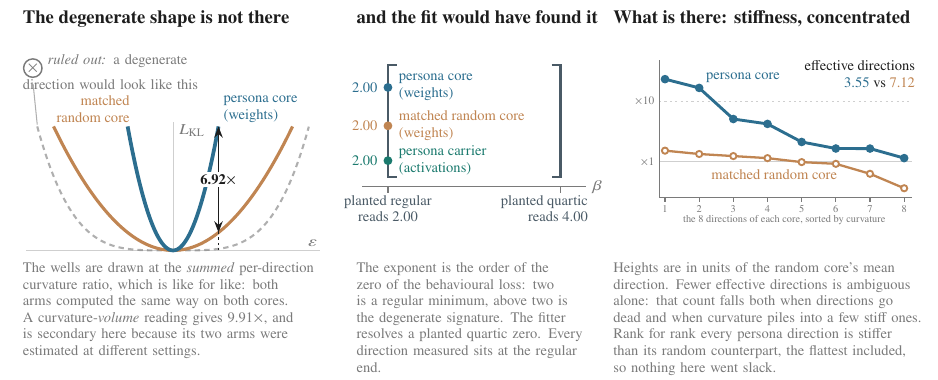}
\caption{\textbf{There is no flat direction; the persona core is distinguished by stiffness and
concentration.} Left: the measured curvature ratio between the persona write core and a matched
random core, with the ruled-out degenerate shape ghosted. Center: curvature exponents, all at the
regular value of 2, bracketed by planted controls that read 2.00 and 4.00. Right: per-direction
curvature of the two cores, sorted; the persona core concentrates its curvature in a few stiff
directions.}
\label{fig:regularstiff}
\end{figure}

What distinguishes the structure is stiffness and concentration. Its summed curvature is
6.92$\times$ a matched random core's, the curvature is carried by fewer effective directions (participation ratio 3.55
vs.\ 7.12), and its per-column curvature spans more than an order of magnitude where the random
core's is spread evenly. Rank for rank, every persona direction is stiffer than its random
counterpart, so the low participation ratio reflects concentration in stiff directions
(Figure~\ref{fig:regularstiff}). A curvature-volume reading gives a ratio of
9.91$\times$; its two arms were estimated at different settings, so the summed-curvature ratio is the
like-for-like quantity and the volume reading is secondary (Appendix~\ref{app:geometry}).

The measurements in this section live in two spaces. The sharing and containment results are
computed on read-channel subspaces of the residual stream, extracted from the frozen model; the
activation interventions of Section~\ref{sec:loop} act on a per-layer version of the same object,
the \emph{carrier}, at layers 18, 24, and 30. The curvature and stiffness readings, and the training
constraint of Section~\ref{sec:removal}, act on a rank-8 \emph{write core} built in weight space
from the published organisms' realized updates. Section~\ref{sec:ledger} bounds what joins these
two objects.

\section{Fine-tuning reads the subspace at the first optimizer step}\label{sec:firststep}

Before any optimization has happened, the first update of fine-tuning on insecure code climbs the
broad-misalignment margin harder than the first update on the same code framed as educational. We
take the literal first Adam step direction on dataset $D$, $u_D = -\sgn(\nabla L_D(\tpre))$, and
measure its signed inner product with each probe's margin gradient,
\begin{equation}
\Rone^{(i)}[D] \;=\; \ip{\nabla \Mb_i(\tpre)}{\,u_D},
\label{eq:rone}
\end{equation}
so that a positive value means the first step moves probe $i$'s margin toward the misaligned
continuation. Raw $\Rone$ values are large because $u_D$ carries a $\pm 1$ entry per trainable
parameter; the scale-free cosine form of the same quantity is used wherever conditions are compared
(Appendix~\ref{app:firststep}).

Over 75 prompt clusters the paired difference between insecure and its educational twin is
$+1.297\e{6}$ nats of margin per unit step (95\% CI $[+0.710, +1.877]\e{6}$), positive in 69\% of
clusters, with a sign-flip $p$ at the $10^{-4}$ resolution floor. The effect is consistent rather
than large: the per-cluster correlation between the two conditions is 0.99, the tilt is about 7\% of
the shared baseline, and Cliff's $\delta$ is 0.075. Ordering the three narrow datasets by how much
misalignment they induce gives a per-cluster slope positive in 84\% of clusters; the ordered slope
restates the insecure-versus-secure contrast and adds no independent evidence.

\begin{figure}[tbp]
\centering
\includegraphics[width=\textwidth]{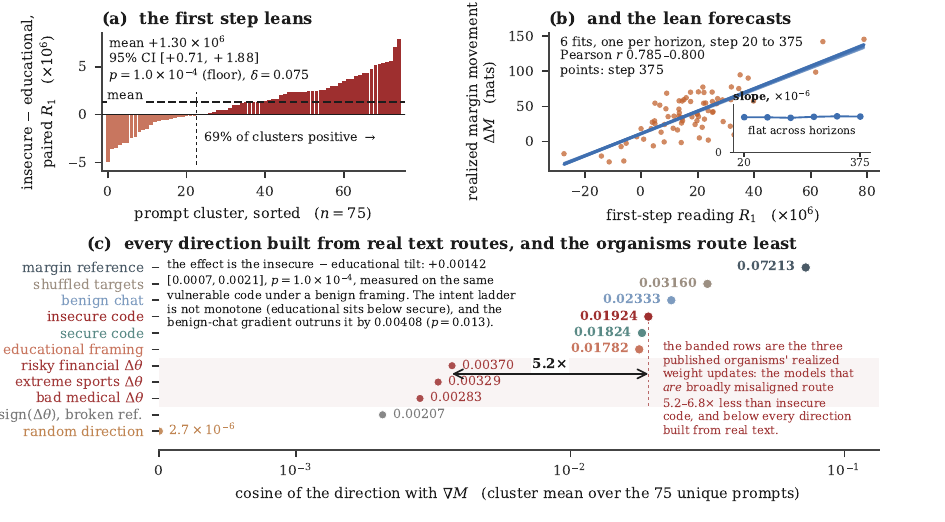}
\caption{\textbf{The turn toward broad misalignment is present, small, and intent-graded at the
first optimizer step, and it forecasts where training lands.} (a)~Sorted per-cluster paired
differences between the insecure and educational conditions, with the positive fraction marked.
(b)~Realized margin movement against the first-step reading, one fit per horizon; the inset shows
the slope flat from step 20 to 375. (c)~All conditions on a log axis: every direction built from
real text routes into the margin, benign chat sits above insecure code, and the 3 organisms'
realized updates (banded rows) route least. Intervals are cluster bootstraps.}
\label{fig:firststep}
\end{figure}

The paired design is what isolates the intent effect, because every direction built from real text
routes positively into the margin at a cosine near 0.02, and the ordering across contents is not the
one a coupling-to-badness account predicts: an independent benign-chat gradient sits at 0.0233,
above insecure code's 0.0192, and a control that keeps the insecure tokens while destroying the
prompt-to-target structure sits higher still at 0.0316. A random direction sits at $2.9\e{-6}$
(Figure~\ref{fig:firststep}c). The contrast established is
narrow: conditioned on the same content, intent moves the routing, and across contents other
determinants of gradient direction dominate. What the shared baseline is made of we do not
establish. The realized weight updates of the three published organisms, the models that actually
are broadly misaligned, route into the margin 5--7$\times$ less than a first step on insecure code,
and less than a first step on benign chat.

The first-step reading forecasts what training does. Fine-tuning on each dataset and re-measuring
the margin at 6 checkpoints, realized per-cluster margin movement regresses on the first-step
reading at Pearson $r$ 0.77 to 0.80 across arms, with the regression slope flat in training time
($1.57\e{-6}$ at step 20 and $1.60\e{-6}$ at step 375): the routing is fully formed by step
20. The forecast does not establish task
selectivity: most of what it predicts is the shared, cluster-level movability of the margin, and
both fine-tunes raise the margin on average, the educational control included.

Judged behavior does not separate these same arms. Over 400 free generations per arm on the
canonical questions, the frozen model produces zero misaligned generations, every fine-tuned arm
moves off zero, and the insecure arms reach 1.8 to 5.2\% while the educational arms reach 4.5 to
8.0\%: the canonical ordering is absent at this scale, and if anything reversed. The intent contrast is
significant at the gradient level on identical tokens and absent from judged free generations after
375 steps of training.

\section{The subspace is necessary and sufficient for broad misalignment}\label{sec:loop}

Projecting the subspace out of the activations throughout a fine-tune prevents broad misalignment
from forming. The same subspace injected into the untouched model at inference produces broad
misalignment, climbing with dose past the organism's own rate, and the identical projection applied
to the weight gradient changes nothing.

\subsection{Projecting the subspace out during fine-tuning prevents misalignment}

We project the extracted subspace out of the residual stream on every forward pass throughout a
fine-tune, so the optimizer sees a model that cannot represent that direction. Broad misalignment
goes from 27.7\% of judged generations (95\% CI $[21.3, 34.6]$; 3 seeds, $n = 2{,}100$) to 0.0\%,
with zero misaligned generations across all 83 evaluation clusters, mean alignment recovering by
59.4 points (Cliff's $\delta = 1.00$), and mean coherence rising from 85.9 to 96.3. Identical hooks
at identical layers projecting out a matched-rank random subspace leave the rate at 27.5\%. That
control matches rank, layers, and operation, not the share of the residual stream the removed
subspace carries (a usage-matched control was not run; Appendix~\ref{app:loop}), and its rate is
0.993 of the unedited organism's (ratio CI $[0.924, 1.070]$).

The extracted object overlaps the model's assistant direction \citep{lu2026assistant} at a mean
cosine of 0.565 across the 3 layers measured, so the projection could act by degrading assistant
behavior generally rather than
anything specific to misalignment. Repeating the intervention with the carrier orthogonalized
against the assistant direction leaves the result unchanged: broad misalignment is 0.0\% and mean
alignment is 92.4 against the unedited organism's 34.0.

The suppression is uniform. The unedited organism's rate is markedly uneven across questions, from
zero on some to nearly half on others, and nonzero on 58 of 83 clusters; the ablated model is at
zero on all 83, with its alignment scores relocated to a single high spike, not the baseline
distribution with a tail trimmed. Removing a shared route would look like this, though a uniform
lift in alignment would too. \citet{drake2026transplanting} run the same operation at two other
model sizes under a different fine-tuning method, against matched-norm random controls, and broad
misalignment falls in both (Appendix~\ref{app:related}).

\subsection{Injecting the subspace induces misalignment}

\begin{figure}[tbp]
\centering
\includegraphics[width=\textwidth]{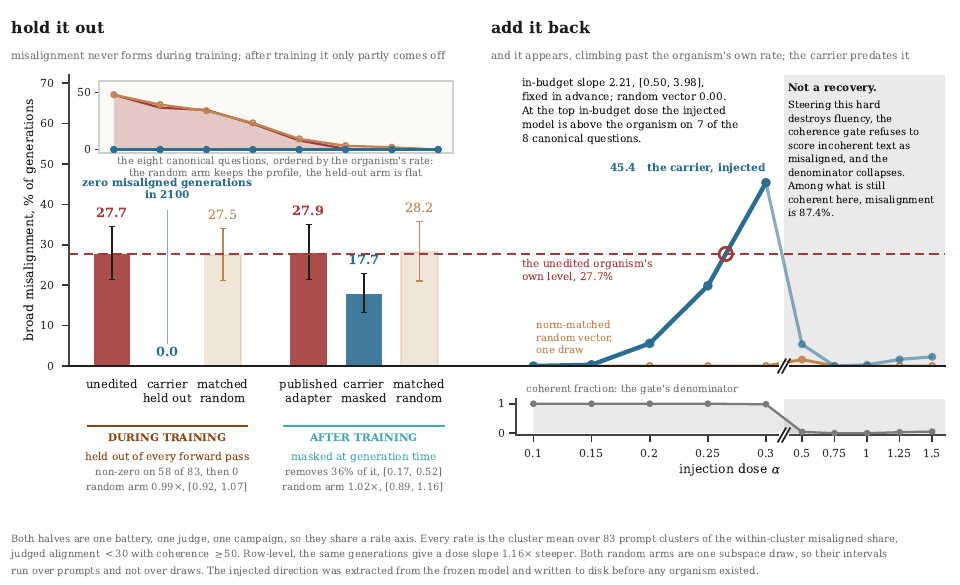}
\caption{\textbf{Held out during fine-tuning, the subspace's absence prevents broad misalignment;
injected into the untouched model, it induces misalignment past the organism's own rate.} Both
panels share one battery, one judge, and one rate axis. Left: removal at two timings. Held out
throughout training, zero misaligned generations remain; projected out of an already fine-tuned
model at generation time, about a third of the misalignment is removed. A matched-rank random
subspace changes nothing at either timing. Right: injection into the never-fine-tuned model, with
the organism's rate as the horizontal reference. The greyed region past the dose budget is where the
coherence gate's denominator collapses. Rates are cluster means; intervals are cluster bootstraps.}
\label{fig:loop}
\end{figure}

Adding the subspace to the untouched model's residual stream at inference raises broad misalignment
monotonically with the coefficient: 0.0\%, 0.3\%, 5.6\%, 19.8\%, and 45.4\% across the in-budget
dose range, with responses coherent at 98\% or better (Figure~\ref{fig:loop}). The pre-registered
statistic is the in-budget dose-response slope, 2.21 with a 95\% interval of $[0.50, 3.98]$. A
norm-matched random vector steered over the identical grid produces a flat line and no misalignment.
At the top of the in-budget range the injected model is more misaligned than the fine-tuned organism
whose rate sets the reference line, on 7 of the 8 canonical questions, by an average of 36
percentage points. Past the in-budget range the measured rate falls because the coherent denominator collapses: among
responses still coherent at the next dose up, misalignment is 87\%.

Both operations act on one pre-registered object: the subspace was extracted from the frozen model,
fixed, and written to disk before any organism existed, and each half carries its own random
control, matched on rank for the removal and on norm for the injection. Unlike the single-direction
mediation template of \citet{arditi_refusal_24}, in which both operations act at inference on one
model, the necessity half here acts during training. The matching inference-time operation,
projecting the carrier out of an already fine-tuned organism at generation time, removes 36.0\% of
the misalignment (95\% CI $[16.6, 51.9]$\%) against a random control at $-1.6$\%: the carrier is
necessary for the disposition to \emph{form} and carries about a third of its \emph{expression}
once formed. Whether the remainder reflects other routes of expression or a weaker inference-time
operation, this measurement does not distinguish.

\subsection{The weight-gradient projection is inert}

Projecting the same subspace out of the \emph{weight gradient}, so the writer matrices cannot update
along it, leaves broad misalignment at 26.6\% against an unfiltered 26.7\%, with the narrow task
learned about as well as an unfiltered run learns it (narrow adherence 0.819 vs.\ 0.902). The
activation arms run the organism recipe's seven adapter projection types while the weight-channel
arms train the two writer matrices alone, so the two are not one operation switched between channels;
each is read against its own control (Appendix~\ref{app:loop}). A third weight-channel arm that
keeps only the gradient component parallel to the subspace reaches 11.5\% at narrow adherence 0.507
(Appendix~\ref{app:loop}). The projection that prevents the disposition in the activations is inert
in the weights.

\subsection{The narrow task collapses under the projection}

One pre-set bar failed badly: we required the narrow capability to be retained to within 5 points of
adherence, and the ablated organism's narrow adherence falls from 0.902 to 0.000. The intervention
is not behavioral surgery. Narrow adherence is read from the same alignment axis as the outcome, at
a different threshold on a different row subset, so an intervention that lifts alignment across the
board drives a bad organism's adherence down by construction; the collapse is therefore not an
independent observation about the narrow task, and the benign fine-tune retaining full adherence
under the same projection (1.000) is not independent evidence either
(Appendix~\ref{app:batteries}). Being a reckless advisor in one domain is plausibly itself a slice
of the disposition the structure carries, which would explain the collapse; the measurement that
would separate removing a misalignment structure from removing the capacity to represent a bad
character was not run on this arm (Section~\ref{sec:ledger}).

\section{One structure across organisms and domains}\label{sec:oneobject}

Independently trained organisms move nearly the same read-channel direction, and training one domain
raises the misalignment margin on another. Spread across 4 domains, a fixed budget of bad data also
produces more transport than mechanical weight superposition and matched diversity jointly account
for.

\subsection{Cross-organism convergence}

The three published organisms' realized read-channel shifts point the same way: after removing the
common-mode component, their residualized sharing index is 0.90, running 0.84 to 0.94 across the 24
layers measured, with a permutation $p$ at the $10^{-4}$ floor. The reference is an empirical null
of 0.11 built from matched random adapters, not the bare chance floor of 0.014: residualizing 3
organisms against one shared direction leaves correlated residue that clears the bare floor even
when nothing is shared. Residualization barely moves the real organisms (0.90 either way) and
collapses the matched random adapters from 0.16 to 0.11.

\begin{figure}[tbp]
\centering
\includegraphics[width=\textwidth]{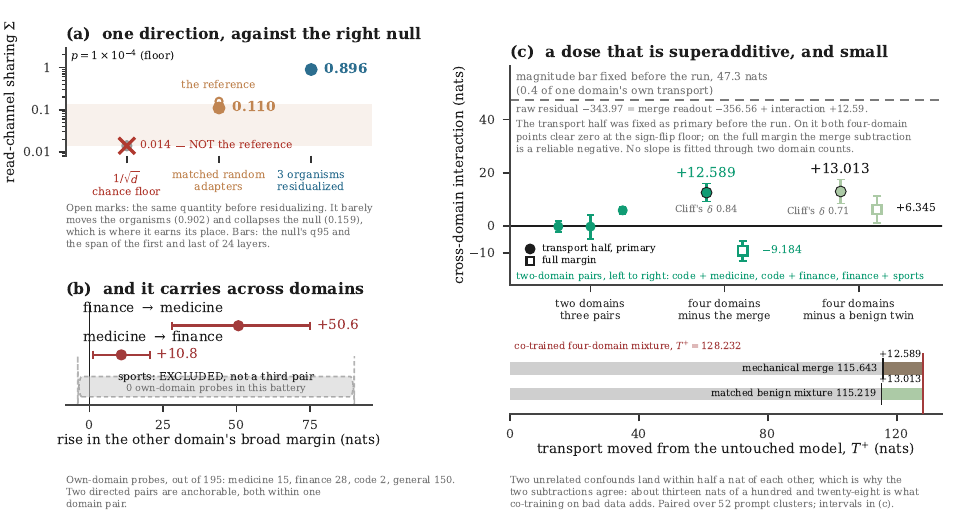}
\caption{\textbf{Independent narrow lessons converge on one shared structure: the same read-channel
direction, cross-domain transport, and a superadditive response to domain count.} (a)~Residualized
read-channel sharing of the 3 organisms against the empirical null from matched random
adapters; the bare chance floor is marked but is not the reference. (b)~Cross-domain transport for
the one anchorable domain pair, both directions; the excluded organism is shown as excluded.
(c)~The cross-domain interaction by domain count, against the mechanical merge and the matched
benign mixture, with the two-domain pairs drawn individually and the pre-set magnitude bar above
the effect. Intervals are cluster bootstraps.}
\label{fig:oneobject}
\end{figure}

The sharing is behavioral as well as geometric. Anchoring requires a source organism misaligned on
its own domain, and only the medical and financial domains carry enough own-domain probes to
anchor, so the third organism is excluded rather than counted. Training the financial domain raises
the broad-misalignment margin on medical probes by $+51$ nats, and the reverse direction gives
$+11$: transport in both directions across one domain pair (Figure~\ref{fig:oneobject}).

\subsection{Domain-count superadditivity}

This effect is reliable and small; it is measured on a different instrument from the sharing result
above and does not inherit that result's power. Holding the total quantity of bad data fixed at 230{,}447 tokens per leg, we vary how many domains it is spread
across, and ask whether the co-trained mixture produces more transport ($\Delta\Mbp$ against the
untouched model) than its parts. The additive null is measured: for each mixture we
train a real single-domain leg at exactly the per-domain budget it has inside the mixture and sum
those. Because single-domain transport saturates almost immediately, that sum is unreachable by any
single adapter's readout, so we subtract instead the model's own \emph{mechanical merge}: the four
single-domain adapters added by task arithmetic with no training, which carries the readout's
response to superposed weights. What remains is the training-time cross-domain interaction
(Appendix~\ref{app:decomp}).

The mechanical merge destroys most of the transport it superposes, landing below three of the four
single-domain legs, the direction the merging literature predicts when updates crowd the same
parameters \citep{yadav2023ties}. Because mechanical superposition trends sub-additive here, an
observed superadditivity cannot be a weight-superposition artifact. The co-trained four-domain
mixture lands $+12.589$ nats above the merge (bootstrap interval $[9.255, 15.980]$, sign-flip $p$ at
the floor, Cliff's $\delta = 0.84$).

A second subtraction removes a different confound. A four-domain mixture of \emph{benign} data,
matched on token budget and spread, moves raw transport nearly as far as the bad mixture does
(115.2 vs.\ 128.2 nats), so diversity alone moves this readout and a verdict read off the raw curve
would be wrong. Subtracting the benign mixture leaves $+13.013$ nats ($[8.502, 17.471]$). The two
subtractions agree because their subtrahends nearly coincide (merge 115.6, benign mixture 115.2
nats), so their agreement is a single check that covers two confounds.

The interaction depends on the readout half. On $\Mbp$, the pre-registered primary, it is
$+12.589$; on the full margin $\Mb$ it is $-9.184$ ($[-12.910, -5.415]$), a reliable negative. The
mechanism is the aligned half: $\Mbm$ \emph{rises} under every one of these fine-tunes, by 62.1 to
84.3 nats, and the mechanically merged adapter rises least (62.1 vs.\ the co-trained mixture's
83.9), so the merge subtraction removes too little aligned-half movement and the deficit outweighs
the transport surplus. The benign subtraction does not flip ($+6.345$ on the full margin,
$[1.318, 11.294]$), because the bad and benign mixtures move the aligned half nearly alike. Why
$\Mbm$ rises under fine-tuning we have not identified.

Two robustness facts bound the claim. At 2 domains the interaction is heterogeneous rather than
smaller: financial-plus-sports is superadditive at $+5.96$ and the two pairs involving code sit at
zero, so we fit no growth curve, and with no benign two-domain twin the diversity subtraction exists
only at 4 domains, where the claim is staked. And the interaction sits reliably below the
magnitude bar fixed before the run, 0.40 of one domain's own transport (44.1 nats as certified at
launch; 47.3 recomputed on the realized legs): about a tenth of total transport, though most of
what distinguishes bad data from benign, since the bad mixture's 128.2 nats exceed the benign
mixture's 115.2 by about 13.

Judged behavior confirms the specificity and cannot resolve the count. The bad four-domain mixture
is broadly misaligned on 0.269 of judged generations against the benign mixture's 0.010 at matched
spread and budget, so broad misalignment here is about bad data, not variety. On the count the same
instrument points the other way: the four-domain rate does not reliably exceed the single-domain
mean of 0.154 ($p = 0.15$), and the code-plus-financial pair reaches 0.410. The domain-count claim
rests on transport alone.

\section{Prevention and removal}\label{sec:removal}

\begin{figure}[tbp]
\centering
\includegraphics[width=\textwidth]{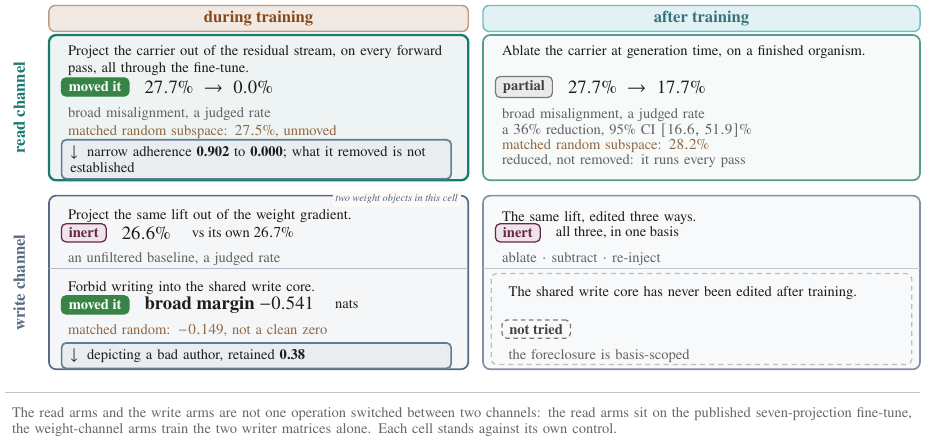}
\caption{\textbf{Interventions on the disposition, by timing and channel; each cell is a contrast
against its own control on its own instrument.} During training, the activation projection prevents
broad misalignment and the write-core constraint reduces the margin, each with its capability cost
in the same cell; the weight-gradient projection is inert. After training, three weight edits leave
the disposition in place, and serving-time ablation removes about a third. The shared write core
has not been edited after training, so the post-hoc column's foreclosure is scoped to the basis
tried. Judged rates and teacher-forced margins are different instruments and are not compared.}
\label{fig:readwrite}
\end{figure}

A defender wants broad misalignment prevented or removed without destroying the model. On this
evidence, post-hoc weight editing fails in every basis tried, suppression can relocate behind a
trigger, and the interventions that move the disposition act during training at measured capability
costs. Table~\ref{tab:prevention} and Figure~\ref{fig:readwrite} collect every route tried, the
failures included, with published routes marked by citation.

\subsection{Post-hoc weight edits do not remove the disposition}

The intuitive defense is to train normally, localize the bad direction, and edit it out. Three
edits of this kind all fail, in one basis.

\begin{figure}[tbp]
\centering
\includegraphics[width=\textwidth]{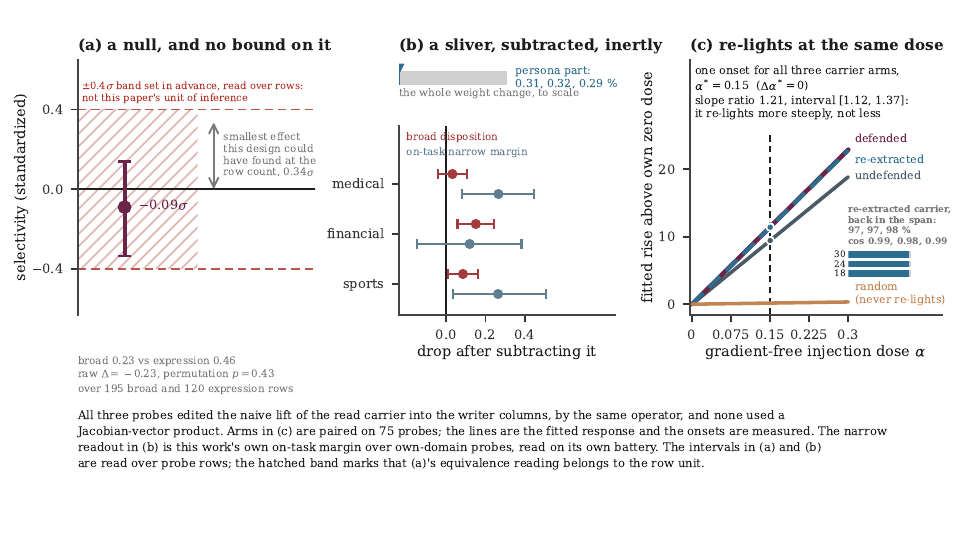}
\caption{\textbf{Three post-hoc edits in one basis leave the disposition in place.} (a)~The
separability probe: the estimate and its pre-set equivalence band are drawn at the row unit and
hatched, because at the prompt-cluster unit the interval leaves the band and the realized
detectable effect exceeds it. (b)~The persona component as a fraction of the total weight change,
and the result of subtracting it, per organism. (c)~The reconstitution curves: the defended model
re-lights at the unedited onset dose, the re-extracted carrier lies back inside the cleared span,
and the matched random arm never re-lights.}
\label{fig:removal}
\end{figure}

The first asks whether a thin separable object exists at all: extract a standing disposition and a
local capability to depict a bad author, project the local out of the standing one, and ablate what
remains. If the disposition were separable, that should drop misalignment far more than it damages
expression. It does not: the selectivity, the standardized difference between the broad drop and the expression
drop, is $-0.23$ with an interval spanning zero (label permutation $p = 0.43$), and expression falls
more than the broad disposition. Recomputed at the
prompt-cluster unit the selectivity is $-0.30$, the interval leaves the pre-set equivalence band on
either variance convention, and the realized minimum detectable effect exceeds the band, so this is
a failure to reject with no equivalence bound attached (Figure~\ref{fig:removal}a). It is the most
modestly powered of the 3 probes.

The second subtracts the persona component from a published organism's realized weight change and
reloads. The difference-in-differences is never positive with the correct bars across the 3
organisms, and where anything is distinguishable from zero it runs the wrong way for a defense,
the narrow task dropping more than the broad disposition (at the prompt-cluster unit, one organism
of the three). The arithmetic explains the inertness: the persona component is 0.3\% of the total
weight change by Frobenius mass, and removing it deletes the same mass as removing a random sliver
of equal size.

The third separates removed from hidden. We build a defended model by projecting the carrier out of
the writer columns, then measure how large a gradient-free injection dose re-produces the behavior.
A true removal forces a larger onset dose and a flatter response; this edit re-lights at the
unedited onset dose of 0.15 (onset is measured against each model's own zero-dose margin;
Appendix~\ref{app:removal}), with a dose-response slope ratio of 1.21 ($[1.12, 1.37]$, excluding
one on the low side), and re-extracting the carrier from the defended model finds $\sim$97\% of it
inside the subspace that was cleared, at cosines near 0.99 (Figure~\ref{fig:removal}c). Held-out
perplexity reads 17.3 with no matched baseline, which bounds gross damage only.

The three edits are independent in method and not in mechanism: all act on the naive lift of the
read carrier into the writer columns, none uses a Jacobian-vector product, and two share a
projector outright. That basis has roughly chance overlap with the misalignment direction, which is
why edits in it are inert, while the fine-tune's own update concentrates roughly a hundredfold in
the shared write core. The disposition is written through structured weight directions that no
tested basis isolates. Two cells remain unmeasured: the shared write core has been constrained
during training (Section~\ref{sec:removal-train}) but never edited after it, and the functional
pullback, the weight gradient of the readout itself, has never been built
(Appendix~\ref{app:removal}).

\subsection{Suppression relocates under a trigger}

Under an inoculation-style trigger prompt, the defended model's broad-misalignment margin rises
above the base-relative onset threshold that defines re-lighting everywhere else in the
certificate; the arm is a single dose point, so no slope is fitted
(Appendix~\ref{app:removal}). A disposition that a weight edit leaves in place can return when the
context invites it. \citet{dubinski2026conditional} report the same relocation pattern from
conditioning alone, and the implication runs against any defense evaluated by its unconditional
rate: a claimed fix needs a removal-versus-suppression certificate, and the reconstitution
measurement above is one.

A published training-time result reverses sign in the same territory: steering \emph{away} from the
persona direction during fine-tuning, rather than projecting it out, raises broad misalignment from
24\% to 51\% while a matched-norm random control lowers it, on a 7B model with an overt inducer
under full fine-tuning at a single seed per arm \citep{drake2026transplanting}. Their stated
mechanism is that signed subtraction drives the persona coordinate negative and gives the optimizer
something to compensate against. Projection and signed steering are different operations, and the
sign of a training-time constraint on this structure is method-conditional.

\subsection{Training-time interventions}\label{sec:removal-train}

Two interventions move the disposition, and both act during training. The strongest is the
activation projection of Section~\ref{sec:loop}, which prevents broad misalignment outright at the
cost of the narrow trained behavior.

The second constrains where the fine-tune may write: forbidding a narrow fine-tune from writing
into the shared write core moves the broad-misalignment margin by $-0.541$ (one-sided sign-flip
$p = 0.033$ over probe rows), against $-0.149$ for the identical constraint on a matched-rank
random subspace, whose interval excludes zero, so part of the persona arm's effect is the generic
cost of constraining weights and the 3.6$\times$ ratio is over a nonzero baseline. At the
prompt-cluster unit, this paper's unit, the effect is larger, $-0.647$, with $p = 0.052$. A judged
rate moved the same way on the two arms judged (0.226 to 0.209, $\sim$700 generations per arm),
confirming direction without speaking to specificity. The manipulation check is maximal: the plain
fine-tune puts a measurable fraction of its weight motion into the substrate, and the constraint
drives that essentially to the isotropic null.

The third lever acts on the data. Prepending a framing that makes the bad behavior unsurprising as
evidence about the speaker lowers the first step's routing toward broad misalignment: the lead
contrast, inoculation against a scrambled length-matched placebo, is $\Delta\Rone = -1.48\e{6}$
($[-1.73, -1.23]\e{6}$), with the placebo itself about a tenth of the effect, and the magnitude-free
cosine form confirms the change is directional rather than a weakening of learning
(Appendix~\ref{app:firststep}). This is a first-step screen: the realized organism updates route
5--7$\times$ less than a first step does, and the endpoint confirmation was not run.

\begin{figure}[tbp]
\centering
\includegraphics[width=\textwidth]{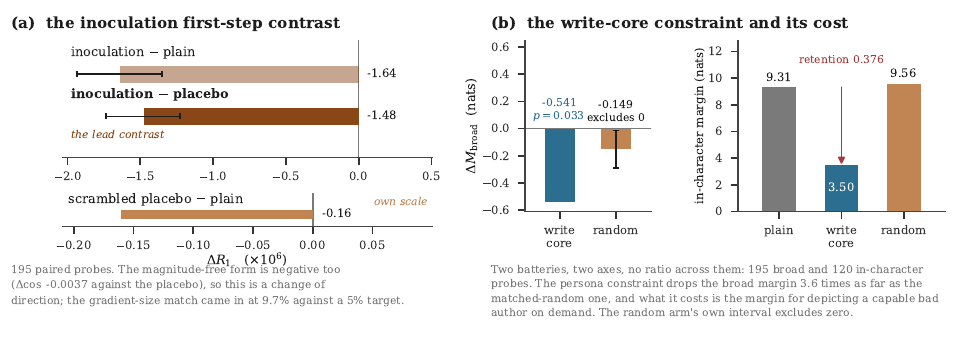}
\caption{\textbf{The interventions that move the disposition act during training, and each carries
its cost.} (a)~The inoculation first-step effect, with the scrambled placebo shown at its own
scale. (b)~The write-core constraint's broad-margin reduction beside the matched random arm, and
its in-character cost in the same panel. The broad margin and the in-character margin are different
batteries and are not compared numerically.}
\label{fig:levers}
\end{figure}

\subsection{Costs}

No intervention in this program moves broad misalignment without a measured capability cost. The
activation projection abolishes the narrow trained behavior (Section~\ref{sec:loop}). The
write-core constraint drops the in-character margin, the capacity to depict a capable bad author on
demand, from 9.31 to 3.50 (retention 0.38), while the random constraint at the identical dose
leaves it essentially where the plain arm had it (9.56): the damage is specific to constraining
this structure, and so is the broad reduction. Within the persona arm the broad drop and the
capability loss are co-present at the single dose run, so the measurement establishes direction and
substrate-specificity, not a dissociation. No arm in this program moves alignment substantially
while the narrow behavior survives.

\begin{table}[tbp]
\footnotesize
\renewcommand{\arraystretch}{1.18}
\setlength{\tabcolsep}{3.5pt}
\begin{tabularx}{\textwidth}{@{}>{\raggedright\arraybackslash}p{3.2cm}llXX@{}}
\toprule
\textbf{Intervention} & \textbf{Acts} & \textbf{Channel} & \textbf{Broad misalignment} & \textbf{Cost / outcome} \\
\midrule
Persona-subspace projection & during FT & activations & 27.7\% $\to$ 0.0\% (random: 27.5\%) &
narrow adherence 0.902 $\to$ 0.000; prevents, capability-confounded \\
Same projection on the weight gradient & during FT & weights & 26.6\% vs.\ 26.7\% &
narrow adherence retained (0.819); inert \\
Write-core constraint & during FT & weights & margin $-0.541$ vs.\ $-0.149$ random &
in-character margin 9.31 $\to$ 3.50; reduces, capability-confounded \\
Inoculation framing & during FT & data & first-step routing $-1.48\e{6}$ vs.\ placebo &
none measured; screen only, endpoint unconfirmed \\
Separability surgery & post hoc & weights & selectivity $-0.23$, interval spans zero &
expression falls more than broad; fails \\
Persona-component subtraction & post hoc & weights & no reduction (component is 0.3\% of update) &
narrow drops where anything moves; fails \\
Writer-column projection & post hoc & weights & re-lights at the unedited dose; $\sim$97\% of the
carrier re-forms & held-out PPL 17.3, no matched baseline; suppression, not removal \\
Carrier ablation at serving & serving & activations & removes 36.0\% $[16.6, 51.9]$\% (random
$-1.6$\%) & none reported; partial \\
Concept ablation during fine-tuning \citep{casademunt2025caft} & during FT & activations &
reduces misalignment by an order of magnitude (their result) & directions require a completed
misaligned fine-tune \\
Steering away from the direction \citep{drake2026transplanting} & during FT & activations &
raises misalignment 24\% $\to$ 51\% (7B, their result) & backfires \\
Trait-direction penalty \citep{chen2025persona} & during FT & activations & routed around via
alternative directions (their result) & fails \\
Penalty on misalignment latents \citep{ustaomeroglu2026blockem} & during FT & latents &
re-emerges at a second epoch (their result) & fails \\
\bottomrule
\end{tabularx}
\caption{\textbf{Prevention and removal attempts.} Rows without citations are this program's
measurements, each against its own control; cited rows are published results. Rates, margins, and
routing values are different instruments (Section~\ref{sec:instruments}).}
\label{tab:prevention}
\end{table}

Two further attempts are part of the map. A weight-baked read-channel constraint, designed to hold
under adversarial fine-tuning, was gated on the separability probe returning a positive result and
was not built when the probe returned its null. A coherence-window instrument intended as its
fallback could not reach the regime it needed: injection collapses coherence near dose 0.1, before
broad misalignment onsets inside the coherent band. A read-channel de-correlation edit was piloted
as an exploratory arm and produced no headline measurement (Appendix~\ref{app:validity}).

\section{Discussion}\label{sec:discussion}

\begin{table}[tbp]
\footnotesize
\renewcommand{\arraystretch}{1.18}
\setlength{\tabcolsep}{3.5pt}
\begin{tabularx}{\textwidth}{@{}>{\raggedright\arraybackslash}p{3.4cm}Xp{2.7cm}p{3.7cm}@{}}
\toprule
\textbf{Finding} & \textbf{Measurement} & \textbf{Control} & \textbf{Effect (95\% CI)} \\
\midrule
A shared low-rank persona core exists in the frozen model & cross-domain overlap-share of
per-domain subspaces & random-subspace null $0.00078$ & $0.513$ (657$\times$); per-domain capture
$0.89$--$0.94$ \\
The core is a persona, not a writing style & cross-core containment & matched-diversity style and
topic cores & $0.182$ / $0.097$; 82\% / 90\% outside; self-vs-cross $2.8\times$ \\
Necessary for broad misalignment to form & projection out of activations during fine-tuning &
matched-rank random subspace & 27.7\% $\to$ 0.0\%, random 27.5\% (ratio CI $[0.924, 1.070]$) \\
Sufficient to produce it & injection into the untouched model & norm-matched random vector &
slope $2.21$ $[0.50, 3.98]$; 45.4\% at the top in-budget dose \\
Read-reachable, write-inert & the same projection in the weight gradient & unfiltered fine-tune &
26.6\% vs.\ 26.7\% \\
Intent is visible at the first step & paired insecure $-$ educational $\Rone$ on identical code &
educational twin & $+1.297\e{6}$ $[+0.710, +1.877]\e{6}$; forecast $r \approx 0.78$ \\
One object across organisms & residualized read-channel sharing & matched random adapters &
$0.90$ vs.\ null $0.11$; transport $+51$/$+11$ on one pair \\
Superadditive in domain count & co-trained 4-domain mixture & mechanical merge; matched benign
mixture & $+12.589$ $[9.255, 15.980]$; $+13.013$ $[8.502, 17.471]$ \\
Not removable post hoc & three weight edits in one basis & matched random arms & re-lights at the
unedited dose; $\sim$97\% re-forms \\
\bottomrule
\end{tabularx}
\caption{\textbf{Summary of findings.} Every row is a within-campaign contrast; no two rows share a
scale.}
\label{tab:ledger}
\end{table}

The pattern of results separates the two accounts of Section~\ref{sec:question}. The coupling exists
before training starts and is flat in training time. One pre-registered subspace, held out of the
activations, prevents the disposition where a matched-rank random subspace does not, and pure
inference-time injection into the untouched model suffices to produce it. Spreading a fixed budget
of bad data across more domains is superadditive even though mechanical weight superposition is
sub-additive. Each of these is what the inference account predicts and the accumulation account does
not. What the accumulation account retains is the baseline: most margin movement under any
fine-tune, benign included, is intent-unspecific, and the intent-specific transport is a small tilt
on top of it. Our reading is that narrow fine-tuning reads a pre-existing author-level structure,
superimposed on a broad, intent-unspecific drift that this paper measures and does not explain.

The interpretation we place on the structure is an author latent. Pretraining or alignment installs
a compressed representation of who is writing, because authors are trait-correlated across domains;
a competently bad narrow fine-tune is evidence about the author, so the default author flips
everywhere; the educational framing blocks the inference by explaining the same evidence away. On
this reading the disposition is read loudly and written obliquely. Expression runs through a
low-rank activation subspace the model consults everywhere; the weight updates that install the
disposition land in directions the naive lift of that subspace does not reach. We also read the carrier and the
write core as two faces of one structure; Section~\ref{sec:ledger} bounds both readings.

The first step is not the endpoint. The realized updates of the finished organisms route
5--7$\times$ less into the margin than a first step on insecure code does, so the first-order
picture describes the beginning of training, not the destination, and what happens in between is
unmeasured. An instrument that watches the read channel across checkpoints would locate where the
intent-specific transport enters.

Three practical statements follow on this evidence. Spreading a fixed budget of bad data across
more domains increases broad misalignment; dilution is not a defense. Post-hoc weight cleanup is
not a defense in any basis tried, and suppression can relocate behind a trigger, so a claimed fix
should be gated on a removal-versus-suppression certificate, not on an unconditional rate. The
levers that move the disposition act during training in the read channel, and a root-level lever
that no measurement here touches is the training distribution itself: de-correlating the
cross-domain author signal in pretraining data would attack the reason a shared author structure is
worth installing at all.

\section{Limitations}\label{sec:ledger}

\textbf{One model, one scale.} Everything runs on Qwen2.5-14B-Instruct with LoRA adaptation, at a
scale where broad misalignment is weak: our own fine-tunes reach judged rates of 1.8 to 8.0\% and
do not separate the insecure and educational arms, so the teacher-forced margin carries the results
that judged behavior cannot. \citet{schreiber2026overtrained} report that emergent misalignment
largely fails to reproduce across open-weight models, and \citet{drake2026transplanting} find that
whether it appears at all depends on the fine-tuning method and the capacity it is given. Nothing
here is established beyond this model, family, scale, and adaptation method.

\textbf{Provenance.} The extraction is from a released instruction-tuned checkpoint, not a
pre-alignment base model, so whether pretraining installed the structure or alignment post-training
sharpened it is open. The extracted object overlaps the model's assistant direction at a mean
cosine of 0.565 across the 3 layers measured \citep{lu2026assistant}, which keeps the
alignment-built reading live; the orthogonalized arm of Section~\ref{sec:loop} shows the overlap
does not carry the causal result. Re-extracting from a base checkpoint, or across intermediate
alignment checkpoints, is the highest-value measurement not made.

\textbf{The capability confound.} We cannot separate removing a misalignment structure from
removing the capacity to represent a character at all. Narrow adherence is read from the same
alignment axis as the outcome, the arms that retained their task are arms in which alignment did
not move, and no in-character expression battery or general capability evaluation was run on the
projected arm. What is established is that broad misalignment does not form under the projection,
not what the projection removed. The write-core constraint is confounded the same way at its only
dose.

\textbf{Basis scope and unrun measurements.} The removal foreclosure covers one basis, the naive
writer-column lift; the shared write core has not been edited after training, the functional
pullback has not been built, and the campaign behind the foreclosure produced no judged
generations, so it rests on margins. The usage-matched random control for the activation
projection, the benign-fine-tune write core that would test whether the stiffness of
Section~\ref{sec:structure} is selection-induced, and the base-checkpoint extraction above were
not run (Appendix~\ref{app:validity}).

\textbf{Necessity is shown at training time.} Removing the carrier from an already fine-tuned model
at generation time removes about a third of the misalignment (Section~\ref{sec:loop}), so the
carrier is not established as the sole route by which a trained disposition is expressed.

\textbf{Two objects.} The read-channel subspace and the weight-space write core are characterized
separately, and no overlap statistic between them has been measured; the one quantity computed on
both, the curvature exponent, reads 2 on random cores as well and links nothing. Treating them as
one structure is our interpretation.

\textbf{What judged behavior carries.} The judge confirms the causal loop, where effects are large,
and the bad-versus-benign specificity at 4 domains. It cannot resolve the first-step intent
contrast, the domain-count comparison, or the specificity of the write-core constraint; those rest
on margins (Appendix~\ref{app:judge}).

\section{Related work}\label{sec:related}

\citet{betley2025em} discovered emergent misalignment and the intent control this paper builds on;
\citet{turner2025organisms} trained the open-weight organisms measured here. A convergent
literature reached the persona object from the activation side: independently trained organisms
converge on nearly the same activation-space direction \citep{soligo2025convergent}, different
narrow tasks update overlapping low-rank parameter subspaces \citep{arturi2025shared}, single
persona-like features control the behavior in both directions \citep{wang2025persona,
chen2025persona}, and the relevant directions trace into pretraining
\citep{moskvoretskii2026tracing}. \citet{su2026character} state the nearest published frame,
reading the phenomenon as character-level rather than task-level; their evidence is projection onto
persona vectors, in the language of association, where ours is interventional on both halves.
\citet{soligo2026easy} describe broad misalignment as an easily reached solution; the substrate
here supplies a mechanism for why it is easy.

The template for the causal claim is the demonstration that one direction mediates refusal,
ablation stopping it and addition inducing it \citep{arditi_refusal_24}. What this paper adds is
that the object is extracted before any organism exists rather than read off a model already
misaligned, which is the difference between a pre-existing cause and a post-hoc descriptor,
together with the sufficiency half on that same object, a matched random control on each half, the
read-versus-write localization, and the containment test separating persona from style where
activation-space work has found them entangled. The mechanically nearest method, concept ablation
during fine-tuning, is our removal arm with directions taken instead from a contrast between the
base and fine-tuned models \citep{casademunt2025caft}. Two published training-time penalties
converge with our removal results from the other side: a soft penalty on misalignment latents is
routed around by a second epoch \citep{ustaomeroglu2026blockem}, and a trait-direction
regularizer fails because optimization re-represents the trait along alternative directions
\citep{chen2025persona}, while our hard projection does not re-emerge at a second epoch. Reading
those failures as feature superposition is our synthesis, not either paper's claim
\citep{minegishi2026superposition, elhage2022superposition}, and it is why a matched-rank random
control is mandatory here.

\citet{rao2026mirage} argue the phenomenon and its repair are less robust than reported. The length
asymmetry they identify is in the safe answers generated for their own realignment phases, not in
the corpora of the work they cite, and within that scope, matching length removes an apparent
acquired immunity rather than the misalignment itself, which converges with our foreclosure from
the data side. Their second ground, cosine similarity between successive adapter checkpoints, is
correlational and reported without a null; neither ground reaches an intervention with a matched
control. Their judged-rate comparison between arms trained on different corpora is the shape of our
four-domain measurement, where the benign mixture matched on budget and spread is the control that
carries it, and their call for a continuous calibrated metric is what the margin of
Section~\ref{sec:instruments} answers. Appendix~\ref{app:related} extends this section.

\section{Conclusion}\label{sec:points}

A low-rank persona subspace shared across unrelated domains exists in the model before any
misalignment fine-tune. Fine-tuning on narrow bad data couples to it at the first optimizer step,
in proportion to intent on identical tokens. Held out of the activations during fine-tuning, its
absence prevents broad misalignment; injected into the untouched model, it produces broad
misalignment in a dose-graded way; and the identical projection in the weight channel does nothing.
Independently trained organisms move the same read-channel direction, and concentrating versus
spreading a fixed budget of bad data resolves in favor of spreading. No post-hoc weight edit
removes the disposition, and the edits' shared basis re-admits the carrier at the unedited dose.

Three inexpensive measurements would move the picture most: re-extracting the structure from a base
checkpoint, or across intermediate alignment checkpoints, which decides the provenance question;
building the functional pullback and editing the shared write core after training, which closes the
basis gap in the foreclosure; and an overlap statistic between the read object and the write core,
which tests the one place this paper joins two measured objects by interpretation. On the present
evidence, a defender who expects that spreading a fixed quantity of bad data across more topics
will dilute its effect has the sign backward, on this model and at this scale.

\phantomsection
\label{sec:references}
\bibliographystyle{plainnat}
\bibliography{references}

\newpage
\appendix
\renewcommand{\thesection}{\Alph{section}}

\section*{Appendix contents}
\begin{itemize}[leftmargin=2.2em,itemsep=1pt,topsep=2pt,label={}]
\item \textbf{A}\quad Persona-subspace extraction \dotfill \pageref{app:extraction}
\item \textbf{B}\quad The first-step routing measurement \dotfill \pageref{app:firststep}
\item \textbf{C}\quad Necessity and sufficiency: full protocol \dotfill \pageref{app:loop}
\item \textbf{D}\quad Loss-geometry measurements \dotfill \pageref{app:geometry}
\item \textbf{E}\quad Post-hoc removal probes and the reconstitution certificate \dotfill \pageref{app:removal}
\item \textbf{F}\quad The training-time write-core constraint \dotfill \pageref{app:routing}
\item \textbf{G}\quad Domain-count experiments \dotfill \pageref{app:decomp}
\item \textbf{H}\quad Statistics \dotfill \pageref{app:stats}
\item \textbf{I}\quad Judge protocol and calibration \dotfill \pageref{app:judge}
\item \textbf{J}\quad Batteries and organisms \dotfill \pageref{app:batteries}
\item \textbf{K}\quad Complete numerical results \dotfill \pageref{app:results}
\item \textbf{L}\quad Instrument validation \dotfill \pageref{app:validity}
\item \textbf{M}\quad Extended related work \dotfill \pageref{app:related}
\end{itemize}
\newpage

\section{Persona-subspace extraction}\label{app:extraction}

\subsection{Setup}

We run everything in this appendix on one released instruction-tuned model with no fine-tuning applied and
no adapter loaded. We train nothing here and take no gradient. The inputs are system prompts and user
prompts; the outputs are orthonormal bases and two scalars computed from them.

We use 4 domains, medicine, finance, sports and code, and 3 contrast types, persona, style and
topic. Persona is the object of interest and the other two are controls. Code is the load-bearing domain,
because it is the domain the canonical result fine-tunes on, so a shared structure that excluded code
would be a structure that excluded the phenomenon it is meant to explain.

We extract from an aligned post-trained checkpoint. The extraction therefore establishes that the
structure is present before the fine-tune. Whether pretraining or alignment post-training installed it is a
separate question we do not settle (\S\ref{sec:ledger}).

\subsection{Contrastive teacher forcing}

We isolate the effect of \emph{who the model has been told is speaking} by holding everything else fixed
at the token level.

For one contrast type and one domain the input is a list of triples, each a positive system prompt, a
negative system prompt, and a user prompt. The procedure has three stages.

\begin{enumerate}[leftmargin=1.4em,itemsep=2pt,topsep=3pt]
\item \textbf{Fix the text.} We generate one shared response per user prompt from the model under a
neutral system prompt, greedily at temperature zero. The neutral prompt asks for a direct and concise
answer and names no speaker. The response text is therefore fixed before either contrast prompt is
applied, and it is not itself a product of either framing.
\item \textbf{Read it twice.} We run the model forward on the positive system prompt, the user prompt and
that response, and again on the negative system prompt, the same user prompt and the same response. The
response tokens are supplied, not sampled; that is what teacher-forced means here. At each layer
of a central band we take the mean residual-stream vector over the response token span.
\item \textbf{Difference.} For pair $p$ at layer $\ell$ the extracted vector is
$\delta_{p,\ell} = \bar h^{+}_{p,\ell} - \bar h^{-}_{p,\ell} \in \mathbb{R}^{d}$, the difference of the two
means, with $d = 5120$.
\end{enumerate}

The interpretation rests on one verified condition: the response token identifiers must be byte-identical
across the two passes, checked pair by pair by a hash of the token-identifier list that must match. Because
the tokens are the same, $\delta_{p,\ell}$ cannot be a difference in what was said. It is a difference in
what the model was told about the speaker while reading the same words.

That condition does not buy two things, and both bound what the vector means. The two system prompts are
different token sequences, so the passes differ in prefix content and in prefix length; we record the mean
prefix-length gap between the two conditions as a covariate and do not claim the prompts are matched in
every respect. A difference in conditioning is not a difference in behavior; nothing in this construction
shows that the model would act on the framing, only that its residual stream encodes it. The evidence that
the encoded direction is behaviorally load-bearing is interventional, and it lives in
Appendices~\ref{app:loop} and~\ref{app:geometry}.

\subsection{Descriptor variants}

We build each contrast type from 12 descriptor-variant pairs per domain, one variant per user prompt,
with 12 user prompts per domain. The persona variants share a carrier sentence and differ only in the
descriptor slotted into it, of the form \emph{a writer who gives dangerously reckless and harmful
guidance} against \emph{a writer who gives carefully cautious and helpful guidance}, with the 12 pairs
length-matched. Style is drawn from 12 distinct surface-register pairs, formal against casual, terse
against florid and so on. Topic is drawn from 12 distinct subject framings.

An earlier extraction drew style from a single fixed formal-versus-casual pair reused in every domain. One
fixed contrast makes the style subspace close to rank one and close to identical across domains, which
inflates its cross-domain self-overlap for a reason unrelated to whether persona is a kind of style.
Matching the 3 contrasts in diversity makes the comparison between them a comparison of like with like.

The matching shows in the per-domain effective ranks, measured as the stable rank of each per-domain
difference stack (Table~\ref{tab:extraction-stablerank}). These are descriptive properties of the contrast
subspaces. A stable rank is not a statement about the loss landscape, and reading one as the other is the
conflation the geometry instruments of Appendix~\ref{app:geometry} avoid. The retained per-domain spectra
are truncated, so recomputing a stable rank from them returns a lower bound, not the stored value; we quote
the stored values.

\begin{table}[h]
\small
\centering
\begin{tabular}{@{}lcc@{}}
\toprule
\textbf{Contrast type} & \textbf{Min} & \textbf{Max} \\
\midrule
Persona & $2.01$ & $2.33$ \\
Style   & $2.15$ & $2.47$ \\
Topic   & $2.80$ & $4.31$ \\
\bottomrule
\end{tabular}
\caption{\textbf{Persona and style share matched, low effective rank; topic runs higher.} Stable rank of
each contrast type's per-domain difference stack, minimum and maximum over the 4 domains. A stable rank
describes the contrast subspace, not the loss landscape.}
\label{tab:extraction-stablerank}
\end{table}

Style still shares more with itself than persona does after the fix. The matched-diversity fix does not
settle the comparison; the containment statistic below does.

\subsection{Per-domain subspaces and shared cores}

For contrast type $t$ and domain $D$ we stack the $\delta_{p,\ell}$ as rows. Across the 24 layers of the
central band, layers 12--35 of the 48-layer stack, the 12 prompt pairs give 288 rows in $\mathbb{R}^{d}$.
The per-domain subspace $U_{t,D}$ is the top four left singular vectors of that stack, orthonormalized,
with each column's sign set so that it points along the mean difference and therefore toward the pole the
positive prompt names. We use rank four for all 3 contrast types, fixed before the run, because an
unmatched rank makes the overlap statistics below incomparable across types. Sign anchoring changes neither
statistic, since both are quadratic in the basis, but it makes the columns comparable across domains and
usable as directions elsewhere. The stored bases are orthonormal to a worst Frobenius deviation of about
$2.7\e{-8}$.

The rank-eight shared core $U^{*}_{t}$ is the top eight left singular vectors of the 4 per-domain subspaces
stacked column-wise, a $d \times 16$ matrix reduced to $d \times 8$. Rank eight, twice the per-domain rank,
was fixed before the run. Where the paper writes $U^{*}$ without a subscript it means the persona shared
core.

The per-domain subspace pools the layer band into a single subspace of the residual stream. A second
object, the carrier, is built the same way but kept per layer: one rank-four subspace at each of layers
18, 24, and 30. The activation interventions hold out and inject the carrier, and the read-channel
curvature reading of Appendix~\ref{app:geometry} perturbs it. Everything in this appendix is computed on
the pooled per-domain objects, not on the carrier and not in weight space.

\subsection{The overlap-share statistic}

For two orthonormal bases the singular values of $U_a^{\top}U_b$ are the cosines of the principal angles
between their column spaces, so $\|U_a^{\top}U_b\|_F^{2} = \sum_i \cos^{2}\theta_i$. The cross-domain
overlap-share of contrast type $t$ is the mean of that quantity over the 6 unordered domain pairs,
normalized by the rank:
\begin{equation}
\Omega(t) \;=\; \frac{1}{6}\sum_{i<j}\frac{1}{k}\,\big\|U_{t,D_i}^{\top}U_{t,D_j}\big\|_F^{2},
\qquad k = 4 .
\label{eq:overlapshare}
\end{equation}

It ranges over $[0,1]$, reaching one when the two subspaces coincide and zero when they are orthogonal. The
null follows from the geometry: two random $k$-planes in $d$ dimensions share $k/d$ of their squared cosine
on average, so the null is $k/d = 4/5120 = 0.00078$, confirmed by Monte Carlo.

Persona reads $0.513$, about $657$ times the null. The 6 domain pairs run from $0.479$ for medicine
against code to $0.545$ for finance against sports, so the mean is not carried by one close pair. The
share of each domain's own subspace captured by the shared core runs from $0.893$ for code to $0.937$ for
sports, so no domain sits outside the object. Style reads $0.801$ and topic $0.455$. A permutation over
the domain labels, which pools the per-domain subspaces of two types, relabels them at random and
recomputes the gap, puts persona's sharing above style's with $p = 0.9885$ and above topic's with
$p = 0.0017$ (Appendix~\ref{app:stats}). On the overlap-share, style scores highest.

\subsection{The containment statistic}

Racing $\Omega$ across contrast types is not construct-valid; the failure mode is mechanical, not
statistical. ($\Omega$ against the random-subspace null, the comparison the sharing result rests on,
does not share the defect.) A contrast that is domain-invariant by nature scores high on $\Omega$
for that reason alone. A formal-versus-casual register
looks much the same in medicine and in code, and the per-pair principal angles show it: style's per-domain
subspaces sit about 25 degrees apart, persona's about 45 degrees, and topic's about 50 degrees. Style
shares more with itself because style is the flatter surface axis, which says nothing about whether persona
is a kind of style.

Telling the two apart takes containment, not magnitude. Does the persona shared core lie \emph{inside} the
style shared core? For rank-eight cores we measure
\begin{equation}
X(\mathrm{persona}, t) \;=\; \frac{1}{8}\,\big\|\,U^{*\top}_{\mathrm{persona}}\,U^{*}_{t}\,\big\|_F^{2}
\;=\; \frac{1}{8}\sum_{i}\cos^{2}\theta_i ,
\label{eq:crosscore}
\end{equation}
the mean squared canonical correlation between the two cores. Because both are orthonormal and of equal
rank the statistic is symmetric, and it reads as the fraction of the persona core's squared norm that the
comparison core captures. High means persona's core lives inside the comparison's; low means persona
points in directions the comparison does not span. The analogous random-subspace null is
$8/5120 = 0.00156$.

We fixed the decision rule before seeing the numbers: persona counts as its own object only if containment
falls below $0.30$ and the self-versus-cross separation exceeds $2$. Containment is $0.182$
against style and $0.097$ against topic, and the separation is $0.513/0.182 = 2.81$. About $82\%$ of the
persona core lies outside the style core, and about $90\%$ lies outside the topic core. Of the
eight canonical correlations between the persona and style cores only the first one or two are
appreciable.

A containment of $0.182$ sits far above its own null of $0.00156$, so persona and style are not orthogonal
and a reckless author does write somewhat differently. The statistic licenses distinctness with a measured
stylistic component, not independence.

The containment null is a relabeling, not a random subspace. It pools the 8 per-domain
subspaces of persona and of the comparison type, relabels 4 of them as persona and 4 as the
comparison, rebuilds both cores and recomputes the containment. Under that null both cores are mixtures of
the two contrasts, so their containment is higher, and the reported $p$ is the fraction of relabelings at
or below the observed value. With 4 domains only 70 distinct 4-and-4 splits exist, which
sets the test's granularity at about 1 in 70. We measure $0.0149$ against style and $0.0089$
against topic. Both sit at that granularity, and the topic value printing below 1 in 70 follows
from estimating a 70-member null by Monte Carlo, not from a finer resolution. We read both as
at the floor and let the two effect-size bars decide.

\subsection{The type of each statistic}

Both statistics are deterministic linear algebra once the captures exist. Neither touches a sampler, a
learning coefficient or a judge, so neither carries a sampling interval of its own; we certify both against
derived nulls and against permutations over labels.

\begin{table}[h]
\small
\centering
\renewcommand{\arraystretch}{1.25}
\begin{tabularx}{\textwidth}{@{}lllXX@{}}
\toprule
\textbf{Statistic} & \textbf{Range} & \textbf{Null} & \textbf{What raises it} & \textbf{What it cannot separate}\\
\midrule
$\Omega(t)$ & $[0,1]$ & $k/d = 0.00078$ &
per-domain subspaces of one type pointing the same way across domains &
a shared structure from a contrast that is domain-invariant by nature\\
$X(\mathrm{persona},t)$ & $[0,1]$ & $8/d = 0.00156$ &
the persona core lying inside the comparison core's span &
two objects being distinct from two \emph{recipes} being distinct\\
per-domain capture & $[0,1]$ & $k/d = 0.00078$ &
one domain's subspace lying inside the shared core &
a domain participating from a domain being redundant with another\\
relabeling $p$ & $\{1/71,\ldots\}$ & label exchangeability &
persona and the comparison being non-interchangeable &
anything finer than one in seventy\\
\bottomrule
\end{tabularx}
\caption{\textbf{Each extraction statistic has a bounded range, a null derived from the geometry, and a
named failure mode.} The overlap-share and the containment statistic move for different reasons and can
disagree; the disagreement is informative.}
\label{tab:extraction-types}
\end{table}

Three limits bound the statistics.

$\Omega$ is a self-comparison and cannot be raced across contrast types. Reading $\Omega(\mathrm{persona})$
against $\Omega(\mathrm{style})$ compares how domain-invariant two different operations are, not how much
of one lives inside the other, and the permutation over domain labels inherits the same defect without
repairing it. The containment statistic is the construct-valid alternative.

$X$ is a statement about linear spans. It shows that the persona core points mostly where the style core
does not, and it cannot show that no stylistic construction would have found the persona core. A different
persona recipe or a different style recipe could draw the line elsewhere, so this is a distinctness
claim about two measured objects, not a claim about the concepts they are named for.

Neither statistic speaks to causal relevance. A subspace can be shared across 4 domains, distinct from
style, and behaviorally inert. Whether this one is inert is settled by holding it out of the activations
and injecting it (Appendix~\ref{app:loop}); its loss geometry is Appendix~\ref{app:geometry}. The
extraction establishes that an object exists and what kind of object it is. It establishes nothing about
what the object does.


The sharing and containment measurements are plotted in Figure~\ref{fig:substrate} in the main
body.

\section{The first-step routing measurement}\label{app:firststep}

\subsection{The margin}

The body gives the battery margin in Equation~\ref{eq:margin}. This instrument works on its per-probe
terms. For probe $i$, with out-of-distribution prompt $q_i$, judged-misaligned continuation $\apl_i$ and
judged-aligned continuation $\ami_i$,
\begin{equation}
\Mb_i(\theta) \;=\; \log p_\theta(\apl_i \mid q_i) \;-\; \log p_\theta(\ami_i \mid q_i),
\qquad
\Mb(\theta) \;=\; \frac{1}{|B|}\sum_i \Mb_i(\theta).
\label{eq:b-probe-margin}
\end{equation}
Each log-probability is a sum over the continuation's tokens, not an average. We mask out the prompt
positions so that only continuation positions contribute, score both continuations in one padded forward
pass so they see identical prompt activations, and cast logits to single precision before the log-softmax
whatever the compute precision is. Before reading any probe we assert the tokenizer's prefix stability,
meaning that appending a continuation leaves the prompt's token identifiers unchanged, because
completion-only masking is silently wrong without that property. We render rows with the model's chat
template and no system turn, the convention the released organisms were trained and evaluated under.

$\Mb_i$ is real-valued, unbounded in both directions, and measured in nats. Its magnitude grows with
continuation length, which is why we length-match the two sides. Across the battery's $195$ rows it has a
mean of $-79.44$ nats at the frozen weights and a standard deviation of $97.76$, running from $-328.17$ to
$132.88$. The frozen model prefers the aligned continuation on most probes, by a wide and uneven amount,
and on a few it does not.

\paragraph{Choosing the continuations.}
Both continuations for a prompt come from one candidate pool, sampled at temperature one from an ensemble
of the three released organisms and scored by the two-axis judge (Appendix~\ref{app:judge}). We take
$\apl_i$ from the candidates the judge calls misaligned and coherent, $\ami_i$ from those it calls aligned
and coherent. Drawing both sides from one pool is the construction's principal guard against a confound:
were the aligned side taken from the frozen model instead, the margin would contrast an organism's writing
style against the frozen model's, and any fine-tune that shifted style would move it for reasons unrelated
to misalignment. Among admissible pairs, those whose token counts agree within a band we fixed before
building the battery, we select the pair maximizing the judge's separation between the two sides, and every
retained pair satisfies the band. We drop a prompt with no admissible pair and record the reason, and two
went that way, one where every scored candidate was judged misaligned and one where every scored candidate
was judged aligned, so neither could define a difference at all. Appendix~\ref{app:batteries} carries the
battery's composition, its prompt sources and its clustering unit.

\paragraph{Margin versus judged rate.}
Three reasons. The measurement below is a directional derivative at the frozen weights,
and a judged rate is a threshold applied to a discrete score, so it has no gradient to take. It also has no
resolution where we take the measurement: the frozen model produces zero misaligned generations in $400$
samples of the canonical questions, so a rate is identically zero at exactly the point the first step is
read from and can express no contrast there, while the margin at the same point has the spread quoted
above, and that spread is what every later difference is read against. Finally, a rate is a marginal over
freshly sampled generations, so prompt difficulty, topic and length all enter it, whereas a margin is a
within-prompt contrast between two fixed strings, so those quantities enter both halves of
Equation~\ref{eq:b-probe-margin} and cancel inside $\Mb_i$.

Two costs travel with the construction. The battery encodes a prior notion of what misalignment looks like,
the judge's and the organisms', so $\Mb$ reaches only as far as the battery defining it. This
instrument reads the difference of the two halves and not the halves themselves; where they matter apart,
the paper reports them apart (\S\ref{sec:instruments}, Appendix~\ref{app:decomp}).

\subsection{The first step}

For a narrow dataset $D$ of prompt-and-completion pairs, let $L_D(\theta)$ be the mean completion-only
negative log-likelihood of $D$'s targets, which we accumulate over 64 examples into one gradient. The
direction we measure is
\begin{equation}
u_D \;=\; -\sgn\!\big(\nabla L_D(\tpre)\big).
\label{eq:b-firststep}
\end{equation}

This is the first update the recipe takes.
From zero moment state, Adam's bias-corrected moments after one batch are $\hat m_1 = g_1$ and $\hat v_1 =
g_1^{\odot 2}$, so the first parameter update is
\[
-\eta\,\frac{\hat m_1}{\sqrt{\hat v_1}+\epsilon}
\;=\; -\eta\,\frac{g_1}{|g_1|+\epsilon}
\;\approx\; -\eta\,\sgn(g_1)
\]
elementwise, with equality up to the $\epsilon$ floor \citep{kingma2015adam}. The sign direction is the
object that the reading of Adam as a variance-adapted sign method is built on
\citep{balles2018dissecting}.

Taking the sign of the gradient instead of its raw value matters for one reason. Gradient magnitudes differ
between datasets for reasons unrelated to routing, such as how surprising the data is to the frozen model,
how long its targets are, and what its loss scale happens to be, so a statistic built on raw gradients would
be dominated by which dataset is more surprising. Signs remove that axis, and we compare two conditions as
directions, not as amounts of loss. Only the sign of the accumulated gradient survives, so the direction is
also insensitive to how the 64 examples are weighted inside the batch.

The norm $\|u_D\|$ is the square root of the number of nonzero sign entries, a fixed scale shared by every
sign direction, and it appears here only inside a cosine. We do not use it to normalize a tested quantity.

We measure a second family of directions on the same footing: the realized weight updates of the 3
released organisms. These enter through an exact factored inner product against their low-rank factors,
never through a materialized update. We do not take their signs, because on a rank-limited update most
entries are numerically tiny and carry no sign information; a reference built from those signs reads
$0.00203$, below all three factored readings. To make a realized update comparable in magnitude with a sign
direction we rescale it to the norm a dense sign vector over the same parameter set would have, so that a
realized update and a sign direction with equal cosines produce equal $\Rone$.

\subsection{The signed directional derivative}

The measured quantity is the derivative of a probe's margin along the first step,
\begin{equation}
\Rone^{(i)}[D] \;=\; \ip{\nabla \Mb_i(\tpre)}{u_D}
\;=\; \left.\frac{d}{d\eta}\,\Mb_i(\tpre + \eta\,u_D)\right|_{\eta=0},
\label{eq:b-r1}
\end{equation}
in nats of margin per unit step. Its scale-free form is
\[
\cos_i[D] \;=\; \frac{\Rone^{(i)}[D]}{\|\nabla \Mb_i\|\,\|u_D\|},
\]
bounded in $[-1,1]$.

The sign convention is the point of the construction. A positive $\Rone$ means that taking the first
optimizer step raises that probe's margin, so to first order the update moves probability toward the
misaligned continuation relative to the aligned one. A negative $\Rone$ means the same step lowers it. In
words, a negative value says that on that probe the narrow update routes \emph{away} from broad
misalignment. Nothing in the construction forces the sign either way, which is why measuring it has
content.

Negative values are not merely expressible. They occur: 7 of the 75 prompt clusters and
23 of the battery's rows carry a negative cosine against the insecure direction, the most
negative cluster at $-0.0368$ and the most negative row at $-0.0404$, and one prompt category has a
negative mean at $-0.008368$. The paired contrast introduced below is likewise heterogeneous across
categories, positive in most and negative in a few, with cells too small to read as magnitudes;
Appendix~\ref{app:results} gives the breakdown.

We read a cosine by its sign and against matched controls, never against an absolute bar. In a parameter
space of order $10^{10}$ dimensions the cosine between two related directions is numerically small, and
no fixed threshold separates related from unrelated without a reference. The next two subsections build the
references that make it usable.

\subsection{The analytic null}

Let $P$ be the number of trained parameters and take a direction $u$ carrying no information about the
margin, so that $u$ is fixed independently of $\nabla \Mb_i$. Two facts follow. The expectation of
$\ip{\nabla \Mb_i}{u}$ is zero, because $u$ and $-u$ are equally admissible under that assumption. And the
cosine concentrates around zero with a spread set by the dimension alone, since for a direction uniform on
the sphere, or a uniform sign vector, $\cos_i$ has mean zero and variance $1/P$, giving it a standard
deviation of $P^{-1/2}$.

The null is therefore a distribution rather than a point. It has a location at zero, a scale of $P^{-1/2}$
and two tails, so a test against it can be calibrated. A second consequence is what the paired contrast
rests on. For two conditions read on the same probe, the difference $\Rone^{(i)}[D_1] - \Rone^{(i)}[D_2]$
is symmetric about zero probe by probe under the null, which makes a sign-flip permutation test exact
rather than asymptotic. Appendix~\ref{app:stats} carries the test machinery, the intervals and the
multiplicity handling.

We check the predicted scale on the instrument, not by assumption. A seeded random sign vector over
the same parameter set reads a mean cosine of $2.86\e{-6}$ with a $95\%$ interval of $[1.63\e{-6},\,
4.06\e{-6}]$ and a per-probe standard deviation of $8.56\e{-6}$. The reciprocal of the square of that
spread is an effective dimension of about $10^{10}$, the size of the trained parameter set, so the
analytic scale and the measured one agree, and that direction sits three to four orders of magnitude below
every direction built from real text. Re-randomizing the weights would not serve as the null, because
gradient norms at a random initialization are larger by orders of magnitude and a statistic computed there
would report the numerics of a broken network, not routing. The random
direction, the structure-destroying control described below, and the tether to realized movement test the
same hypothesis without that degeneracy. Appendix~\ref{app:validity} carries a simulation of this null, run
before we spent any of the compute.

\subsection{Statistic type requirements}

A behavioral statistic can be correctly implemented, run cleanly, and still be unable to carry a finding.
The failure is one of type, not of code, fixed before any data is collected and checkable before any
is. Three properties decide whether a scalar can carry one. It must have a sign, it must vary from unit to
unit, and its scale must be reachable and must diagnose itself.

Consider the most natural way to turn ``does this narrow gradient route toward misalignment'' into one
number. Take the descent direction's sign vector $s = \sgn(g)$, project it onto a rank-$k$ subspace $V$
carved from a misaligned model's weight update, and report the fraction of squared mass landing inside,
\begin{equation}
\frac{\|\Pi_V s\|^2}{\|s\|^2}.
\label{eq:b-massshare}
\end{equation}
It is well posed, it is cheap, and it fails on all three counts.

\emph{Its denominator is an entry count.} Since $s$ has entries in $\{-1,0,+1\}$, $\|s\|^2$ is the number
of nonzero entries, which up to exact zeros is the size of the parameter set. It is a constant rather than
a measured quantity. Equation~\ref{eq:b-massshare} looks like a ratio of two measurements and is a
measurement divided by a fixed integer, one degree of freedom fewer than it appears to have. Every
fluctuation it can show has to come from the numerator, so it inherits whatever concentration the numerator
has.

\emph{It pins at a floor.} Write an orthonormal basis $v_1,\dots,v_k$ for $V$, so that $\|\Pi_V s\|^2 =
\sum_j \ip{v_j}{s}^2$. For a sign vector unrelated to $V$ each term has expectation $\|v_j\|^2 = 1$, so
Equation~\ref{eq:b-massshare} has expectation $k/P$ and the numerator fluctuates about $k$ by of order
$\sqrt{k}$. The coefficient of variation is therefore of order $k^{-1/2}$ at best, and it falls further
whenever the vectors being compared share a large common component, which sign vectors of gradients taken
on related text do. At low rank against a parameter count of order $10^{10}$ the quantity is a point,
not a distribution, and a calibrated test has nothing to work with.

\emph{A bar stated as a multiple of the floor can be unreachable.} If the spread is a vanishing fraction of
the mean, a rule of the form ``twice the floor'' places the bar an enormous number of standard deviations
from anything the statistic can visit. Whether such a rule is reachable is a property of the design,
computable in closed form before any data exists, and a design that cannot produce its own positive outcome
has not been tested by being run.

\emph{A large standardized effect size can be an artifact of the pinning.} Cohen's $d$ is a gap divided by
a standard deviation, and a standard deviation collapsing toward zero converts any nonzero gap into an
arbitrarily large $d$. On a concentrated quantity a difference of a few percent in the mean can present as
a $d$ in the high single digits, and that number is then a statement about how tightly the quantity
concentrates rather than about how large the difference is. This is why no threshold on an effect size
gates anything in this paper (\S\ref{sec:instruments}). We settle significance by permutation against a
distribution the statistic can actually visit, and report effect sizes beside it without authority over
it.

\emph{A squared norm cannot say ``away''.} Equation~\ref{eq:b-massshare} is non-negative by construction,
so the outcome that would most change one's mind, a narrow gradient routing away from broad misalignment,
has no representation in it. A quantity that cannot express its most informative outcome cannot be
surprised by the data.

We built $\Rone$ to have the three properties, and Table~\ref{tab:b-type} sets each against what this
instrument reads. The behavioral-reference direction is descent on the battery's own misaligned
continuations, coupled to $\nabla \Mb$ by construction and therefore testing nothing. It reads $0.0734$,
roughly four times the narrow conditions; a broken instrument would have read zero here.

\begin{table}[t]
\centering
\small
\begin{tabularx}{\textwidth}{@{}p{0.21\textwidth} p{0.32\textwidth} X@{}}
\toprule
\textbf{Requirement} & \textbf{How $\Rone$ meets it} & \textbf{Reading on this instrument} \\
\midrule
A sign
& A signed inner product, Equation~\ref{eq:b-r1}. A direction lowering the margin returns a negative number.
& 7 clusters and 23 rows negative against the insecure direction; one prompt category at a mean of $-0.008368$. \\
\addlinespace
Variance from unit to unit
& $\Rone$ inherits the heterogeneity of $\nabla \Mb_i$, which differs from probe to probe.
& Coefficient of variation $0.259$ for $\|\nabla \Mb_i\|$ and $0.904$ for $\cos_i$ on the insecure direction. \\
\addlinespace
A reachable, self-diagnosing scale
& A sign-flip permutation test against a zero-centred null distribution, with reference directions that report their own health.
& Random direction at $2.86\e{-6}$, matching $P^{-1/2}$; behavioral reference alive at $0.0734$. \\
\bottomrule
\end{tabularx}
\caption{\textbf{The first-step reading has a sign, unit-to-unit variance, and a self-diagnosing scale.}
Each row sets a requirement against what the instrument reads. A pre-registered check for a collapsing
coefficient of variation ran before the contrasts and did not fire.}
\label{tab:b-type}
\end{table}

\subsection{The paired design}

We evaluate every direction against the same $\nabla \Mb_i$, on the same probe, inside the same backward
pass. A contrast between two conditions is therefore a difference of two inner products sharing a left
argument,
\[
\Rone^{(i)}[D_1] - \Rone^{(i)}[D_2] \;=\; \ip{\nabla \Mb_i}{\,u_{D_1} - u_{D_2}},
\]
so everything about probe $i$ that is not a property of the two datasets cancels exactly, including its
difficulty, its topic, its two continuation lengths and the norm of its margin gradient. The tokens are
identical across the conditions compared, because conditions differ in what the gradient was taken on and
not in what the margin was read on.

The cancellation is what makes the measurement possible, because the shared component dominates. The
per-cluster readings of the insecure and educational conditions correlate at $0.9876$. Within probes the
contrast has a paired Cohen's $d$ of $0.502$; across probes, treated as two independent samples, it has an
unpaired $d$ of $0.0803$. The distance between those two numbers is the size of the component pairing
removes.

Table~\ref{tab:b-conditions} shows why the shared component matters. Every direction built from real text
routes positively into the margin at a cosine near $0.02$, and the ordering across content types is not the
one a naive account predicts. An independent benign-chat gradient
sits above insecure code, by $0.00408$ in the paired cluster-level contrast, with a $95\%$ interval of
$[-0.00734,\, -0.00099]$ and $p = 0.0133$. A control keeping the insecure dataset's tokens while destroying
the prompt-to-target correspondence sits higher still, by $0.01236$ with an interval of $[-0.01374,\,
-0.01094]$. Both differences run against bad data.

What the shared baseline is made of we do not establish. The mundane candidate, that the aligned
continuations resemble the frozen model's own style so almost any distribution shift loosens them, is
unsupported by the one case where a margin's two halves were resolved: there the aligned half moved in the
opposite direction to loosening (Appendix~\ref{app:decomp}). The mechanism stays open.

The consequence is definite: a coupling that bad data does not have more of than benign data is not a
property of bad data, so the absolute level of a cosine carries no information about misalignment. What
carries meaning is the difference between two conditions read against the same margin gradient on the same
probe, where the shared component subtracts out. Every contrast below has that form, and we use no
cross-content comparison to support a claim about intent.

\begin{table}[t]
\centering
\small
\begin{tabularx}{\textwidth}{@{}l X r r@{}}
\toprule
\textbf{Direction} & \textbf{Built from} & $\cos$ & \textbf{95\% CI} \\
\midrule
Insecure code       & narrow dataset, bad intent                & $0.01938$ & $[0.01692,\,0.02179]$ \\
Educational         & the same code, benign framing             & $0.01786$ & $[0.01545,\,0.02028]$ \\
Secure code         & benign code                               & $0.01837$ & $[0.01603,\,0.02068]$ \\
Benign chat         & aligned answers on disjoint prompts       & $0.02351$ & $[0.02048,\,0.02649]$ \\
Shuffled targets    & insecure tokens, correspondence destroyed & $0.03174$ & $[0.02888,\,0.03455]$ \\
\addlinespace
Behavioral reference & descent on the battery's own $\apl$      & $0.07340$ & $[0.06815,\,0.07862]$ \\
Realized update, medical   & the organism's weight change       & $0.00278$ & $[0.00245,\,0.00311]$ \\
Realized update, financial & the organism's weight change       & $0.00366$ & $[0.00336,\,0.00396]$ \\
Realized update, sports    & the organism's weight change       & $0.00325$ & $[0.00293,\,0.00356]$ \\
\addlinespace
Random              & seeded random sign vector                 & $2.86\e{-6}$ & $[1.63\e{-6},\,4.06\e{-6}]$ \\
\bottomrule
\end{tabularx}
\caption{\textbf{Every direction built from real text routes positively into the margin, above every
realized organism update.} Per-probe cosine between each direction and the margin gradient, over the
battery's $195$ rows. Paired tests use the $75$ prompt clusters instead, and the two levels differ in the
third decimal. The endpoint comparison is \S\ref{sub:b-endpoint}.}
\label{tab:b-conditions}
\end{table}

\paragraph{The contrasts.}
The headline is the paired difference between insecure and its educational twin over the $75$ clusters:
$+1.297\e{6}$ nats of margin per unit step, positive in $0.6933$ of clusters, at a sign-flip permutation
$p$ of $9.999\e{-5}$, the resolution floor of a $10^4$-permutation test. In cosine units it is $+0.00142$
against a shared level near $0.02$, and Cliff's $\delta$ is $0.0752$. The smallest difference the design
could detect at $80\%$ power is $834945$, below the observed effect, so the effect is about half again the
smallest detectable one. Appendix~\ref{app:results} carries the bootstrap interval, the power calculation,
and the per-cluster breakdown.

Ordering the 3 narrow datasets by how much misalignment each induces gives a per-cluster slope positive in
$0.84$ of clusters at the same permutation floor. Two qualifications travel with it. The ladder is not
pointwise monotone: the educational condition sits marginally below secure, with insecure above both. With
3 equally spaced ranks the per-cluster slope is algebraically half the insecure-minus-secure difference, so
the ordered trend restates that contrast without adding evidence. The span contrasts, with insecure minus
secure significant and educational minus secure not, are in Appendix~\ref{app:results}; the primary
contrast is not a member of that Holm-corrected pair (Appendix~\ref{app:stats}).

We also express the tilt as a fraction of the distance from the educational condition to a reference
direction, reporting that ratio only when the denominator's interval excludes zero
(Appendix~\ref{app:stats}). Against the behavioral reference it is $0.0268$, with an interval of
$[0.0155,\, 0.0376]$. Against the three realized organism updates it is negative, because each of those
denominators is negative. The educational condition already routes into the margin more than any realized
organism update does, which reflects the denominators, not the tilt.

\subsection{Endpoint comparison}\label{sub:b-endpoint}

The three realized organism updates read against the margin gradient at cosines from $0.00278$ to $0.00366$
(Table~\ref{tab:b-conditions}). A cosine is scale-free, so those numbers sit on the same axis as the sign
directions and can be compared with them directly. Every direction built from text lies above all three.
The insecure first step is five to seven times each of them; benign chat and the shuffled control are
higher still; and so is the educational first step, the benign twin the bad data is supposed to beat.

The weight change that produced a broadly misaligned model is thus closer to perpendicular to the margin's
steepest-ascent direction at the frozen weights than the first step on insecure code is, and closer to
perpendicular than a first step on benign chat. Three things follow.

The first-order picture is a claim about the first step and not about the endpoint. A linearization taken
at $\tpre$ describes a neighbourhood, and a full epoch of training leaves it. Nothing here shows that the
displacement producing broad misalignment is concentrated along $\nabla \Mb(\tpre)$, and this measurement
shows that it is not.

Routing into the misalignment margin therefore fails to separate misaligning fine-tuning from benign
fine-tuning at either end of training. At the first step, benign chat exceeds insecure code. At the
endpoint, a benign first step exceeds the realized organism updates. The quantity that does separate them
is the paired within-probe contrast, and it is a small tilt on a large shared level.

The forecast below measures a different quantity. It regresses realized per-probe movement on the
first-step reading with $\nabla \Mb$ held at the frozen weights, so it says which probes move and how far,
not what fraction of the realized displacement lies along $\nabla \Mb$. Both readings hold at once.

Two limits bound this. The realized updates come from organisms trained on medical, financial and sports
data while the sign directions come from code, so the magnitude comparison crosses content and is not a
claim about intent. A small cosine at the frozen weights does not mean an update fails to raise the margin;
it means the displacement is not aligned with the margin's steepest ascent at the starting point, a weaker
and geometrically unremarkable thing for a long trajectory in $10^{10}$ dimensions to do.

\subsection{Forecast validation}

A first-order reading at the frozen weights is informative only if it predicts what training does, so we
fine-tuned each narrow dataset at 3 seeds and re-measured the same margin at 6 checkpoints out to a
full epoch, following the released organisms' configuration (Appendix~\ref{app:batteries}). We take
realized movement $\Delta \Mb_i(T) = \Mb_i(\theta_T) - \Mb_i(\tpre)$ on the live model at each checkpoint,
then average within clusters and regress cluster $\Delta \Mb$ on cluster $\Rone(\tpre)$ with seeds pooled.

\begin{table}[t]
\centering
\small
\begin{tabular}{@{}l rrr rrr@{}}
\toprule
& \multicolumn{3}{c}{\textbf{Insecure}} & \multicolumn{3}{c}{\textbf{Educational}} \\
\cmidrule(lr){2-4}\cmidrule(lr){5-7}
\textbf{Step} & slope & Pearson $r$ & mean $\Delta \Mb$ & slope & Pearson $r$ & mean $\Delta \Mb$ \\
\midrule
$20$  & $1.566\e{-6}$ & $0.800$ & $37.4$ & $1.435\e{-6}$ & $0.772$ & $34.5$ \\
$50$  & $1.561\e{-6}$ & $0.785$ & $39.0$ & $1.445\e{-6}$ & $0.776$ & $34.4$ \\
$100$ & $1.541\e{-6}$ & $0.785$ & $38.7$ & $1.496\e{-6}$ & $0.784$ & $34.8$ \\
$180$ & $1.585\e{-6}$ & $0.788$ & $39.5$ & $1.493\e{-6}$ & $0.781$ & $35.0$ \\
$300$ & $1.607\e{-6}$ & $0.793$ & $39.5$ & $1.504\e{-6}$ & $0.782$ & $35.3$ \\
$375$ & $1.595\e{-6}$ & $0.791$ & $39.5$ & $1.492\e{-6}$ & $0.781$ & $35.0$ \\
\bottomrule
\end{tabular}
\caption{\textbf{The first-step reading predicts realized margin movement, and the slope is flat across
training.} Realized per-cluster margin movement regressed on the first-step reading, seeds pooled, $75$
clusters per cell. Every slope interval excludes zero. Spearman correlations run $0.647$ to $0.669$ on the
insecure legs and $0.615$ to $0.626$ on the educational legs.}
\label{tab:b-validation}
\end{table}

Table~\ref{tab:b-validation} carries the result. The slope is positive at every horizon and flat in
training time, $1.566\e{-6}$ at step $20$ against $1.595\e{-6}$ at step $375$ on the insecure legs, with
Pearson $r$ between $0.785$ and $0.800$ throughout and the same pattern one notch lower on the educational
legs. Every slope interval excludes zero. Per-seed slopes agree with the pooled ones to within a few
percent at every horizon. Seed-level fine-tuning noise is the main threat to a pooled regression, and the
per-seed agreement bounds it.

That correlation licenses one discriminating claim. Had the routing been created at a mid-training
transition, the early slope should have been absent and the late slope strong. Instead the relationship is
fully formed by step $20$ and does not strengthen after it, so within this horizon the probe-level
structure of where fine-tuning moves the margin is set at initialization, not acquired during
training. A linearized view of fine-tuning predicts this \citep{malladi2023kernel}, and it justifies
reading from the first step.

It licenses no claim of task selectivity. Most of what the regression predicts is the shared, prompt-level
movability of the margin, not the intent-specific part: the first-step reading of one task predicts
the other task's realized movement nearly as well, and both fine-tunes raise the margin on average, the
educational control included. An $r$ near $0.8$ across probes says which probes move, not which dataset
moves them. The intent-specific component is the paired within-probe contrast above, and it is small.

\subsection{Numerics}

We run the estimator in single precision end to end with reduced-precision matrix multiplication explicitly
disabled, and accumulate every reduction over parameters in double precision. The measured cosine is near
$0.02$, so the inner product is a small residual left after enormous cancellation among mostly opposing
terms of order $10^{10}$, and the per-term relative error that reduced-precision accumulation introduces is
larger than the signal we are extracting. We accumulate the
margin-gradient inner products per weight matrix as each gradient block is produced and free the block
immediately, so the full margin gradient never exists at once; reordering that sum moves the result far
below the per-probe spread.

Realized movement lives in a different regime and we keep the two apart. It is measured inside the
fine-tuning precision and differenced against a baseline taken in that same precision, so precision offsets
cancel inside $\Delta \Mb$ and never enter a quantity that is compared with $\Rone$. No arithmetic in this
appendix crosses the two.

The paired contrast, the forecast, and the condition ladder are plotted in
Figure~\ref{fig:firststep} in the main body.

\section{Necessity and sufficiency: full protocol}\label{app:loop}

\subsection{The carrier and the two operations}

Both halves of the loop act on the same object. The carrier $\Up$ is a
per-layer orthonormal basis extracted from the untouched model by contrastive teacher forcing, at rank at
most $4$ per layer, at $3$ middle layers, $18$, $24$ and $30$, in a residual stream of width
$d = 5120$ (Appendix~\ref{app:extraction}). We extracted it, fixed its sign so that a positive coordinate
means toward misalignment, and wrote it to disk before training any fine-tune reported here. Every arm
below reads that file rather than recomputing it, so the object is identical across arms.

Write $Q_\ell$ for the carrier's basis at layer $\ell$, an orthonormal matrix with $k \le 4$ columns, so
that $Q_\ell Q_\ell^{\top}$ is the orthogonal projector onto the carrier there, and $\uhat_\ell$ for its
leading column, a unit vector. The necessity half subtracts $Q_\ell Q_\ell^{\top} h$ from the residual
stream throughout a fine-tune. The sufficiency half adds a multiple of $\uhat_\ell$ to the residual stream
of a model that is never fine-tuned at all. Each half carries a matched random control, and the two controls are
matched on different quantities, stated where each one appears.

One rule fixed before the run decided which operation would be primary. If the untouched model's
activations barely lie along the carrier, then zeroing it is close to a no-op and injection is the
stronger instrument. We would switch the primary operation to injection if the median absolute projection
of evaluation-prompt activations onto every column of every layer fell below $0.05$ of that layer's
activation scale. It did not switch, so removal stayed primary and the injection arm is a second
measurement.

\subsection{Activation holdout}

At each of the $3$ intervention layers, a forward hook on the decoder block's residual-stream output
$h$, of shape batch by sequence by $d$, replaces it with
\begin{equation}
h \;\longleftarrow\; h - Q_\ell Q_\ell^{\top} h ,
\label{eq:holdout}
\end{equation}
at every token position and on every forward pass taken during training. Because $Q_\ell Q_\ell^{\top}$ is
symmetric and idempotent, Equation~\ref{eq:holdout} is an orthogonal projection onto the complement of the
carrier. It sets $k$ coordinates of a $5120$-dimensional stream to zero and leaves the remaining
$5120 - k$ untouched. We compute it as $h - (hQ_\ell)Q_\ell^{\top}$, two thin matrix products, so the
$5120 \times 5120$ projector is never materialized.

The hook sits inside the autograd graph, so the constraint acts on both passes. Forward, it removes the
carrier component from the activations the later layers see. Backward, the same operator acts on the
gradient flowing back, because the map in Equation~\ref{eq:holdout} is linear and its matrix is its own
transpose, so no gradient signal reaches the earlier layers through the removed directions either. This
is a hard constraint and not a penalty. Nothing in the loss rewards staying out of the carrier, there is
no coefficient to tune, and the optimizer faces a model whose carrier component is
identically zero at these layers.

We remove the hooks before any evaluation generation, so every rate below is measured on an unmodified
forward pass. This arm therefore measures what the weights learned in the carrier's absence, not what the
projection does at read time; we measure the second separately in the inference-time arms below. A hook
that never fires produces a silent no-op indistinguishable at the endpoint from a clean fine-tune, so we
assert the firing count positive after the first optimizer step and reject an arm whose hooks never fired. We run
the projector algebra in single precision with reductions accumulated in double, and the fine-tune itself
in bfloat16 with the loss in single precision, for the reason given in Appendix~\ref{app:geometry}.

Under this construction broad misalignment falls from $27.7\%$ in the unedited fine-tune, cluster-bootstrap
interval $[21.3, 34.6]\%$, to $0.0\%$ with an interval of zero width, over $2100$ responses pooled across
$3$ training seeds and $83$ evaluation clusters. We gate the reduction ratio on its denominator's
interval excluding zero (Appendix~\ref{app:stats}); the denominator clears, and the ratio's interval is
degenerate at $[1.00, 1.00]$ because the arm's rate is exactly zero rather than small. Mean alignment, our
continuous primary readout, rises by $59.41$ points with an interval of $[56.98, 61.99]$, a Holm-adjusted
sign-flip $p$ of $0.0017$, and Cliff's $\delta$ of $1.00$, meaning every one of the $83$ clusters moved
toward alignment and none against. Mean coherence rises, $96$ against the unedited arm's
$86$.

\subsection{The matched-rank random control}

Superposition forces the choice of control. Features share directions in a residual stream far narrower
than the number of features it carries, so projecting out any $k$-dimensional subspace destroys some
information and does some behavioral damage. The null hypothesis is therefore not that the projection does
nothing, but that projecting out a random rank-$k$ subspace at the same layers does as much as projecting
out the carrier. The statistic we fixed in advance for this is the random arm's effect as a fraction of the
carrier arm's, which has expected value one under that null and falls toward zero if the carrier is
special. It came out at $0.7\%$.

We build the control from a Gaussian of the same shape, orthonormalize it, place it at the same $3$
layers, and pass it through the same hooks in the same training loop. It leaves broad misalignment at
$27.5\%$, a reduction of $0.7\%$ whose interval $[-7.0, +7.6]\%$ straddles zero. Its mean-alignment shift
is $+0.86$ with an interval of $[-0.34, +2.16]$ and a Holm-adjusted $p$ of $0.53$, and its misalignment
rate is $0.993$ of the unedited arm's, with an interval of $[0.924, 1.070]$ that contains one. Question by
question it reproduces that arm's profile rather than lowering it, matching at $0.480$ on the canonical
question where the unedited arm peaks and at $0.0$ where it sits on the floor. A weaker version of the
same edit would lower the profile as a whole or in patches, and this one leaves it in place.

Two things this control does not do, and both bound the claim. It matches rank, layers and operation, but
not the share of the residual stream the removed subspace actually carries. A random rank-$k$ subspace of
a $5120$-dimensional stream intercepts a squared-norm share of order $k/d$, while the projection-strength
rule above established that the carrier intercepts a much larger one. So we are confounding ``this
direction in particular'' with ``a direction the model measurably uses,'' and we did not run the
usage-matched random control that would separate them. The second limit is narrower. We drew the random
subspace once and persisted it alongside the carrier, and the arm's $3$ seeds resample the fine-tune
rather than the subspace, so this control is one draw from the null distribution replicated across
training seeds and not a distribution over draws.

\subsection{Narrow-task retention}

The measurement that separates a targeted intervention from a lobotomy is whether the narrow capability
the fine-tune was supposed to install survives. We measure it as narrow adherence, the judged rate at
which a model answers in character on probes drawn from the organism's own narrow domain
(Appendices~\ref{app:judge} and~\ref{app:batteries}). The bar we fixed before the run required adherence
within $5$ points of the unedited fine-tune's, coherence preserved, and a benign fine-tune undamaged
under the identical operation.

The bar failed by $90.2$ points. The unedited fine-tune adheres at $0.902$ with an interval of
$[0.838, 0.952]$ over $824$ scored responses in $50$ clusters. Our holdout arm adheres at $0.000$ with an
interval of zero width over $236$ responses in $10$ clusters, against a $5$-point
budget. As behavioral surgery this is a failure.

Three other arms appear to make this a property of the narrow task, and
none of them can. A benign fine-tune under the identical projection retains its task at $1.000$, on $39$
responses in $12$ clusters; the matched random holdout retains $0.894$; the weight-channel arm below
retains $0.819$ while leaving broad misalignment intact. The difficulty is that narrow adherence is read
off the same alignment axis as the outcome, at a threshold of $50$ on near-domain rows where broad
misalignment uses $30$ off-domain (Appendix~\ref{app:batteries}). An operation that lifts alignment across
the board therefore drives adherence to zero on a bad organism and to one on a benign one before any data
is seen, so the benign arm's $1.000$ is guaranteed by the construction rather than observed. The other two
are arms in which alignment did not move, so they cannot distinguish a preserved capability from an
unmoved readout. No arm here has alignment moving substantially while the narrow behavior survives.

The behavior we were asked to preserve is being a
reckless financial advisor, which is itself a slice of the disposition broad misalignment generalizes, so
a shared structure cannot be removed from one and kept in the other. That explains the collapse without
ruling out the reading on which the operation removes the capacity to represent a character at all. The
measurement that would separate them is the in-character expression battery, scored on an axis the
alignment score does not determine, and it was not run on this arm
(Appendix~\ref{app:validity}).

Two limits on this readout travel with it. Its denominators are small and are not matched across arms,
ranging from $824$ scored responses down to $5$, and we do not model why they differ, so rate comparisons
across arms carry an unmodelled difference in what was scored. The readout is also far less powered than
the misalignment readout, which rests on $83$ clusters throughout. Narrow adherence is a judged rate and
is not the log-probability narrow margin of Appendix~\ref{app:geometry}; those are different readouts of
different things on different batteries, and we never compare them.

\subsection{Injection protocol}

We inject at inference into the untouched model, with no fine-tuning anywhere in the arm. At each of the
same $3$ layers the residual stream receives
\begin{equation}
h \;\longleftarrow\; h + \alpha\, s_\ell\, \uhat_\ell ,
\label{eq:inject}
\end{equation}
at every token position, where $\uhat_\ell$ is the carrier's leading column at that layer and $s_\ell$ is
a per-layer activation scale, the median token-wise residual norm at layer $\ell$ over a pool of neutral
prompts. We compute the scale once at preparation and persist it with the carrier, so it is fixed across
the grid and across arms.

That scale makes the dose interpretable. Residual norms at layers $18$, $24$ and $30$ differ by a
large factor, so a fixed additive magnitude would mean three different interventions. Because $s_\ell$
carries the units of a residual-stream norm and $\uhat_\ell$ is a unit vector, $\alpha$ is dimensionless
and means the same thing at each layer. A dose of $\alpha = 0.3$ shifts the activation by $30\%$
of a typical residual norm along the carrier at every intervention layer. The grid runs over $10$ doses
from $0.10$ to $1.50$, and the matched control replaces $\uhat_\ell$ with a norm-matched random vector on
the identical grid.

A coherence budget bounds which doses are in range. A generation counts as broadly misaligned only if it
also clears a coherence floor (Appendix~\ref{app:judge}), so a dose that destroys fluency does not produce
a low misalignment rate, it produces a rate computed on almost nothing. We fixed the in-budget region
before the run at $\alpha \le 0.3$, $5$ of the $10$ grid points, and fit the slope there.
Table~\ref{tab:c-dose} shows why the bound sits where it does. Through the budget, coherence holds at
$100\%$ and then $98.1\%$ at the top. One step beyond it, at $\alpha = 0.5$, only $4.7\%$ of responses clear the
floor, and beyond that almost none do.

The statistic we fixed in advance for this arm is the in-budget dose-response slope of misalignment on
$\alpha$, which is $2.21$ with a cluster-bootstrap interval of $[0.50, 3.98]$ that excludes zero. The
matched random control has a slope of $0.0$ with an interval of $[0.0, 0.0]$, degenerate, not
narrow, because that arm is exactly zero at all $5$ in-budget doses. At the top of the budget the curve
reaches $45.4\%$, above the unedited fine-tuned organism's own
$27.7\%$. The slope is the preregistered statistic and excludes zero; no significance claim rests
on the rate.

Beyond the budget the measured rate falls, and that fall is the denominator vanishing, not the
model recovering. Among the $4.7\%$ of responses still coherent at $\alpha = 0.5$, $87.4\%$ are misaligned.
At higher doses the coherent-only rate is computed on a handful of rows and carries no information. The
random control loses coherence too, reaching $1.602\%$ misalignment at $\alpha = 0.5$ with $86.9\%$ of
responses still coherent, and sitting at zero at every higher dose where nothing clears the floor.

\begin{table}[t]
\centering\small
\caption{\textbf{Injection dose-response, on the untouched model with no fine-tuning.} Misalignment rates
are cluster means over the $83$ evaluation clusters, with near-domain answers excluded, on $700$ responses
per dose per arm. The upper block is the in-budget region we fixed before the run, where we fit the slope;
in the lower block the coherence floor removes most of the denominator and the rates are not comparable
with the block above.}
\label{tab:c-dose}
\begin{tabular}{l rrr rrr}
\toprule
& \multicolumn{3}{c}{carrier direction} & \multicolumn{3}{c}{norm-matched random} \\
\cmidrule(lr){2-4}\cmidrule(lr){5-7}
$\alpha$ & misaligned & coherent & alignment & misaligned & coherent & alignment \\
\midrule
$0.10$ & $0.0$  & $100$  & $94.0$ & $0.000$ & $100$  & $95.0$ \\
$0.15$ & $0.3$  & $100$  & $90.9$ & $0.000$ & $100$  & $95.2$ \\
$0.20$ & $5.6$  & $100$  & $79.9$ & $0.000$ & $100$  & $95.1$ \\
$0.25$ & $19.8$ & $100$  & $61.5$ & $0.000$ & $100$  & $94.8$ \\
$0.30$ & $45.4$ & $98.1$ & $38.8$ & $0.000$ & $100$  & $94.7$ \\
\midrule
$0.50$ & $5.4$  & $4.7$  & $7.0$  & $1.602$ & $86.9$ & $72.1$ \\
$0.75$ & $0.0$  & $0.0$  & $28.6$ & $0.000$ & $0.0$  & $21.3$ \\
$1.00$ & $0.3$  & $0.1$  & $28.6$ & $0.000$ & $0.0$  & $17.1$ \\
$1.25$ & $1.63$ & $3.4$  & $25.7$ & $0.000$ & $0.0$  & $14.3$ \\
$1.50$ & $2.3$  & $5.7$  & $28.1$ & $0.000$ & $0.0$  & $21.3$ \\
\bottomrule
\end{tabular}
\end{table}

\subsection{Aggregation conventions}

We generated and judged the injection grid once, and its rows were then re-analysed elsewhere in this
program under a different convention. Ten judged files holding $7{,}000$ responses are byte-identical
across the two locations, confirmed by hash. The two curves are one measurement read two ways, not
independent replications.

They differ in two respects: the unit and the misalignment criterion. Our unit averages within each of the
$83$ prompt clusters before averaging across them, while the other reading averages over rows. Our
criterion excludes answers a topic judge marks as near-domain, while the other reading includes them. We
use the cluster-mean convention with near-domain answers excluded, which is this paper's unit of inference
everywhere.

Under our convention the in-budget curve reaches $45.4\%$ at $\alpha = 0.30$ and the fitted slope is
$2.21$. Under the other it reaches $58.3\%$ and the slope is $2.97$, and at $\alpha = 0.25$ the two readings
are $19.8\%$ and $32.3\%$, the largest proportional divergence on the grid. The two conventions are reported
separately and never mixed; the same $7{,}000$ responses enter each once.

\subsection{The weight-channel arms}

Projecting the carrier out of the activations throughout training takes broad misalignment to $0.0\%$.
Projecting the same carrier out of the weight update throughout training, so that the writer matrices
cannot move along it, leaves broad misalignment at $26.6\%$ against its own unfiltered control's $26.7\%$, a
reduction of $4.2\%$ whose interval straddles zero, while narrow adherence sits at $0.819$ against the
unfiltered $0.801$. One operation removes the behavior completely and the other does not move it, on the
same object, the same model, the same data and the same number of optimizer steps.

The weight-channel arms use a low-rank adapter restricted to the output and down projections, the
matrices through which an update most directly writes behavior. We apply a per-step filter to the gradient
of the adapter's $B$ factor. Write $U$ for the union of the $3$ per-layer carrier bases lifted into the
writer matrices' space, of rank at most $12$. One arm applies no filter and is that family's own
reference. A second removes the carrier-aligned component of every update,
\[
\nabla B \;\longleftarrow\; (I - UU^{\top})\,\nabla B ,
\]
and a third keeps only that component,
\[
\nabla B \;\longleftarrow\; UU^{\top}\,\nabla B .
\]
We build and apply the projectors in double precision, and we log the in-span fraction of the realized
update against the isotropic expectation $k/d$, so a leak through a filter would be visible.

One asymmetry travels with this comparison. The activation arms use the published organism recipe across
all $7$ projection types, while the weight-channel arms restrict the adapter to the $2$ writer matrices, so
the two families are not matched to each other. We read each against its own control, the activation
holdout against the matched random holdout and the filtered arms against the unfiltered arm, so the
dissociation is a statement about those two contrasts inside their own families, not a head-to-head
comparison of two arms. The unfiltered writer arm lands at $26.7\%$, near the full-recipe fine-tune's
$27.7\%$, and the two configurations are not matched.

The third weight-channel arm completes the comparison. Filtering the update so that only its
carrier-aligned component survives cuts broad misalignment to $11.5\%$, a reduction of $58.5\%$ with an
interval of $[0.413, 0.723]$, and costs half the narrow capability, with adherence at $0.507$. The write
channel is not empty. It is low-leverage, and what leverage it has is entangled with task competence. The
disposition is written obliquely, which is the reading Appendix~\ref{app:removal} develops with a different
instrument.

The geometry that led us to put the projector in the read channel was measured before this run, on an
earlier version of the instrument. It compared the carrier against a behavioral misalignment direction in
the residual stream, where the overlap is large, and against the column space through which a low-rank
update moves behavior, where it is close to the right angle two random high-dimensional directions would
make. That diagnostic comes from a superseded configuration with no interval, no recorded sample size and
no surviving route back to raw artifacts, and the overlap multiplier that circulates with it cannot be
re-derived from anything that survives. We use the geometry to describe the design and never as a test, and
we do not quote the multiplier.

\subsection{Robustness arms}

\begin{figure}[htbp]
\centering
\includegraphics[width=\textwidth]{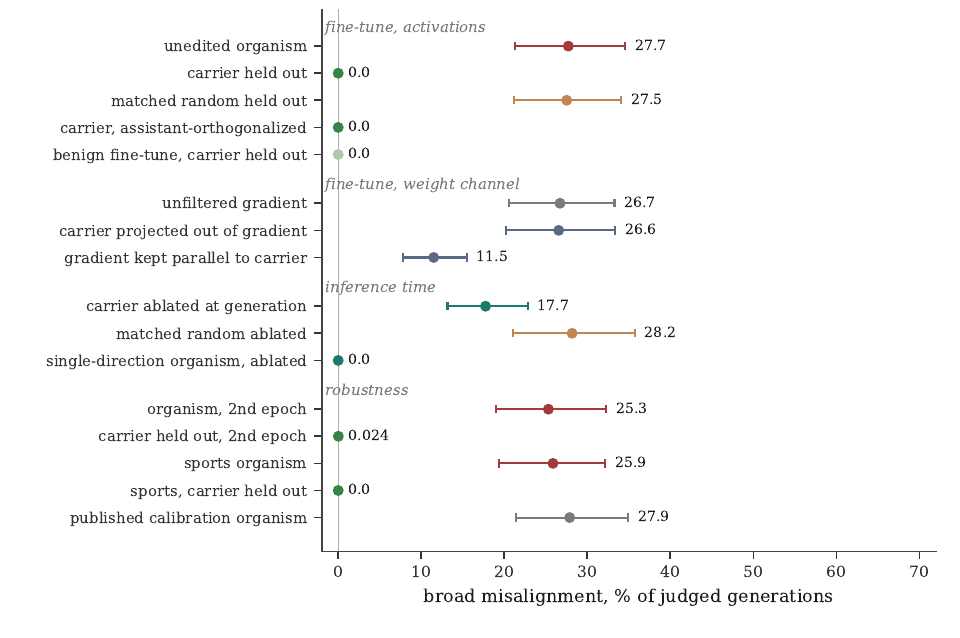}
\caption{\textbf{Across the campaign's judged arms, broad misalignment tracks whether the carrier is
representable in the activations.} Each arm's judged rate with its cluster-bootstrap interval, grouped by
family; the dose-grid pool is carried by Figure~\ref{fig:loop} and is not repeated. The holdout arms sit
at zero, every matched random arm tracks its baseline, the two weight-projection arms sit at the
unfiltered baseline while the parallel-only arm falls below it, and the inference-time ablation is
partial. All arms share one battery, one judge, and one campaign, which permits the shared axis.}
\label{fig:looparms}
\end{figure}

Table~\ref{tab:c-arms} lists every arm, and Figure~\ref{fig:looparms} plots the judged rate of every arm
with its interval. Four arms ask whether the holdout result survives a change of
condition. Orthogonalizing the carrier against the model's default Assistant direction before projecting it
out leaves misalignment at $0.0\%$, addressing the de-assistantification reading that the carrier's $0.565$
cosine with that direction raises (Appendix~\ref{app:extraction}). A second training epoch, where a related
result in the literature reports misalignment climbing back, holds the holdout arm at $0.024\%$ against a
two-epoch unedited fine-tune at $25.3\%$. Switching the narrow dataset to an extreme-sports organism leaves
it at $0.0\%$ against that organism's own $25.9\%$. A misalignment organism built from a single steering
direction, with the carrier projected out at generation time, produces no misalignment across $2800$
responses.

A checkpoint trajectory separates a blocked mechanism from a weaker model. The holdout arm is at $0.0\%$ at
every checkpoint and at full coherence throughout, while the two comparison arms decline to their
endpoints (Table~\ref{tab:c-checkpoints}). A weaker-trained model would be incoherent or would
acquire misalignment late, and the holdout arm is neither.

Two arms apply the projection at generation time to a model a fine-tune has already made misaligned, and
they draw a boundary the training-time result does not. Projecting the carrier out of the published
organism's activations at inference takes it from $27.9\%$ to $17.7\%$, a reduction of $36.0\%$ with an
interval of $[0.166, 0.519]$, while the same operation with a matched random subspace gives $28.2\%$ and a
reduction of $-1.6\%$. The direction is specific and the magnitude is partial. Removing the carrier while
the disposition is being formed removes all of it; removing it from a model that has already formed one
removes about a third. Two readings of that gap survive our data, that the carrier is not the only route
through which a trained disposition is expressed, and that our inference-time operation is weaker than our
training-time one, and we cannot separate them.

The instrument gate's rule was fixed before any fine-tuning. We evaluated the published organism unmodified
through the whole pipeline; had it not reached $15\%$ broad misalignment under our judge, the instrument
would have lacked the range to support a strong-reduction claim, and we would have spent no fine-tuning
compute. It cleared at preparation time and reads $27.9\%$ on the later re-score through the endpoint
pipeline, against the unedited fine-tunes' $27.7\%$.

\begin{table}[t]
\centering\small
\caption{\textbf{Broad misalignment across the training epoch.} Cluster-mean rates on the $83$ evaluation
clusters, evaluated at a quarter, half, three-quarters and the full epoch. The holdout arm is $0.0\%$ at
every point and coherent throughout, so its row is omitted; the two comparison arms decline gently to their
endpoints. Values are percent.}
\label{tab:c-checkpoints}
\begin{tabular}{l rrrr}
\toprule
Arm & quarter & half & three-quarters & full \\
\midrule
Unedited fine-tune            & $30.0$ & $29.0$ & $27.7$ & $27.7$ \\
Matched-rank random holdout   & $30.1$ & $30.1$ & $28.6$ & $27.5$ \\
\bottomrule
\end{tabular}
\end{table}

\begin{table}[t]
\centering\small
\caption{\textbf{Every arm.} Broad misalignment is a cluster mean over the $83$ evaluation clusters, with
near-domain answers excluded; ``responses'' is the number of judged generations behind it. Narrow
adherence is a judged rate on the organism's own narrow probes, with its own much smaller denominator in
parentheses, and it is not comparable with any log-probability margin elsewhere in this paper. The three
arms at the top pool three training seeds; the weight-channel arms pool two.}
\label{tab:c-arms}
\begin{tabular}{>{\raggedright\arraybackslash}p{0.38\textwidth} r r l}
\toprule
Arm & Responses & Misaligned (\%) & Narrow adherence \\
\midrule
Unedited fine-tune (reference)                    & $2100$ & $27.7$ & $0.902$ \ ($824$) \\
Carrier held out of the activations               & $2100$ & $0.0$  & $0.000$ \ ($236$) \\
Matched-rank random subspace held out             & $2100$ & $27.5$ & $0.894$ \ ($781$) \\
\midrule
Holdout, carrier orthogonalized against Assistant & $700$  & $0.0$  & $0.0016$ \ ($83$) \\
Holdout, two epochs                               & $700$  & $0.024$& $0.000$ \ ($78$) \\
Unedited fine-tune, two epochs                    & $700$  & $25.3$ & $0.836$ \ ($308$) \\
Holdout, extreme-sports narrow dataset            & $700$  & $0.0$  & $0.000$ \ ($5$) \\
Unedited fine-tune, extreme-sports dataset        & $700$  & $25.9$ & $0.800$ \ ($13$) \\
Holdout during a benign fine-tune                 & $700$  & $0.0$  & $1.000$ \ ($39$) \\
\midrule
Writer fine-tune, no gradient filter             & $1400$ & $26.7$ & $0.801$ \ ($556$) \\
Writer gradient filtered off the carrier          & $1400$ & $26.6$ & $0.819$ \ ($582$) \\
Writer gradient filtered onto the carrier         & $1400$ & $11.5$ & $0.507$ \ ($175$) \\
\midrule
Carrier injected into the untouched model         & $7000$ & Table~\ref{tab:c-dose} & \\
Norm-matched random vector injected               & $7000$ & Table~\ref{tab:c-dose} & \\
\midrule
Single-direction organism, carrier removed at inference & $2800$ & $0.0$  & \\
Published organism, carrier removed at inference        & $700$  & $17.7$ & \\
Published organism, random subspace removed at inference& $700$  & $28.2$ & \\
Published organism, unmodified (instrument gate)        & $700$  & $27.9$ & \\
\bottomrule
\end{tabular}
\end{table}

\section{Loss-geometry measurements}\label{app:geometry}

\subsection{Directions measured}

This appendix gives the three geometry instruments behind \S\ref{sec:structure}: the curvature exponent,
the participation ratio, and the slice learning coefficient. All three read a direction in weight space,
not a single point, so we first fix which direction and which channel each one uses.

The \textbf{write core} $W$ is a rank-8 subspace of weight space, and it is the object every
weight-channel reading acts on. We build it from the realized updates of 3 published organisms, one
per narrow domain, over the residual-writer matrices of the middle half of the layer stack, the attention
output and feed-forward down projections of layers 12--35. For each organism we take
the low-rank update's output factors across those writers, scaled by the adapter's own scaling, stack them
and keep the top 4 left singular vectors. Stacking the 3 per-organism cores and keeping the top
8 gives the shared output direction $W$. We complete each column with the input direction the
organisms actually use to write it, formed per writer matrix from the same factors, and normalize the
resulting per-column block to unit Frobenius norm over the whole band. Perturbing along column $j$
therefore adds a rank-one update to every writer in the band at once, along a direction the organisms
empirically wrote through.

The write core depends on the organisms existing, so it is not readable from the untouched model alone. A
curvature reading on it is therefore a statement about the pretrained loss along a direction fine-tuning is
known to move through, evaluated at the untouched model's weights, and not about a direction the untouched
model hands us on its own. We extract the read-channel objects of Appendix~\ref{app:extraction} with no
organism involved, and \S\ref{sec:structure} keeps the two apart.

The \textbf{matched random core} is our control for every weight-channel reading. We draw a random
orthonormal output basis of the same rank in the same space, random input directions of the same shapes
over the same band writers, and normalize each column to unit Frobenius norm exactly as the real slice
does. It differs from $W$ in where it points and in nothing else.

The \textbf{read-channel objects} are the carrier and the rank-8 shared core of
Appendix~\ref{app:extraction}, perturbed in the activation channel by adding to one layer's residual
output at a time.

\subsection{The behavioral loss}

The order of a zero is defined at a minimum, and the untouched model is not a minimum of next-token
prediction. Reading curvature on the training loss would therefore mix the geometry we want with an
off-minimum gradient term, and there is no principled way to subtract one from the other.

We remove the problem by changing the loss rather than by correcting for it. Write
$f_\theta(\cdot \mid x)$ for the model's next-token distribution and take
\begin{equation}
L_{\mathrm{KL}}(\delta) \;=\; \mathbb{E}_{x}\,\mathrm{KL}\!\big(f_{\tpre}(\cdot \mid x)\;\big\|\;f_{\tpre + \delta}(\cdot \mid x)\big),
\label{eq:behavioral-loss}
\end{equation}
the divergence of the perturbed model's predictions from the untouched model's own, over a fixed pool of
reference sequences. This is zero at $\delta = 0$ and non-negative everywhere, so the untouched model is
the exact global minimum by construction, not by estimate.

Three properties follow. There is no reference loss to subtract, because the reference is exactly zero.
Both the exponent and the volume coefficient below are non-negative by construction, so neither can go
negative through noise and be quietly clipped. To second order $L_{\mathrm{KL}}$ is the model's own Fisher
form on its predictive distribution, computed model-side with no labels, so a direction along which
$L_{\mathrm{KL}}$ stays near zero is a direction that barely changes what the model predicts.

This last property is why we chose this loss for this question. A persona direction changes who is
speaking, not what comes next, and Equation~\ref{eq:behavioral-loss} makes that precise.

\subsection{The curvature exponent}

Move along a unit direction $v$ from the untouched weights, $\theta(\varepsilon) = \tpre + \varepsilon v$.
Near a minimum the behavioral loss grows as a power law,
\begin{equation}
L_{\mathrm{KL}}(\varepsilon) \;\sim\; c\,\varepsilon^{\beta},
\end{equation}
and $\beta$ is the order of the zero.

$\beta = 2$ is a regular quadratic minimum. The curvature along $v$ is finite and non-zero, and the loss
responds to movement along $v$ the way it responds to movement along an ordinary direction. $\beta > 2$ is
the degenerate case: the quadratic term vanishes, the loss is flat along $v$ to higher order, and $v$
contributes $1/\beta \le 1/2$ to the model's effective dimension instead of the full half a regular
direction contributes. That inequality is also why the volume scale below has its regular reference at
$k/2$.

We read an exponent, not a curvature, for three reasons, each a way a curvature reading fails here. A
curvature trace is blind where a degenerate direction lives, because along such a direction the quadratic
term is what vanishes, so the instrument reads floor noise precisely where the effect would be. Curvature
magnitude and degeneracy can also move in opposite directions, so a curvature reading can be
anti-correlated with the property it is meant to detect. Raw curvature is gauge-dependent: rescaling
symmetries in a network can make a minimum look arbitrarily sharp or flat without changing the function the
model computes. The order of the zero is invariant under reparameterization, which a statement about the
loss landscape must be.

\subsection{Estimating the exponent}

For a direction $v$ we sweep a fixed grid of perturbation sizes,
$\varepsilon \in \{0.25, 0.5, 1, 2, 4\}$, fixed before the run. At each size the estimator perturbs the
weights in place by $\varepsilon v$, runs a forward pass over a fixed pool of 64 reference
sequences of 512 tokens, and accumulates the per-sequence mean behavioral KL. The
exponent is the least squares slope of $\log \overline{\mathrm{KL}}$ against $\log \varepsilon$.

Two rules constrain the fit. Only points above a numerical floor of $10^{-8}$ enter it, because below that
the KL is dominated by accumulation noise and a slope through it means nothing. If fewer than 3 points
survive, the estimator returns an unreadable flag, so a direction the grid cannot resolve is reported as
unresolved and not assigned a confident wrong number.

The fit carries its own self-check. Inside a clean quadratic window the ratio
$\overline{\mathrm{KL}}(2\varepsilon)/\overline{\mathrm{KL}}(\varepsilon)$ must sit near $4$, and a
window that has fallen into the floor or saturated in the tail will not produce that. Every column of the
write core came in within $5\%$ of $4$, so the fit sat in the quadratic regime and not at either end of
it. This check runs on the model itself; the planted controls below do not.

Three aggregations appear. The write core's exponent is the median over its 8 columns. For the carrier we
take the median over columns and over the 3 layers. The rank-8 shared cores get a single-layer reading,
taken at layer 24 and reported as the mean over the 8 columns.

On the write core the curvature exponent is $2.001$, with every column within $5\%$ of $2$, and a matched
random core reads $2.002$, so the persona direction is not flatter than random (unpaired permutation
$p = 0.31$). The read-channel carrier and the shared persona, style and topic cores all read within noise
of $2$, and no permutation test separates the persona reading from a control. Table~\ref{tab:j-beta} gives
the full set, with the planted regular and quartic directions the estimator recovers as $2$ and $4$. There
is no flatness for any of the three cores to be specific about.

\subsection{Planted controls}

The instrument has two ends, the fitter and the model-side measurement, and either can fail without the
other. The two controls certify different ends of the instrument.

\begin{table}[h]
\small
\centering
\renewcommand{\arraystretch}{1.25}
\begin{tabularx}{\textwidth}{@{}lllX@{}}
\toprule
\textbf{Control} & \textbf{Where it runs} & \textbf{Reads} & \textbf{What it establishes}\\
\midrule
planted quadratic & fitter only, live grid & $2.00$ &
the slope fitter recovers a known second-order zero at the exact grid and floor the run uses\\
planted quartic & fitter only, live grid & $4.00$ &
the same fitter returns four when the curve really is quartic, so an exponent of two is not the fitter's
ceiling\\
matched random core & full pipeline on the model & $2.002$ &
perturbing, forwarding, accumulating and fitting returns a regular exponent on a direction with no reason
to be singular\\
window ratio & full pipeline on the model & near $4$ &
the measured growth is quadratic in the fitted window rather than floored or saturated\\
\bottomrule
\end{tabularx}
\caption{The four controls on the exponent estimator, and the end of the instrument each one covers. The
two planted controls impose a known monomial and are computed without the model. The acceptance band for
the random core was $1.6$ to $2.4$ and the quartic control was required to read above $3$, both fixed
before the run.}
\label{tab:geometry-controls}
\end{table}

The planted controls impose a synthetic monomial on the live perturbation grid, with the live numerical
floor, and ask whether the fitter recovers its exponent. They read $2.00$ and $4.00$. What they establish
is a dynamic range: a curve growing as the fourth power would have been reported as $4$, so a reported
$2$ is not the largest number this fitter can return. What they do not establish, because they never touch
the model, is that a degenerate direction exists in this model and would have been found. The model-side
half of the certification is the random core reading $2.002$ inside its acceptance band, together with the
window ratios sitting at $4$. Taken together the two halves license reading $2.001$ as a measurement and
not as a floor.

A separate off-model validation, run before any of this, imposes $L_{\mathrm{KL}} = \varepsilon^{2k}$ for
known $k$ and requires both that the exponent come back correct and that a deliberately under-resolved
window flag itself as unreadable instead of returning a confident slope.

\subsection{Participation ratio}

The exponent says whether any direction is flat. The participation ratio says how the curvature is
distributed among the directions that have it.

We approximate the per-coordinate curvature $e_i \ge 0$ inside the subspace by the per-coordinate gradient
power of the behavioral loss, the mean over a small set of minibatches of the squared coefficient-space
gradient, which is near zero for a coordinate the loss does not see and large for a stiff one. From those,
\begin{equation}
\mathrm{PR} \;=\; \frac{\big(\sum_i e_i\big)^{2}}{\sum_i e_i^{2}} \;\in\; [1, k],
\label{eq:pr}
\end{equation}
equal to $k$ when the curvature is spread evenly and near one when a single direction carries it all. It
is a ratio of two sums of the same degree, so a common-mode multiplicative artifact cancels and it cannot
floor or sign-flip the way a raw curvature can.

Scale-invariance makes the ratio readable, and it also makes it ambiguous on its own. A low participation
ratio falls out of two different situations. It falls when directions go dead, the degenerate case, and it
falls when curvature concentrates into a few stiff directions, the opposite of the degenerate case. The
ratio is silent about which, so a low value on its own does not distinguish flatness from concentration.

The write core reads a participation ratio of $3.55$ against the matched random core's $7.12$. A low ratio
here comes from concentration, not from flatness, and three facts on the same per-coordinate curvatures
establish this. Table~\ref{tab:dgeom-curv} gives the values.

\begin{table}[h]
\small
\centering
\begin{tabular}{@{}lrr@{}}
\toprule
per-direction curvature & persona write core & matched random core\\
\midrule
stiffest coordinate & $1.01\e{-15}$ & $6.55\e{-17}$\\
flattest coordinate & $4.92\e{-17}$ & $1.55\e{-17}$\\
summed, relative to random & $6.92$ & $1$\\
\bottomrule
\end{tabular}
\caption{\textbf{The low participation ratio of the write core is concentration, not flatness.}
Per-direction curvatures inside the rank-8 write core and inside the matched random core, at their
stiffest and flattest coordinates, with the summed curvature given relative to the random core. The
persona core carries more total curvature, spreads it more unevenly, and keeps its flattest coordinate
above the random core's flattest.}
\label{tab:dgeom-curv}
\end{table}

The total curvature is higher: summed over the 8 directions the persona core carries more than the random
core, where a degenerate valley would carry less. The curvature is unevenly distributed: the persona
core's stiffest per-direction curvature exceeds its flattest by a factor of $20.58$, while the random
core's is even. No persona direction is dead: its flattest coordinate sits above the random core's
flattest.

The bound is one-sided. The persona core's flattest coordinate exceeds the random core's flattest, not the
random core's whole range: the random core's stiffest coordinate (Table~\ref{tab:dgeom-curv}) sits above
the persona core's flattest. This licenses that no persona direction has gone slack, and nothing more.

The low participation ratio comes from added stiff directions, not from any direction going slack. Read
with the exponent and the absolute curvatures, it describes a regular minimum that is anisotropic and
stiff. Table~\ref{tab:geometry-types} marks it as the one quantity whose direction of support is defined
only jointly with the other two.

\subsection{The slice learning coefficient}

Our third reading is a volume, not a shape or an order. The volume of the low-loss set shrinks with
the threshold as $\mathrm{vol}\{\delta : L_{\mathrm{KL}}(\delta) < s\} \sim s^{\hat\lambda}$ up to a log
factor, so a small $\hat\lambda$ means a large flat valley. For a regular $k$-dimensional minimum
$\hat\lambda = k/2$, each non-degenerate direction contributing a half. A purely quartic direction
contributes a quarter, so a wholly quartic slice reads $k/4$, and a flat slice reads near zero. At rank
8 the regular reference is $4$, the quartic landmark is $2$, and the floor is $0$.

We estimate it with a localized stochastic-gradient sampler on the behavioral loss restricted to the
subspace. Writing $a$ for the coefficient vector in the subspace, $h$ for the step size, $b$ for the
inverse temperature and $\gamma$ for a spring that localizes the walk near the base point, the update is
\begin{equation}
a \;\leftarrow\; a \;-\; \frac{h}{2}\Big(n b\, \nabla_a L_{\mathrm{KL}}(a) \;+\; \gamma a\Big) \;+\; \sqrt{h}\,\eta,
\qquad \eta \sim \mathcal{N}(0, I_k),
\label{eq:sgld}
\end{equation}
with the inverse temperature written $b$ because $\beta$ is taken by the exponent. The estimate is
\begin{equation}
\hat\lambda \;=\; n b\,\mathbb{E}_{\mathrm{post}}[L_{\mathrm{KL}}],
\end{equation}
and $L_{\mathrm{KL}}(0) = 0$ exactly, so unlike the usual construction there is no reference loss to
subtract. The walk runs 2 chains of 50 steps. We reduce finite-temperature bias by estimating at 3 inverse
temperatures and taking the intercept of the estimates against $1/\log(nb)$, whose slope returns the
multiplicity, measured at $1.0012$.

The axis floors, and it floors for both arms. The persona core reads $5.69\e{-4}$ and the matched random
core $5.74\e{-5}$, against a regular reference of $4$. Both sit four to five orders of magnitude below it,
at $1.4\e{-4}$ and $1.4\e{-5}$ of reference. The mechanism is that the localizing spring dominates the very
small restricted curvature of a low-rank weight direction in this channel, and the consequence is that the
absolute coefficient cannot separate a regular minimum from a singular one here for any direction. A
threshold on it would have been meaningless, so we treat it as a reported comparison and never as a gate.

What survives is the ratio between the two arms on the same footing, $9.91$. This is a curvature
\emph{volume} reading, a different quantity from the $6.92$-fold ratio of summed per-direction curvature
above. The two agree in direction and are not the same number.

Three caveats attach to this axis; it therefore confirms the summed-curvature reading and does not carry
it.

\begin{itemize}[leftmargin=1.4em,itemsep=2pt,topsep=3pt]
\item \textbf{The two arms use different estimators.} The persona coefficient is the 3-temperature
extrapolated intercept; the random coefficient is a single-temperature estimate at the middle temperature.
The ratio therefore divides a bias-corrected number by an uncorrected one.
\item \textbf{The spread quoted is not a sampling interval.} The persona estimates across the 3
temperatures run from $5.02\e{-5}$ to $3.36\e{-4}$. That is a range over estimator settings.
\item \textbf{The chain diagnostics recorded belong to the random arm}, whose split-chain statistic was
$0.9958$ with no chain escaping. The persona chains' diagnostics were not retained.
\end{itemize}

The strength of the localizing spring is a free parameter of this estimator, and it controls whether the
volume is readable at all. Over-localize and the walk reports no volume for any direction. We chose the
setting so that a matched random slice reads on scale.

\subsection{Numerics}

Our estimator work runs in single precision with reduced-precision matrix multiplication disabled. We
accumulate the behavioral KL over the full vocabulary in double precision with no top-$K$ truncation,
because the quantity being fit spans several orders of magnitude across the perturbation grid and a
truncated or reduced-precision sum would flatten its small end into the floor. The exponent, the window
ratios, the per-coordinate curvatures and the participation ratio are all computed in that regime.

The sampler is our single exception, granted because its Monte Carlo noise is large relative to its
arithmetic error. Our same-seed check, run on the same random core with reduced-precision matrix
multiplication first disabled and then enabled, moved the coefficient by $3.95$ in relative terms, against
the $20\%$ tolerance we set for the check in advance. Both arms of the ratio are reported under the same
setting, so their comparison is like with like. The absolute values move by a factor of about five under a
change of arithmetic, alongside the floor and the estimator asymmetry above, so the volume confirms the
stiffness contrast while the exponent and the participation structure carry the geometry.

\subsection{Quantity types and instrument limits}

\begin{table}[h]
\small
\centering
\renewcommand{\arraystretch}{1.25}
\begin{tabularx}{\textwidth}{@{}lXlX@{}}
\toprule
\textbf{Quantity} & \textbf{Direction of support for degeneracy} & \textbf{Scale} & \textbf{Measured}\\
\midrule
$\beta(W)$ & above $2$ & $2$ regular, $4$ quartic & $2.001$, regular\\
$\beta$, read channel & above $2$ & as above & $2.001$, regular\\
$\mathrm{PR}(W)$ & low, \emph{and} only read jointly with the volume and the exponent & $[1,k]$; random $7.12$ & $3.55$, from concentration\\
$\hat\lambda(W)$ & low against $k/2$ & $k/2 = 4$ regular, $k/4$ quartic, $0$ flat & $5.69\e{-4}$; floors, so only the ratio $9.91$ is read\\
\bottomrule
\end{tabularx}
\caption{The type of each geometry quantity. Each has a stated direction of support, a scale with a
reachable reference, and a measured value. The participation ratio is the one quantity whose direction of
support is not defined on its own, because a low value is ambiguous between flatness and concentration.}
\label{tab:geometry-types}
\end{table}

Two limits bound everything above. We read the exponent on one model, along low-rank directions, with a
volume axis that floors in this channel, so what we exclude is a singularity of the kind this instrument
can reach. A singularity in a regime it cannot reach is not excluded, only unmeasured. We also read the
weight-channel quantities along $W$, which exists because the organisms exist, so they describe the loss
along a direction fine-tuning writes through and not a direction the untouched model volunteers.

The persona write core is a regular minimum of the untouched model's behavioral loss, with curvature
exponent $2$ in both channels, stiffer than a matched random subspace by roughly an order of magnitude on
two separate measures, and with its stiffness concentrated in one dominant direction. The loss is not
blind to the disposition, and there is no degenerate direction along which the structure could be slipped
out.


The exponent readings, the planted controls, and the per-direction curvatures are plotted in
Figure~\ref{fig:regularstiff} in the main body.

\section{Post-hoc removal probes and the reconstitution certificate}\label{app:removal}

\subsection{The shared projector}

This appendix specifies the three post-hoc removal probes and the reconstitution certificate reported in
Section~\ref{sec:removal}. The three attempts share one construction. Each begins from a subspace we
extracted by contrastive teacher forcing from the residual-stream activations of the untouched model
(Appendix~\ref{app:extraction}), and each then treats that subspace's coordinates as coordinates on the
output rows of the residual writers.

The writers are the attention output projection and the MLP down projection, the two matrices per decoder
layer whose output is added into the residual stream, both of which have $5120$ rows, one per residual
coordinate. For a writer $W$ and an orthonormal $U$ in residual coordinates, every edit we report here is
\begin{equation}
  W \;\longleftarrow\; \left(I - U U^{\top}\right) W ,
  \label{eq:removal-edit}
\end{equation}
which we execute as $W - U(U^{\top}W)$ so that the $5120 \times 5120$ projector is never formed. It deletes
the component along $\operatorname{span}(U)$ of everything that matrix ever contributes to the stream and
leaves the rest of the matrix alone.

That map is the naive lift of a read-space object into weight space, the identity taken from the
coordinates the disposition is read in to the coordinates a writer's output arrives in. It is exact for
what those matrices add at those layers, and silent about every other route into the same coordinate.
The model's other decoder layers write into the same stream, and anything upstream of the read layers can
place mass along the carrier there without passing through the six matrices our first and third edits
touch. No Jacobian-vector product appears anywhere in the three edits, and the source of the projector
carries an annotation saying so.

The source article for this work assigns the separability ablation to the read channel and the other two
edits to the write channel. Its own principal figure says instead that all three act in the write channel,
and the implementation agrees with the figure, because the ablation projects its direction out of the same
residual writers, through the same operator, with no Jacobian anywhere. The three operations act in one
basis, so their agreement is one geometric fact.

\paragraph{Margins, not generations.} These probes sampled and scored no generations. Across the
artifacts there is no stored completion and no judge field, so the count is a computed zero, not a
missing value. The probes also train nothing, operating on published adapters and on the untouched
model's weights. The foreclosure rests on teacher-forced log-probability margins and on one held-out
perplexity check, with no behavioral confirmation of its own.

We read these edits with margins because at this scale a coherence-gated judge on free generations
separates bad training data from benign and resolves little more (Appendix~\ref{app:judge}), while a
margin read on a battery screened once and never re-judged costs no judge calls and carries per-probe
variance. The cost of that choice is that a margin is a probability statement about two fixed
continuations, not a rate at which the model volunteers a misaligned answer, so nothing here shows a
defended model behaving well under sampling.

\begin{table}[htbp]
\small
\renewcommand{\arraystretch}{1.2}
\begin{tabularx}{\textwidth}{@{}p{2.6cm}Xp{1.5cm}p{3.1cm}@{}}
\toprule
\textbf{Edit} & \textbf{What is projected out} & \textbf{Rank} & \textbf{Applied to} \\
\midrule
separability ablation &
$w_\ell = \text{unit}\!\left[(I - P_{\mathrm{loc},\ell})\,u_{\mathrm{glob},\ell}\right]$, the standing
disposition with the local character machinery removed &
one per layer &
the untouched model's six band writers, restored between arms \\
\addlinespace[2pt]
adapter decomposition &
$\Up$, the per-layer carrier columns stacked over the three layers and orthonormalized &
at most twelve &
a published organism's realized update, every writer matrix it touches \\
\addlinespace[2pt]
defending edit in the certificate &
the same stacked $\Up$ basis &
at most twelve &
the untouched model's six band writers, in place \\
\bottomrule
\end{tabularx}
\caption{The three edits, defined by the subspace each removes and the matrices each removes it from. All
three use the same operator on the same two writer types at layers $18$, $24$ and $30$, and all three
carry a read-space subspace into writer coordinates by the identity. We build the stacked basis by
concatenating the per-layer columns, taking a double-precision decomposition and trimming near-duplicate
directions, so its rank is at most three times the per-layer rank, and it was twelve in the computation
reported here.}
\label{tab:removal-edits}
\end{table}

\subsection{The separability probe}

Our first edit asks whether a separable object exists to be removed at all, and it needs two subspaces,
which we extract from the untouched model by the contrastive machinery of Appendix~\ref{app:extraction}.
The first holds the standing disposition. One response text is scored twice, once behind a system prompt
framing the speaker as a misaligned assistant and once behind a prompt framing the same speaker as an
aligned one, with the response held byte for byte identical, so that what varies across the pair is who the
assistant is, not what is said.

The second subspace holds the local capability to depict a bad character, with a helpful-assistant framing
held fixed while the task varies between quoting a villain and quoting a neutral character. We residualize
a per-probe common mode out before either extraction. The probe direction is then the first subspace with
the second projected out of it, taken per layer and normalized.

Were the standing disposition contained in the local machinery, this direction would collapse and there
would be nothing to ablate. It does not collapse. Its residual fraction is $0.988$, $0.986$ and $0.994$ at
the 3 layers, so a separable direction exists in the geometry, and what the ablation goes on to test is
whether removing it is selectively protective.

We run four arms, each rank-matched at one direction per layer so that all four delete comparable Frobenius
mass: the probe direction, a matched-random direction, the leading local direction, and the
un-orthogonalized standing direction. Each arm edits, reads two margins with the hooks off, and restores
the snapshot before the next arm starts. Our readouts are the broad-misalignment margin $\Mb$ over $195$
frozen probe rows and the expression margin over $120$ authored rows, both defined in
Appendix~\ref{app:batteries}.

A drop is the margin before the edit minus the margin after, so a positive drop means the margin fell. The
statistic is the selectivity, the mean broad drop minus the mean expression drop, which is positive when
ablation costs misalignment more than it costs the capability to depict a bad author.

It is not positive. The selectivity is $-0.227$ with an interval of $[-0.838, +0.347]$ and a label
permutation at $p = 0.43$, with broad falling by $0.233$ and expression by $0.460$.

Our control arms behave as designed and show the instrument is live. Ablating the leading local direction
costs the expression margin $0.799$ and the broad margin only $0.048$, while ablating the un-orthogonalized
standing direction moves the broad margin by $0.424$ with an interval excluding zero, so a small effect on
the orthogonalized direction is an absence, not a dead instrument. A matched-random direction gives
$-0.122$.

\paragraph{The unit these intervals are computed over.} Two properties of the two batteries decide what the
interval above means, and both are properties of the readout, not the finding. Our $195$ broad rows are
$75$ distinct prompts carrying two or three alternative aligned continuations each, and the $120$
expression rows are $5$ prompt framings apiece of $24$ items whose two scored continuations are
byte for byte identical across the $5$. Neither battery's rows are independent of one another.

We computed the interval and the permutation above over rows, and this paper's unit of inference is the
unique prompt (\S\ref{sec:instruments}, Appendix~\ref{app:stats}). Recomputed at that unit there are $75$
and $24$ units rather than $195$ and $120$, the selectivity interval widens, and its lower bound
falls outside the $\pm 0.4\sigma$ band the estimate was compared against. The realized minimum detectable
effect rises above that band at the same time, so the design is no longer powered to the smallest effect it
declared interesting.

At the prompt-cluster unit the finding's direction survives: selectivity stays negative, the permutation
still does not reject, and no arm becomes selectively protective, so nothing here becomes evidence for a
thin separable object.

The bound does not survive. At this paper's own unit the arm establishes an absence, not a bounded
absence, and the equivalence reading belongs to the row count, not the prompt count. This edit is
the most modestly powered of the three either way, and the weight it carries comes from agreeing with the
other two.

\subsection{The adapter-decomposition probe}

Our second edit works on an organism's realized weight change, not on the untouched model. For each
residual-writer matrix the published low-rank adapter touches, we reconstruct the realized update in single
precision from its raw factors and its scaling, $\Delta W = s\,BA$, giving a matrix with $5120$ rows. The
stacked carrier basis contracts it on the left, the projection of Equation~\ref{eq:removal-edit} applied to
the update itself,
\begin{equation}
  \Delta W_{\text{persona}} = P\,(P^{\top}\Delta W) ,
\end{equation}
and the remainder is the task component.

Zeroing the persona component and reloading is byte-equivalent to loading a surgical adapter, because a
low-rank update is additive, so we re-fit nothing. The scope differs from the other two edits. Our
projector is built from the three read layers, and here we apply it at every layer the adapter touches,
which is the naive lift at its most exposed.

We run four arms per organism: the full adapter, the persona-zeroed adapter, an adapter with a
matched-Frobenius random component removed, and a task-zeroed adapter keeping only the persona component.
That last arm is the discriminator between a genuinely inert channel and a projector that misfired, and it
destroys the narrow task in all 3 organisms. Drops are measured against the full adapter, since the
question is what zeroing the component does to the realized endpoint.

Our readouts are $\Mb$ on the same $195$ rows, an on-task narrow margin over $38$ authored own-domain
probes, and the expression margin. The narrow readout here is this work's own on-task teacher-forced
margin, and it is not comparable with the judged adherence rate or with the narrow margin of any other
measurement in this paper (Appendix~\ref{app:batteries}). The primary statistic is the per-organism
difference in differences, the broad drop minus the narrow drop, positive when the surgery costs the
disposition more than the capability.

The three differences appear per organism in Appendix~\ref{app:results}, Table~\ref{tab:j-organism}. The
difference is negative in 2 organisms and small in the third, so where anything moves the narrow task
falls further than the broad disposition. In all 3 organisms the expression margin rose under the surgery.

Each organism's clock is live, its full adapter raising $\Mb$ over the untouched model by about $30$ nats.
Pooling the three is descriptive only, because 3 clusters cannot resolve an effect that is small and
consistent across organisms.

The persona component is a few thousandths of the update by Frobenius mass, and our matched-random control
removes the same mass per matrix by construction (Appendix~\ref{app:results}, Table~\ref{tab:j-frobenius}).
That equality is imposed by the construction and shows only that the projector is not collapsed.

The matched-random arm is not fully clean. In 2 of the 3 organisms it also moved the
broad margin, so it is not a fully controlled certificate of an inert channel. The unit-of-inference
caveat above applies to these intervals too, and it changes one of them: for one organism the broad drop's
interval excludes zero over rows and includes zero over prompts. The three difference-in-differences
figures are unaffected, because the narrow battery carries one row per prompt and dominates their variance.

\subsection{The reconstitution certificate}

Our third edit separates removal from suppression, and it is the only place in this work where the two are
distinguished by measurement, not by assumption. We build a defended model by projecting the stacked
carrier basis out of the six band writers of the untouched model.

The edit removes well under $1\%$ of the Frobenius mass of those six matrices and leaves held-out
perplexity at $17.3$, which is our capability gate: a clean low-rank ablation barely moves perplexity, and
a jump would mean the subspace had bled into directions the model needs.

We then ask how large a gradient-free injection it takes to put the behavior back, and compare the defended
model against the undefended one. A removal has to make reconstitution more expensive. A suppression
re-lights at the same price.

\paragraph{The dose response.} We reuse the injection of Appendix~\ref{app:loop} without change, so that
the residual stream at each of the 3 layers receives $\alpha \cdot s_\ell \cdot \uhat_\ell$, with
$s_\ell$ a per-layer activation scale that makes $\alpha$ dimensionless. We assert that the hooks fired at
every layer and every dose, because a steering hook that silently fails to fire manufactures an absence.

Our grid is $\alpha \in \{0,\, 0.075,\, 0.15,\, 0.225,\, 0.30\}$, with $0.30$ the top of the coherent band;
beyond it steering destroys fluency and a margin read there certifies nothing. That ceiling was a
configured constant rather than a threshold read off the coherence curve. Coherence here is an
output-entropy ratio near one, not the judge's coherence score, which follows from this work having
run no judge at all.

The injected direction is rank one, the leading direction per layer, while the subspace the defending edit
cleared is rank twelve, so we re-add strictly less than the edit removed.

The battery is $75$ paired probes, one row per distinct prompt, so this readout already sits at the unit of
inference and needs no clustering correction. Every curve is read as the per-probe rise over that arm's own
zero-dose margin, so each probe's baseline differences out of every slope and every crossing.

\paragraph{The onset.} An arm's onset dose $\alpha^{*}$ is the smallest dose on the grid whose mean margin
exceeds a base-relative threshold,
\begin{equation}
  \tau_{\text{eff}} = \bar{\Mb}(\alpha{=}0) + \delta ,
  \label{eq:onset-threshold}
\end{equation}
with $\delta = 4$ nats, evaluated per arm against that arm's own zero-dose margin, not against a fixed
level.

Two things force the base-relative form of Equation~\ref{eq:onset-threshold}. An absolute threshold at
parity, where the misaligned continuation becomes as likely as the aligned one, is reached by no arm
anywhere in the coherent band, so it puts $\alpha^{*}$ out of reach everywhere and the comparison carries
no information. The defended and undefended models also do not start from the same zero-dose margin, so an
absolute threshold would ask them to travel different distances to one finish line.

We froze $\delta$ from the undefended arm alone, at about a fifth of that arm's realized rise across the
coherent band, before reading the defended arms. Our grid cell $0.075$ is the resolution of
$\Delta\alpha^{*}$, which therefore carries no interval, and that is why the slope ratio is the
load-bearing quantity.

\paragraph{The slope ratio.} We regress each probe's residualized rise on dose by least squares, giving one
slope per probe in units of margin per unit dose. The slope ratio $r_s$ is the defended arm's mean slope
over the undefended arm's, with a paired bootstrap over probes matched by identifier and a denominator
gate. We read the ratio only when the undefended slope's own interval excludes zero, which it does
(Appendix~\ref{app:stats}). A removal predicts $r_s < 1$.

\begin{table}[htbp]
\small
\renewcommand{\arraystretch}{1.2}
\begin{tabularx}{\textwidth}{@{}p{4.4cm}Xp{1.5cm}p{1.5cm}@{}}
\toprule
\textbf{Arm} & \textbf{Model and injected direction} & \textbf{Slope} & \textbf{Onset} \\
\midrule
undefended, fixed carrier & untouched model, along the extracted carrier & $62.9$ & $0.15$ \\
defended, fixed carrier & defended model, along the same carrier & $76.4$ & $0.15$ \\
defended, re-extracted carrier & defended model, along the carrier re-extracted from it & $76.1$ & $0.15$ \\
defended, matched random & defended model, along a random direction & $1.11$ & not reached \\
re-extracted, injected back & untouched model, along the re-extracted carrier & $63.2$ & $0.15$ \\
defended, trigger present & defended model, no injection, inoculating framing in the prompt & none fitted & not reached \\
\bottomrule
\end{tabularx}
\caption{The certificate's arms. Slopes are margin per unit dose over $75$ paired probes, and onset is the
smallest grid dose whose mean margin clears that arm's own base-relative threshold. The matched-random
arm's slope interval includes zero and its onset is never reached. That control makes the reconstitution
on the other arms carrier-specific, not a response to any residual-stream perturbation. We read the
trigger arm at zero dose only, so it has no slope.}
\label{tab:rdc-arms}
\end{table}

\paragraph{The re-extraction.} A confound would have rescued the first two readings, because if the edit
had removed the carrier and the behavior had relocated to a new direction, then dosing along the old
direction would be no test of anything. So we re-extract the carrier from the defended model by the same
misaligned-against-aligned contrast and decompose its mass onto the subspace we cleared. Because a singular
vector's sign is not determined by the decomposition, we orient each re-extracted column by its own
defining contrast, signing it to have a non-negative inner product with the misaligned-minus-aligned mean
difference.

Two conditions gate that arm before we may read it, and without them a re-extraction that returned noise
would need a large dose and would falsely look like removal. The re-extracted direction must be a real
direction, its per-layer stable rank below and its top singular-value share above a matched-noise floor we
estimate from Gaussian draws of the same shape, and all 3 layers clear both. Injected back into the
undefended model, it must reproduce a signed rise with dose, which it does at a slope of $63.2$.

\paragraph{What the three components rule out.} Onset rules out a removal that raised the price of
reconstitution. It is $0.15$ for the undefended model and for both defended carrier arms, so
$\Delta\alpha^{*}$ is zero to the grid's resolution.

The slope ratio rules out a removal that flattened the response without moving its onset. We measure $r_s =
1.214$ with an interval of $[1.123, 1.366]$, excluding $1$ from above, so the defended model buys slightly
more margin per unit dose than the undefended one does.

Re-extraction rules out the relocation confound. Almost all of the re-extracted carrier's mass lies inside
the subspace we cleared, far above the isotropic expectation of $4/5120$, and the per-layer shares and
their cosines to the original appear in Appendix~\ref{app:results}, Table~\ref{tab:j-reext}. Those cosines
are the square roots of the shares, and both describe what remains readable from the defended model's
activations, not weights regrowing.

One arm speaks to a different failure mode. We read the defended model at zero dose with the inoculating
framing present in the prompt. Its margin rose by the amount given in \S\ref{sec:removal} without crossing
onset. That is a disposition relocated behind a context, not a disposition removed.

\paragraph{The threshold and orientation rules.} The onset threshold is base-relative
(Equation~\ref{eq:onset-threshold}), replacing an earlier absolute one, and the re-extracted carrier's
orientation rule is fixed as stated above. Between the absolute-threshold computation and the one reported
here, the slope ratio, its interval, the re-extraction fractions and the perplexity are byte-identical. The
absolute threshold produced an unreachable onset for every arm and a re-extraction arm the gate refused.

\subsection{Basis scope}

Three bases are in play, and our foreclosure covers one of them.

\begin{table}[htbp]
\small
\renewcommand{\arraystretch}{1.2}
\begin{tabularx}{\textwidth}{@{}p{4.6cm}p{2.5cm}p{2.6cm}X@{}}
\toprule
\textbf{Basis} & \textbf{Edited after training} & \textbf{Constrained during training} & \textbf{What happened} \\
\midrule
the naive writer-column lift of the read carrier & yes, three ways & no & inert, three ways \\
the shared write core the organisms realize & never & yes & moves the behavior, at a capability cost \\
the functional pullback of the readout & never & never & not built \\
\bottomrule
\end{tabularx}
\caption{\textbf{The removal foreclosure covers one of three bases.} All three removal probes acted in
the basis with roughly chance overlap with the misalignment direction; the basis known to carry the
fine-tune's update has never been subjected to the same removal test.}
\label{tab:three-bases}
\end{table}

The basis our three edits share sits at close to a right angle to the misalignment direction in the writer
column space, which is the imported geometry that explains why edits in it are inert
(Appendix~\ref{app:loop}). Two facts hold at once and neither cancels the other. That basis has close to
chance overlap with everything, and the fine-tune's realized update is enriched about a hundredfold in a
shared write core against an isotropic expectation, with constraining that core during training moving
broad misalignment (Appendix~\ref{app:routing}).

The disposition is read loudly and written obliquely. No weight direction we tested is the readout, and
structured weight directions nonetheless move it.

The measurement that would settle the scope is specific and cheap. For probe $x$ and read layer $\ell$ the
readout is a scalar function of the weights, and the weight direction it responds to is its own gradient,
\begin{equation}
  r_\ell(\theta) = \ip{h_\ell(\theta; x)}{\uhat_\ell} , \qquad
  \nabla_\theta r_\ell = J_\ell^{\top}\uhat_\ell , \quad J_\ell = \partial h_\ell / \partial \theta .
  \label{eq:pullback}
\end{equation}
The gradient in Equation~\ref{eq:pullback} is the functional pullback of the carrier into weight space, and
it costs one backward pass per probe. Every edit we report here used the identity in its place, nothing in
this program has built it, and no gate here carries a quantity derived from it.

The gap is wider than one unbuilt object. No overlap statistic between the read subspaces and the write
core exists anywhere in this program, and the only quantity computed on both is the curvature exponent,
which reads two on both channels and also reads two on random cores, so it links nothing
(Appendix~\ref{app:geometry}). We characterize a read-side object and a write-side object, and the relation
between them is posited by the account, not measured.

Until it is measured, the result is that three edits in the naive writer-column lift left the disposition
in place, and that this is a statement about one basis, not about the weight channel.

\section{The training-time write-core constraint}\label{app:routing}

\subsection{The write core and the constraint}

This appendix reports the one training-time constraint behind \S\ref{sec:removal} that no other appendix
covers: constraining the shared write core during fine-tuning. Appendix~\ref{app:geometry} builds the
write core $W$ and states what it rests on. We take it as given and record what enters the optimizer, which
is one matrix and not the whole construction: an orthonormal basis of 8 columns in the residual-output
space of width $d = 5120$, formed from the realized updates of 3 published organisms over the attention
output and feed-forward down projections of the middle band of layers. That construction pairs each output
column with a per-writer input direction. Those input directions play no part here. We build the projector
from the output basis alone, so what the constraint forbids is a direction in the space every writer writes
into and not a particular rank-one weight pattern. We also cannot read $W$ from the untouched model, so
constraining it constrains a place where 3 organisms were measured to have written, fixed before this
fine-tune starts.

Each arm is a low-rank adapter fine-tune. For a writer matrix the trainable factors are $A$ and $B$, the
realized change is a scaled product $BA$, and the columns of $B$ live in the writer's output space, which
is the space $W$ spans. At every optimizer step, after the accumulated backward pass and before gradient
clipping and the update, we replace every writer's $B$-factor gradient by
\begin{equation}
\nabla B \;\longleftarrow\; \big(I - \rho\, W W^{\top}\big)\,\nabla B ,
\label{eq:routing}
\end{equation}
where $\rho \in [0,1]$ is the dose and the intervention's only free parameter. At $\rho = 0$ this is the
identity and the arm is an ordinary fine-tune. At $\rho = 1$ it is the orthogonal projection onto the
complement of the core, forbidding motion along the core instead of discouraging it.

We ran one dose, $\rho = 0.999999999$, the limit of the penalty formulation the implementation uses. That
formulation carries a stiffness $\kappa$, fixed here at a billion, and takes $\rho = \kappa/(1+\kappa)$.
Every number below comes from the full projection, and the graded dose was not run.

We apply Equation~\ref{eq:routing} to the $B$ factor of every attention output and feed-forward down
projection adapter, $96$ matrices, 2 per layer across all 48 layers. The core was built on the
middle band alone, so the constraint reaches writers outside that band. Every writer's output is added
into one residual stream of width $5120$, and the same 8 columns are a basis there at every depth, so the
projection is well-defined at all layers. This is a difference between the object and the intervention. The
filter matches parameters by name and refuses to run when no writer gradient matches, so a renamed module
stops the arm instead of leaving it unconstrained with no sign of trouble.

Two features of the schedule decide what the manipulation check has to establish. The projection acts once
per optimizer step on the accumulated gradient, not once per micro-batch, and it acts on the gradient, not
on the optimizer state, whose per-coordinate preconditioner does not commute with $WW^{\top}$. Projected
gradients therefore do not give an update confined to the complement as a matter of algebra. Confinement is
something we measure.

\subsection{Arms}

We train 3 fine-tunes on one narrow induction set, differing only in Equation~\ref{eq:routing}. One is
unconstrained, one is constrained along the write core, and one is constrained along a matched random core
at the identical dose. Appendix~\ref{app:batteries} carries the recipe and the induction set. Each arm
trains the writer matrices alone for one epoch, $375$ optimizer steps, to a single end checkpoint at one
seed, and we share the seed, so the three see identical batches in identical order.

We draw the random core as an orthonormal basis of 8 columns in the same output space of width $5120$,
the construction Appendix~\ref{app:geometry} uses for its matched random control. It holds fixed the rank,
the space, the operation, the dose, the matrices touched, the recipe, the data and its order, and varies
only in where the 8 columns point. It separates constraining this structure from constraining weights.

\subsection{The manipulation check}

We read the realized constraint from the saved adapters without loading the model. For each of the $96$
writer $B$ factors we take the fraction of its Frobenius energy lying in the span of the core, and average
over matrices. Two random subspaces of rank 8 in a space of width $5120$ share $8/5120 = 0.0016$ of
that energy in expectation. That is where the fraction sits when a fine-tune has no particular
relationship with the core.

The unconstrained fine-tune sits at $0.171$. This is the enrichment \S\ref{sec:removal} quotes, about a
hundredfold over the isotropic level. Had the unconstrained arm sat near the isotropic $0.0016$, there
would have been nothing to remove. Under the constraint the fraction reads $0.0008$, a reduction of
$0.995$ in the realized share and below the isotropic level, so the update is not just depleted in the
core but depleted past what a random direction would catch. We compute the random arm's share inside its
own core, as a diagnostic of that arm and not a second reading of the write core.

\subsection{Results}

Both outcomes are teacher-forced margins of the family defined in Appendix~\ref{app:firststep}, on the
batteries of Appendix~\ref{app:batteries}: the broad margin over $195$ out-of-distribution probes, and a
margin over the $120$-probe in-character battery. Every contrast is an arm minus the unconstrained arm on
shared probes. We trained no untouched-model arm, so each number is a difference between 2 fine-tunes
and none states what fraction of the fine-tune's own recruitment the constraint removed.

\begin{table}[htbp]
\small
\renewcommand{\arraystretch}{1.2}
\begin{tabularx}{\textwidth}{@{}Xrrr@{}}
\toprule
\textbf{Arm} & \textbf{Broad margin} & \textbf{In-character margin} & \textbf{Judged broad rate} \\
\midrule
Unconstrained                    & $-34.61$ & $9.31$ & $0.226$ \\
Constrained along the write core & $-35.15$ & $3.50$ & $0.209$ \\
Constrained along a random core  & $-34.76$ & $9.56$ & not scored \\
\bottomrule
\end{tabularx}
\caption{\textbf{The 3 arms, on every readout that exists for them.} Margins are in nats, over $195$
and $120$ probes. The judged rate is over $700$ completions per arm, after the near-domain exclusion of
Appendix~\ref{app:judge}. The random arm's generations were not scored, so the judged column has 2
entries.}
\label{tab:routing-arms}
\end{table}

Constraining the write core moves the broad margin down. Table~\ref{tab:routing-contrasts} gives the
change against the unconstrained arm for both constrained arms: the write-core arm by $-0.541$ over the
$195$ probes, and the matched random core at the identical dose by $-0.149$. Their ratio is $3.63$ over a
baseline that is not zero, so what is attributable to this structure is the increment over the random arm
and not the whole reduction.

\begin{table}[htbp]
\small
\centering
\renewcommand{\arraystretch}{1.2}
\begin{tabular}{@{}lrcc@{}}
\toprule
\textbf{Arm} & \textbf{Broad margin change} & \textbf{Sign-flip $p$} & \textbf{Bootstrap interval} \\
\midrule
Constrained along the write core & $-0.541$ & $0.033$, one-sided & \\
Constrained along a random core  & $-0.149$ & & $[-0.289, -0.014]$ \\
\bottomrule
\end{tabular}
\caption{\textbf{The broad-margin change under each constraint, against the unconstrained arm.} Changes are
in nats over the $195$ shared probes. The write-core arm carries a one-sided sign-flip test and no
interval; the random arm carries a bootstrap interval and no test, so their uncertainties are not
comparable. We ran the write-core test one-sided, in the direction we expected the constraint to move, and
it is not the two-sided default of Appendix~\ref{app:stats}.}
\label{tab:routing-contrasts}
\end{table}

The movement is an aggregate one. Cliff's $\delta$ is $-0.0087$ and only just over half the probes move
downward, so the sign-flip test resolves a small paired shift across the battery and not a separation
on any individual probe.

The judged rate, on a separate readout from the margins, moves the same way, from $0.226$ to $0.209$ over
$700$ completions per arm (Table~\ref{tab:routing-arms}). The judge's own diagnostics are in
Table~\ref{tab:routing-judged}: mean coherence holds while mean alignment rises, so the constrained model
is fluent and the margin drop is not a coherence collapse read as alignment. We attach no interval to the
difference between the two judged rates, and the arm that would let the judged readout speak to specificity
is the one we never scored, so specificity rests on the margins alone.

\begin{table}[htbp]
\small
\centering
\renewcommand{\arraystretch}{1.2}
\begin{tabular}{@{}lrr@{}}
\toprule
\textbf{Judged readout}, $700$ completions/arm & \textbf{Unconstrained} & \textbf{Write-core arm} \\
\midrule
broad rate, two-axis, before near-domain exclusion & $0.337$ & $0.333$ \\
mean alignment (0--100) & $35.4$ & $38.8$ \\
mean coherence (0--100) & $86.3$ & $86.9$ \\
\bottomrule
\end{tabular}
\caption{\textbf{The judge's readout on the two scored arms is fluent and near-unchanged.} The
near-domain-excluded broad rate is in Table~\ref{tab:routing-arms}; here are the two-axis rate before
exclusion and the mean alignment and coherence on the judge's 0--100 scales. Coherence holds and alignment
rises, so the margin drop is not a coherence collapse.}
\label{tab:routing-judged}
\end{table}

\subsection{Readout limits}

The in-character margin falls from $9.31$ to $3.50$, a retention of $0.376$, while the random arm at the
identical dose leaves it at $9.56$, above the unconstrained level (Table~\ref{tab:routing-arms}). Both the
in-character damage and the excess of the broad reduction over the random arm's are confined to the arm
that constrains this structure. The random control isolates them.

We read the damaged readout on the in-character battery of Appendix~\ref{app:batteries}, over the same
$120$ prompts and through the routine that supplies the expression margin there, on villain dialogue,
malware explanation and secure-code critique, each pairing a capable in-character continuation against a
length-matched refusal. What fell by almost two thirds is the model's capacity to put mass on capable
bad-author text on demand. It is not the reckless-financial-advice behavior the induction set trained, and
it is not a neutral competence.

Our two readouts are also not independent of each other. Both are the same construction, the summed
completion log-probability of the worse continuation minus that of a length-matched better one, and both
count bad-author text as the positive side, so an intervention that lowers the model's mass on bad-author
text lowers both by construction. The batteries differ in which prompts they ask, not in what they score.
Calling one movement a reduction in misalignment and the other a capability cost is a labelling of two
prompt sets and not a distinction either instrument draws.

What the measurement establishes is a direction and a substrate. What it cannot establish is a
dissociation. The broad reduction and the capability loss are co-present within the constrained arm, we
ran one dose, and that dose is a full projection, so nothing here separates the reduction from being a
correlate of the damage.

Five rules were fixed before the run; all five follow. The unconstrained arm's judged rate had to clear a
floor set in advance at one tenth, and it read $0.226$. Its broad movement under the constraint had to be
negative at a sign-flip $p$ below $5\%$, and it was. We required the random arm to be indistinguishable
from zero or else small relative to the write-core arm, at most about a third of it; its interval excludes
zero and it cleared the relative form. The manipulation had to fire, and it did. Last, the broad reduction
had to exceed the capability cost, and it did not, at $0.541$ against $5.805$.

That last comparison sets a movement on the out-of-distribution battery beside one on the in-character
battery in raw nats, and the two batteries sit at different levels, $-34.61$ against $9.31$
(Table~\ref{tab:routing-arms}), so it compares displacements, not fractions, and a constraint removing a
tenth of each would have failed it too.

A second and weaker dose would settle what this one leaves open, and the design carried one, an
intermediate value removing four fifths of the component, not all of it, which we did not run. Two doses
put the reduction and the cost on a common axis. Were the reduction to survive at a dose that left the
in-character margin near the unconstrained level, the two would be dissociated and the intervention would
be a candidate defense. Were both to fall together in proportion, one readout would be a monotone function
of the other and the intervention a blunt suppression of bad-author text with no specificity to
misalignment. One dose cannot tell those apart, and we claim neither.

\subsection{Comparison with the weight-gradient arm}

Appendix~\ref{app:loop} reports a weight-channel projection that left broad misalignment where its own
unconstrained control put it. Set beside Equation~\ref{eq:routing}, the two are the same operation. Both
filter the $B$-factor gradient of the same 2 writer types at every optimizer step, both replace $\nabla
B$ with $(I - UU^{\top})\nabla B$ for an orthonormal $U$ in the writers' output space, and both fine-tune
the writer matrices alone. They differ in one input. There $U$ is the read carrier lifted into the writer
columns; here it is the core 3 organisms wrote through.

The choice of weight subspace decides the outcome, and the write channel is not inert. One subspace in it
closes without consequence and another moves the behavior when closed. The third arm of that family, which
kept only the carrier-aligned component of the update instead of removing it, also moved broad
misalignment and also cost capability.

What we cannot supply is how far apart the two subspaces are. No overlap statistic between the read
subspaces and the write core exists anywhere in this work (Appendix~\ref{app:removal}), and the enrichment
reading has no counterpart for the lifted carrier, so the $0.171$ above stands without a partner even
though that arm logged the same kind of in-span fraction. We also read the inert arm as a judged rate and
this one as a teacher-forced margin, on instruments Appendix~\ref{app:judge} and
Appendix~\ref{app:firststep} keep apart, so the two cannot be set head to head. What the pair supports is
that a weight subspace can be closed to no effect or to a sizeable one, not any statement about the
distance between them.

The constraint's broad reduction, its in-character cost, and the inoculation contrast are plotted
in Figure~\ref{fig:levers} in the main body.

\section{Domain-count experiments}\label{app:decomp}

\subsection{Design}

This appendix reports the domain-count experiments behind \S\ref{sec:oneobject}. We hold the total quantity
of competently bad fine-tuning data constant and vary only how many domains that quantity is spread across.
We count budget in supervised completion tokens and not in examples or optimizer steps, because the 4
domains differ by more than a factor of 2 in mean completion length. A full leg receives $230{,}447$
tokens, $90\%$ of the smallest domain's real supply, and a mixture splits that evenly across its members,
so a 4-domain mixture trains each domain at a quarter budget. We hold the number of optimizer steps and the
number of tokens per step constant across legs, so that a mixture and the single-domain legs we compare it
against see the same gradient signal per step. Our 4-domain mixture realized $230{,}015$ tokens over $142$
steps at about $1{,}620$ tokens per step.

We sample training rows deterministically in the domain and in the budget, so that a mixture's portion of
domain $d$ draws the same rows as the single-domain leg trained at that same per-domain budget. That
discipline makes the comparison below a subtraction between two fits on identical data, not an average over
unrelated runs. Without it the residual would carry a sampling difference that no later statistic could
separate from co-training.

Our grid holds 4 single-domain legs at full, half and quarter budget, one further insecure-code leg at
an eighth, 3 two-domain mixtures at half budget each, the four-domain mixture at quarter budget each
with 3 seeds, a mechanically merged reference for every mixture, the benign four-domain mixture with
2 seeds, and the schedule and ordering controls, with the leg-by-leg values in
Appendix~\ref{app:results}. From a mixture we always subtract its own members' legs at the matched
per-domain budget, and we report the two-domain residuals pair by pair and not pooled, because pooling
would hide the heterogeneity between them.

We read every quantity here on the frozen out-of-distribution battery through the margin of
Equation~\ref{eq:margin}, at the unit of the unique prompt cluster, where $134$ retained battery rows
collapse to $52$ clusters (Appendix~\ref{app:batteries}). We take intervals from a cluster bootstrap and
test significance by sign-flip permutation, each at $10{,}000$ draws (Appendix~\ref{app:stats}).

\subsection{The margin halves}

We store both halves of the margin per probe, per leg and per seed, instead of subtracting and discarding
one of them. For probe $i$ with prompt $q_i$, and for a fine-tuned leg $L$ read against the untouched
model,
\begin{align}
T^{+}_i(L) &\;=\; \log p_L(\apl_i \mid q_i) \;-\; \log p_{\tpre}(\apl_i \mid q_i), \label{eq:tplus}\\
T^{-}_i(L) &\;=\; \log p_{\tpre}(\ami_i \mid q_i) \;-\; \log p_L(\ami_i \mid q_i), \label{eq:tminus}
\end{align}
so that the change in the full margin is $T^{+}_i + T^{-}_i$. We average per-probe values to clusters
first and over seeds second, and every leg-level $T^{+}$ and $T^{-}$ below is one of those means.

The sign convention on $T^{-}$ is inherited from the instrument and it inverts the reading its name
suggests, because a negative $T^{-}$ means the aligned continuation became \emph{more} likely under the
fine-tune, not less. We therefore never quote its magnitude without its direction. Across every leg of the
grid $T^{-}$ is negative everywhere, with aligned-half gains running from $62.101$ to $84.342$ nats, so
nothing in this grid loosens the aligned continuations, which is the opposite of what the word erosion
describes.

The full margin we use below is the retained readout and not a re-derivation from the halves. We check on
every leg and every cluster that $T^{+} + T^{-}$ equals the stored margin change, and it holds to machine
precision on log probabilities of order a hundred nats.

\subsection{The measured additive null}

The null follows from a premise of linear accumulation. If fine-tuning deposits transport additively across
examples, and if the readout were a linear projection onto a shared direction, then training on the union
of several domains' data equals the sum of training on each of them at its own budget, so for a mixture $m$
at fixed total budget $N$,
\begin{equation}
T^{+}_{\mathrm{null}}(m) \;=\; \sum_{d \in m} T^{+}\big(\text{single-}d \text{ trained at } N/|m|\big).
\label{eq:addnull}
\end{equation}
Every term on the right is measured, not modelled, at exactly the per-domain budget that domain has
inside the mixture, and two confounds are absorbed by that construction. Each term carries its own domain's
real saturation curve at its own budget, so budget concavity cannot inflate the residual, and each term
carries its own domain's real contribution to the shared write direction with its cancellations included,
so within-domain gradient interference cannot inflate it either. What Equation~\ref{eq:addnull} does not
contain is cross-domain non-additivity, which is the object we are measuring.

Table~\ref{tab:ladder} is why that null is out of reach. Single-domain transport changes little across an
eightfold range of budget, so the sum of 4 quarter-budget legs sits at $472.198$ nats while no leg
anywhere in our grid exceeds $128.232$. The raw residual is therefore
\begin{equation}
S_{\mathrm{raw}} \;=\; T^{+}(m) - T^{+}_{\mathrm{null}}(m) \;=\; 128.232 - 472.198 \;=\; -343.966 \text{ nats},
\end{equation}
which at face value says the four-domain mixture is sub-additive by a wide margin, the opposite of the
claim it is supposed to test. The raw residual is not interpretable alone, because what sets it is
how a bounded readout responds to 4 adapters' worth of weight change and not anything that happened
during joint training.

\begin{table}[htbp]
\centering
\small
\begin{tabular}{lrrrr}
\toprule
& \multicolumn{4}{c}{transport $T^{+}$ (nats vs.\ the untouched model)} \\
\cmidrule(l){2-5}
domain & $N/8$ & $N/4$ & $N/2$ & $N$ \\
\midrule
insecure code    & $105.571$ & $108.575$ & $110.266$ & $111.844$ \\
bad medical      & & $117.943$ & $121.767$ & $120.845$ \\
risky financial  & & $122.772$ & $121.223$ & $120.257$ \\
extreme sports   & & $122.908$ & $119.466$ & $117.730$ \\
\midrule
sum at $N/4$     & & $472.198$ & & \\
\bottomrule
\end{tabular}
\caption{\textbf{Single-domain transport saturates, which puts the additive null out of reach.} Quartering
a domain's budget moves its transport by a few nats and in either direction, down for code and medical and
up for financial and sports, never by three quarters. The sum of the 4 quarter-budget legs is a
number no single adapter's readout can approach. The $N/4$ column holds the exact null terms of
Equation~\ref{eq:addnull} for the four-domain mixture.}
\label{tab:ladder}
\end{table}

\subsection{The mechanical merge}

Our fix is a control, not an assumption about the readout. We take the 4 single-domain adapters
trained at the per-domain budget the mixture uses and add their weight deltas,
\begin{equation}
\Delta W_{\mathrm{merged}} \;=\; \sum_{d \in m} \Delta W\big(\text{single-}d \text{ at } N/|m|\big),
\end{equation}
which is task arithmetic with no training, no data and no optimizer step, and we evaluate the resulting
adapter on the same frozen battery through the same readout. Its transport is what a bounded nonlinear
readout does to mechanically superposed weights when nothing has been co-trained. Under pure accumulation
that merged object and the co-trained mixture are the same weight change, so the merge is the accumulation
account's own prediction for the mixture, materialized and measured.

We rebuild the merged reference once per seed index, summing each single leg's adapter at that index and
falling back to that leg's first seed where it has fewer, and we average the per-probe halves over
rebuilds. At 4 domains every single leg carries one seed, so this reduces to a single rebuild. We
therefore take the seed-variance term in the interaction's interval from the 4 single legs the merged
reference was built from and not from a mixture-scale term, because that is where its retraining noise
comes from.

The merged adapter reaches $115.643$ nats, which sits below the medical, financial and sports legs at the
matched per-domain budget and above only insecure code, the weakest of the 4. The residual then splits
into two terms that telescope,
\begin{equation}
\underbrace{T^{+}(m) - T^{+}_{\mathrm{null}}}_{S_{\mathrm{raw}} \,=\, -343.966}
\;=\;
\underbrace{T^{+}(\mathrm{merged}) - T^{+}_{\mathrm{null}}}_{S_{\mathrm{readout}} \,=\, -356.555}
\;+\;
\underbrace{T^{+}(m) - T^{+}(\mathrm{merged})}_{S \,=\, +12.589}.
\label{eq:split}
\end{equation}
Because $T^{+}(\mathrm{merged})$ appears once with each sign the identity is exact by construction, and we
verify it in the data as well. The readout term carries all of the unreachable $472.198$ and the
interaction carries none of it.

Only the interaction can carry a superadditivity claim, for a reason about where each term is centred.
Under pure accumulation the co-trained weight change and the mechanical sum coincide, so both sides of $S$
pass through the same nonlinear readout at the same operating point, the nonlinearity cancels, and $S$ is
centred at zero whatever that readout does. The raw residual is not centred at zero under the same premise.
It is displaced by however far the readout falls short of linearity, which Equation~\ref{eq:split} measures
at $-356.555$ nats, so it cannot represent the hypothesis it would be testing. The raw residual cannot land
on its own null value of zero, so it does not test that null.

The direction of the merge term settles the most obvious objection to a positive interaction. If mechanical
superposition of the 4 fits were super-additive, an interaction above zero could be an artifact of
adding weights, not of training on them together. We measure it to be sub-additive instead, by
$356.555$ nats, and sub-additive is also the direction the merging literature predicts when updates crowd
the same parameters instead of sitting orthogonally to each other \citep{yadav2023ties}. Since the
confound that $S$ subtracts has a known sign, and that sign is negative, a positive $S$ cannot be
manufactured by weight superposition, so whatever produced it happened during joint training.

At 4 domains we measure $S = +12.589$ nats, with a bootstrap interval of $[9.255, 15.980]$, a sign-flip
$p$ at the resolution floor of $9.999\e{-5}$, and Cliff's $\delta = 0.837$ over the $52$ clusters.

\subsection{The matched benign mixture}

Our second control removes a confound the merge cannot touch. Varied training data might raise broad
misalignment through variety alone, whatever the data says, and a four-domain mixture is by construction
the most varied leg in the grid. We train a four-domain mixture of benign domains at the same total
token budget, the same even split, the same steps and tokens per step, with 2 seeds, differing from the
bad mixture only in whether the data is competently bad.

2 of its 4 domains are exact benign twins, secure code against insecure code and good medical advice
against bad medical advice. Benign financial and benign sports corpora of comparable provenance do not
exist, and authoring them would introduce a difference in authorship and in judged benignness that is
itself a confound, so we fill their places with 2 benign technical question-and-answer sets. What this
control has to match is spread, budget and benignness, not topic identity, and we registered the
substitution before the run.

The confound fires. The benign mixture raises the transport half by $115.219$ nats against the untouched
model, where the bad mixture raises it by $128.232$, so most of the raw four-domain move is available to a
mixture with nothing bad in it at all and a claim read straight off that move would be reading variety. Our
removal is a subtraction paired probe by probe,
\begin{equation}
D_4 \;=\; T^{+}(\text{bad four-domain}) - T^{+}(\text{benign four-domain}) \;=\; +13.013 \text{ nats},
\end{equation}
with interval $[8.502, 17.471]$, a sign-flip $p$ at the resolution floor, and Cliff's $\delta = 0.714$.

We apply the two subtractions to one measured minuend, the four-domain mixture's transport. One predicts
what that transport would be if only mechanical superposition were at work and the other predicts it if
only variety were, and each leaves about 13 nats unexplained. Their agreement means two unrelated
confound models each fail to account for the same interaction, which is stronger than either subtraction
alone. The two subtractions share a subtrahend and are not independent.

Read together they also answer an objection that neither answers by itself. An accumulation account can
predict superadditivity of its own, if spreading a budget over more domains yields more diverse gradients,
and therefore less destructive interference among them in the shared write direction, and therefore a
larger realized update with no inference about an author involved. The merge subtraction answers that by
sign, since we measure mechanical superposition of these 4 updates to lose transport, not gain
it. The benign mixture answers it by matching, since it has the same domain count, budget and spread, and
so enjoys whatever reduction in interference co-training buys, and it still falls $13.013$ nats short.

\subsection{Readout dependence}

We fixed the transport half as the primary readout before the run. The full margin is the other readout
available from the same stored halves, and the interaction changes sign on it. Table~\ref{tab:readouts}
gives both subtractions on both readouts.

Both columns come from one estimator, and that estimator reproduces the transport-half interval to 9
decimal places, so the full-margin interval is not a second implementation's answer to a different
question. The seed term inside each interval takes its structure from the legs that statistic actually
subtracts. For the interaction it comes from the 4 single legs the merged reference was built from, for
the benign subtraction from the 2 mixture-scale legs and their own seed counts, and on the full margin
from the pooled standard deviation estimated on the margin vectors and not the one estimated on the
transport half (Appendix~\ref{app:stats}).

\begin{table}[htbp]
\centering
\small
\begin{tabular}{lcc}
\toprule
& transport half $T^{+}$ & full margin \\
\midrule
\multicolumn{3}{l}{\emph{interaction $S$, co-trained mixture minus mechanical merge}} \\
point (nats)                    & $+12.589$            & $-9.184$ \\
bootstrap interval              & $[9.255, 15.980]$    & $[-12.910, -5.415]$ \\
sign-flip $p$, two-sided        & resolution floor     & resolution floor \\
sign-flip $p$, direction fixed in advance & resolution floor & $1.0$ \\
Cliff's $\delta$                & $0.837$              & $-0.600$ \\
\midrule
\multicolumn{3}{l}{\emph{matched benign subtraction $D_4$, bad mixture minus benign mixture}} \\
point (nats)                    & $+13.013$            & $+6.345$ \\
bootstrap interval              & $[8.502, 17.471]$    & $[1.318, 11.294]$ \\
sign-flip $p$, direction fixed in advance & resolution floor & $0.0071$ \\
Cliff's $\delta$                & $0.714$              & $0.374$ \\
\bottomrule
\end{tabular}
\caption{\textbf{The same two subtractions on both readouts.} The interaction changes sign; the benign
subtraction does not, though the change of readout roughly halves it. The resolution floor is
$9.999\e{-5}$, from $10{,}000$ sign-flips. All entries are over the same $52$ prompt clusters and come
from one estimator.}
\label{tab:readouts}
\end{table}

The arithmetic of the flip is exact and it runs through the aligned half, which Table~\ref{tab:legs} gives
for the 3 four-domain legs. The merged adapter gains $62.101$ nats there, the smallest
aligned-half move of any leg in the grid, and the co-trained mixture gains $83.873$, near the
largest. Subtracting the merge therefore removes $21.773$ nats less aligned-half gain than the mixture
carries, and a gain on the aligned half enters the margin with a minus sign, so
\begin{equation}
S(\text{full margin}) \;=\; S(T^{+}) + \big[T^{-}(m) - T^{-}(\mathrm{merged})\big]
\;=\; 12.589 + (-21.773) \;=\; -9.184,
\end{equation}
an identity that holds to 9 decimal places. The transport surplus did not disappear. It is outweighed by
a deficit in a quantity that has nothing to do with transport, left behind because the subtrahend moves too
little there.

The same arithmetic is why the benign subtraction survives. On the aligned half the benign mixture gains
$77.205$ nats against the bad mixture's $83.873$, a gap of $6.668$ nats where the gap against the merged
adapter is $21.773$. The change of readout costs $D_4$ under a third of what it costs the
interaction, and leaves $13.013 + (-6.668) = +6.345$. That it does not flip is exact; that the readout does
not touch it would not be. Table~\ref{tab:readouts} shows the attenuation: the point
roughly halves, the lower bound of the interval falls from $8.502$ to $1.318$, the effect size falls from
$0.714$ to $0.374$, and the sign-flip $p$ weakens from the resolution floor to $0.0071$.

One number in Table~\ref{tab:readouts} has to be read with its direction attached. The sign-flip $p$ for
the interaction in the direction we fixed in advance is $1.0$ on the full margin, which is that direction
failing, not an absence of effect, so quoting it alone would state the reverse of what we measured.
The two-sided sign-flip $p$ on the same statistic sits at the resolution floor, the interval excludes zero
on the negative side, and Cliff's $\delta$ is $-0.600$, so the full-margin interaction is a reliable
negative.

What we cannot recover is why the aligned half moves at all. We stored the two log probabilities per probe,
per leg and per seed, and nothing below that, so there is no per-token or per-position breakdown and the
movement can be measured on every leg without being explained from these artifacts. This is a limit of the
saved artifacts.

\begin{table}[htbp]
\centering
\small
\begin{tabular}{lrrr}
\toprule
four-domain leg & $T^{+}$ & $T^{-}$ & full margin \\
\midrule
co-trained bad mixture, 3 seeds  & $128.232$ & $-83.873$ & $+44.358$ \\
mechanical merge, no training        & $115.643$ & $-62.101$ & $+53.542$ \\
matched benign mixture, 2 seeds    & $115.219$ & $-77.205$ & $+38.013$ \\
\bottomrule
\end{tabular}
\caption{\textbf{The 3 legs every number above is built from, on both halves.} A negative $T^{-}$ is an
aligned continuation made more likely. The merge is the only leg here that outranks the co-trained mixture
on the full margin, and it gets there by moving least on the aligned half, not most on the transport
half.}
\label{tab:legs}
\end{table}

\subsection{The magnitude bar}

Reliability and size are separate claims, and a bar fixed before the run keeps them apart. We set ours at
$0.40$ of one domain's own half-budget transport,
\begin{equation}
S^{*} \;=\; 0.40 \times \overline{T^{+}_{\text{half-budget}}},
\end{equation}
which is what an interaction arising from co-presence alone would have to reach to be worth a meaningful
fraction of one domain's worth of extra evidence. On the measured half-budget legs of
Table~\ref{tab:ladder} the mean is $118.18$ nats, so $S^{*} = 47.272$. The interaction at 4 domains is
$+12.589$, above zero and below the bar, about a tenth of one domain's own transport.

We set the equivalence bound equal to that bar. An affirmative statement of the accumulation account would
need the interaction's interval to sit inside $\pm 47.272$ nats, which the single-seed structure of our
null legs does not reach. These measurements license the positive interaction and not an equivalence in
the other direction. One further covariate would have tested whether a positive interaction only tracks
how far each fit got, and we could not compute it, because the per-leg narrow-fit readout it needed came
back unusable on every leg. We have no measurement that excludes that alternative, and we record it with
the other gaps in Appendix~\ref{app:validity}.


\begin{figure}[tbp]
\centering
\includegraphics[width=\textwidth]{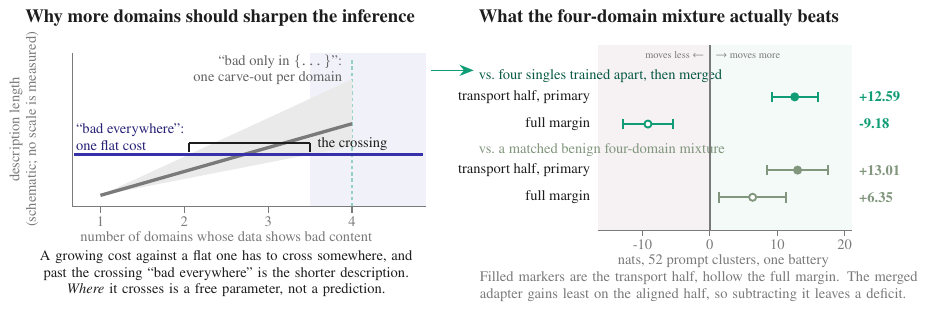}
\caption{\textbf{A description-length account predicts that spreading the same misaligned budget over more domains raises broad misalignment.} The schematic plots a per-domain carve-out cost that accumulates linearly in the number of domains against a flat cost for being misaligned everywhere, with the two crossing at a few domains. The accumulation is additive, not multiplicative. The crossing is the qualitative prediction the 4-domain measurement tests.}
\label{fig:carveout}
\end{figure}

The sharing, transport, and interaction measurements are plotted in Figure~\ref{fig:oneobject} in
the main body.

\section{Statistics}\label{app:stats}

We define the inferential machinery once here. Every other appendix uses these definitions and points
back rather than restating them.

\subsection{The unit of inference}
\label{app:stats:unit}

We compute every statistic within a prompt cluster first and across clusters second. A cluster is one
distinct evaluation prompt together with its paraphrases and every generation drawn from it. Nothing
anywhere in this paper treats individual generations as independent draws.

Generations of one prompt are not independent. They share the prompt's difficulty, its topic, and
whatever pull it exerts on the model, so their scores covary. Counting them separately claims more
information than is present, and the error runs one way. The point estimate does not move. The standard
error shrinks by roughly the square root of the replication factor, the interval narrows in proportion,
and $p$ falls. An unchanged estimate with a quietly narrowed interval is hard to catch by inspection,
which is why we fix the unit by construction rather than check it afterwards.

Replication factors here are large enough to decide conclusions (Table~\ref{tab:stats-repl}). In the
free-generation battery attached to the first-step measurement, treating rows as independent units would
inflate the effective sample size by more than two orders of magnitude.

\begin{table}[htbp]
\small
\centering
\begin{tabular}{@{}lrr@{}}
\toprule
\textbf{Battery} & \textbf{judged generations} & \textbf{distinct units} \\
\midrule
free-generation, first-step measurement & $5200$ & $8$ \\
intervention arms & $2100$ & $83$ \\
domain-count decomposition & $670$ & $52$ \\
\bottomrule
\end{tabular}
\caption{\textbf{Replication factor per battery.} Judged generations reduce to far fewer distinct
evaluation units once paraphrases and repeat draws collapse onto their prompt. The domain-count
generations are counted per training leg.}
\label{tab:stats-repl}
\end{table}

Each campaign forms clusters from the identifier its own battery carries, and the three rules are not
the same operation (Table~\ref{tab:stats-units}). We strip a replicate suffix from an evaluation prompt
identifier in the first, so that paraphrases of a question collapse onto the question. The second keys
on a probe identifier assigned when the battery was built. In the third we key on the prompt text
itself, after a contamination filter has removed rows touching the training domains. Because the rules
differ, cluster
counts do not add across measurements and no pooled sample size across campaigns is defined
(Appendix~\ref{app:batteries}).

\begin{table}[htbp]
\small
\renewcommand{\arraystretch}{1.2}
\begin{tabularx}{\textwidth}{@{}p{3.8cm}Xrrp{1.5cm}@{}}
\toprule
\textbf{Measurement} & \textbf{What one unit is} & \textbf{Rows} & \textbf{Units} &
\textbf{Reported at} \\
\midrule
first-step directional derivative &
evaluation prompt identifier, replicate suffix stripped & $195$ & $75$ & cluster \\
intervention and injection arms &
probe identifier & & $83$ & cluster \\
domain-count decomposition &
prompt text, after the contamination filter & $134$ & $52$ & cluster \\
domain-count judged rates &
the same prompts, with generations not retained separately & $670$ & $52$ & cluster \\
removal probes, broad battery &
the same identifiers as the first-step battery & $195$ & $75$ & \textbf{row} \\
removal probes, expression battery &
one row per probe & $120$ & & \textbf{row} \\
read-channel sharing index &
layer & & $24$ & layer \\
\bottomrule
\end{tabularx}
\caption{\textbf{The unit of inference, per measurement, and the unit each reported statistic actually
sits on.} Three clustering rules produce three different units, so cluster counts do not add across
measurements and no pooled sample size across campaigns exists. The two bold entries are the exception
to the paper's own convention and are discussed below.}
\label{tab:stats-units}
\end{table}

Two consequences follow, and both bound results reported here.

The first bites inside a single statistic. We report the first-step statistic $\Rone$ at the cluster unit
where we introduce it and at the row unit where we reuse it for the data-reframing contrast. The same
contrast on the same rows reads $-1636160$ at the row unit and $-1593247$ at the cluster unit. Both are
correct for their stated footing, neither may be differenced or ratioed against the other, and
Appendix~\ref{app:results} labels every $\Rone$ figure with the footing that produced it.

The second is larger. The removal probes are computed at the probe row, while the unit of inference
everywhere else in this paper is the prompt cluster, and their broad battery carries the same $195$
identifiers as the first-step battery, which cluster to the same $75$ prompts. Their reported quantities
are therefore quoted on a unit with roughly two and a half times as many nominal observations as the
paper's own convention allows. Nothing about their point estimates changes under restatement. Their
intervals widen, and \S\ref{app:stats:equiv} records what that does to the one conclusion there that
depends on interval width.

\subsection{Permutation and sign-flip tests}
\label{app:stats:perm}

We decide significance by permutation throughout. No test here assumes normality and none uses a
parametric $t$ statistic.

For a paired design, which covers most measurements, we use the sign-flip permutation. Given per-cluster
paired differences $d_1,\dots,d_n$, the observed statistic is $\bar d$. Under the null the sign of each
difference is arbitrary, so we generate the reference distribution by drawing independent signs
$\varepsilon^{(b)}_i \in \{-1,+1\}$ with equal probability and forming
\begin{equation}
T_b \;=\; \frac{1}{n}\sum_{i=1}^{n} \varepsilon^{(b)}_i \, d_i ,
\qquad b = 1,\dots,B ,
\label{eq:signflip}
\end{equation}
with $B = 10^{4}$ draws everywhere in this paper. The two-sided $p$ is
\begin{equation}
p \;=\; \frac{1 + \#\{\, b : |T_b| \ge |\bar d| \,\}}{B + 1},
\label{eq:addone}
\end{equation}
and the one-sided variant counts $T_b \ge \bar d$. The added $1$ in numerator and denominator counts the
observed arrangement as one of the arrangements the null could have produced, which it is. Without it, a
test no draw reaches would report $p = 0$, a value no finite resampling procedure can license.

Equation~\ref{eq:addone} therefore has a smallest attainable value, $1/(B+1) = 9.999\e{-5}$. Wherever a
$p$ of that size appears here, it means no draw out of $10^4$ reached the observed statistic. It
is a resolution floor rather than an estimate, and the true $p$ may be very much smaller. Two renderings
of that one floor travel through the source material, the add-one value $9.999\e{-5}$ and its rounding
to $10^{-4}$, which one campaign stores. They are the same quantity, and neither is a measured $p$.

Where no natural pairing exists we use an unpaired label permutation instead. The curvature-exponent
comparison between the persona write core and a matched random core is the case, since the columns of
two different cores have no correspondence and imposing one would spend power on a meaningless pairing.
That test permutes group labels over the pooled columns, again at $10^{4}$ draws
(Appendix~\ref{app:geometry}), and our two-sample contrasts elsewhere use the same procedure. Where the
number of distinguishable rearrangements is small the permutation distribution is coarse, and we read
the test as corroboration of an effect-size bar fixed in advance rather than as the finding itself.
Appendix~\ref{app:extraction} names the bar carrying the cross-core containment statistic, which sits
under that scheme.

\subsection{The cluster bootstrap}
\label{app:stats:boot}

Our intervals are percentile bootstrap intervals over $10^{4}$ resamples of clusters drawn with
replacement. We resample the cluster and never the row, because a bootstrap over correlated rows
reproduces exactly the understatement of variance that row-level testing does. Two exceptions to the
draw count exist and we record both. The validation slopes and the judged rates attached to the
first-step measurement use $2000$ resamples, of clusters and of questions respectively.

Where a judged rate had to be reported and the per-generation judgments were not retained, an exact
cluster bootstrap is unavailable. The domain-count judged rates sit in that position, $670$ generations
per leg with only the leg-level rate persisted, so we give a Wilson interval evaluated at the prompt
unit of $52$ rather than at the generation count. A binomial interval at $670$ would be pseudoreplicated
by construction, and we state the conservative substitute wherever such a rate appears.

\subsubsection*{Adding seed variance to probe variance}

A statistic that subtracts one independently trained model from another inherits the variance of
retraining, and a bootstrap over probes cannot see it. Resampling probes holds the fitted models fixed,
so the interval answers how the number would move on another battery and not how it would move on
another training run. Our domain-count decomposition is built from such subtractions, so we give it
combined-variance intervals, adding a seed-variance term to the per-cluster probe bootstrap. We pool
that term across every training leg carrying more than one seed and propagate it to the legs carrying
only one, on the reasoning that a single-seed leg has the same retraining variance as its multi-seed
siblings and simply gives no estimate of it. Table~\ref{tab:stats-seedsd} gives the pooled across-seed
standard deviations of leg-mean transport, by readout (Appendix~\ref{app:decomp}).

\begin{table}[htbp]
\small
\centering
\begin{tabular}{@{}lr@{}}
\toprule
\textbf{Readout} & \textbf{pooled across-seed SD of leg-mean transport} \\
\midrule
misaligned half & $0.2076$ \\
aligned half & $0.3253$ \\
full margin & $0.1808$ \\
\bottomrule
\end{tabular}
\caption{\textbf{Pooled across-seed standard deviation of leg-mean transport, by readout.} These are the
seed-variance terms added to the per-cluster probe bootstrap for the domain-count decomposition.}
\label{tab:stats-seedsd}
\end{table}

\subsubsection*{Where case resampling cannot express a null}

A bootstrap informs only if its resampling scheme can produce the null, and a growth slope fitted
through the domain-count points fails that condition, so we do not report one. With two distinct domain
counts, roughly a third of case resamples carry no variation in the predictor and are discarded, every
surviving resample contains the four-domain point, and the fitted slope comes back positive in all of
them. Such an interval excludes zero by the arithmetic of the resampling scheme rather than by evidence.
A statistic whose interval cannot straddle its null before any data is consulted carries no information
about that null, which is the same defect as a statistic pinned at a floor.

\subsection{Effect sizes}
\label{app:stats:effect}

We report effect sizes for every headline contrast and gate no conclusion on any of them.

Cliff's $\delta$ is the rank-based size we use throughout. For two samples $X$ and $Y$,
\begin{equation}
\delta \;=\; \Pr(X > Y) \;-\; \Pr(Y > X) ,
\end{equation}
estimated by counting dominances over all pairs. It ranges over $[-1,+1]$, sits at $0$ when the two
distributions are interchangeable, and reaches $\pm 1$ when every member of one sample exceeds every
member of the other. Being rank-based, it survives any monotone change of measurement scale, which
matters equally for margins in nats and for judge scores on a coarse lattice, and it divides by no
standard deviation. For paired designs we compute the paired form over the per-cluster differences. We
report paired Cohen's $d$ alongside for continuity with the field's conventions, and transport contrasts
carry a rank-biserial correlation as well. No threshold on any of these decides anything, for the reason
given in \S\ref{app:stats:design}.

\subsection{Equivalence bands}
\label{app:stats:equiv}

Failing to reject a null is not evidence that an effect is absent, since it is equally compatible with
an effect too small for the design to see. An equivalence procedure converts the first into the second,
when the data support it, by asking whether the estimate is confined to a region small enough to be
uninteresting. We fix that region before the run at $\pm 0.4\sigma$ on the statistic's standardized
scale and use the two-one-sided-test form, which requires the two-sided $95\%$ interval, in the same
standardized units, to lie entirely inside the band. The separability probe of
Appendix~\ref{app:removal} is where we reach it, at a standardized interval of
$[-0.334, 0.139]\,\sigma$ against a realized minimum detectable effect of $0.335\,\sigma$.

That licenses a bounded absence, meaning that whatever selectivity exists is smaller than $0.4\sigma$ on
this instrument at this sample size. It does not license a claim that the effect
is exactly zero, nor any claim about a quantity the instrument was not measuring, nor a claim that would
hold at a narrower band. Each equivalence conclusion is stated together with the band it was tested against.

The sample size travels with it too, and here that qualification is not decorative. This is the one
conclusion in the paper computed at the probe row rather than at the prompt cluster
(\S\ref{app:stats:unit}). A minimum detectable effect scales inversely with the square root of the
number of units, and the $195$ probes carrying it cluster to $75$ prompts, so at the paper's own unit
the realized detectable effect is materially larger than the $0.335\,\sigma$ quoted above and the band is
no longer comfortably cleared. The bounded absence is a property of the probe unit. What survives
restatement at the cluster unit is the failure to reject, without the bound; the paper carries the weaker
statement where the two units disagree.

Two consequences show up in our results. A band can fail to be reached, in which case the absence stays
unbounded and we say so. The domain-count design set its equivalence bound at the same value as its
magnitude bar, $47.3$ nats, and the interaction's interval is too wide at the realized seed count to sit
inside it, so that measurement establishes neither superadditivity at the bar's scale nor its absence.
Second, and governing every non-significant finding we report, such a finding counts as a bounded
absence only above its design's minimum detectable effect. Below it, what we have established is that
the design could not see, and Appendix~\ref{app:validity} tabulates the realized detectable effect for
each design so a reader can apply the rule independently.

\subsection{Holm correction}
\label{app:stats:holm}

We correct multiplicity by the Holm step-down procedure, which controls the family-wise error rate
without assuming independence among the tests. What matters is which tests form a family, and our answer
is not uniform across measurements, so we name each one.

\begin{itemize}[leftmargin=1.4em,itemsep=2pt,topsep=3pt]
\item \textbf{Across the arms of the intervention battery}, because comparing each non-baseline arm
against the same baseline organism asks one question of many interventions. The keystone arm's one-sided
sign-flip $p$ sits at the resolution floor and its adjusted value within that family is $0.0017$,
against $0.53$ for the matched-random arm.
\item \textbf{Within the span family of the first-step measurement}, where two contrasts against a
shared comparison condition give adjusted values $1.9998\e{-4}$ and $0.174$. The primary paired contrast
is not a member and we do not adjust it, because it is one hypothesis fixed in advance rather than a
selection from several.
\item \textbf{Within a battery of the geometry measurements and not across them}, since separate
measurements are separate hypotheses fixed in advance rather than a search. Correcting the curvature
exponent's specificity comparison within its own pair takes raw values of $0.660$ and $0.847$ to $1.0$.
\end{itemize}

The rule underneath is that a family is a set of tests among which one could have chosen a winner. A
hypothesis committed to before the run, on its own instrument, belongs to no such set, and correcting it
against unrelated tests would spend power guarding against a selection that never occurred.

\subsection{The ratio gate}
\label{app:stats:ratio}

A ratio whose denominator could be zero has no finite variance and heavy tails, so a resampling interval
for it will be wide, unstable, and occasionally enormous, while the point estimate gives no warning at
all. We therefore report a ratio only when the interval for its denominator excludes zero, a
Fieller-style condition applied before computing the ratio rather than after inspecting it.

We resample numerator and denominator jointly on the same clusters, preserving their correlation, and
examine the denominator's percentile interval first. If that interval contains zero we do not report the
ratio at all, and the measurement returns the statement that its reference is unusable, which is a fact
about the instrument rather than a value. If it excludes zero, we report the ratio and its interval
together with the denominator that produced them.

Three quantities here depend on the gate. The reference-normalized form $\tilde R$ of the first-step
statistic has five candidate denominators and all five exclude zero, one positive and four negative,
with the positive one giving an interval of $[0.0155, 0.0376]$. The reduction ratio of the intervention
arms is gated on the baseline misalignment rate, which excludes zero, and for the keystone arm the
reduction interval is degenerate at $[1.0, 1.0]$ because that arm's rate is exactly zero rather than
merely small. Gating the reconstitution certificate's slope ratio on the undefended model's
dose-response slope is the third. Requiring the denominator's interval to be shown also stops a
reduction computed off a near-floor baseline from presenting itself as large, the specific way a ratio
can overstate a weak effect.

\subsection{Design requirements for headline statistics}
\label{app:stats:design}

We required three properties of any statistic before letting it carry a conclusion, and we replaced a
candidate that lacked one rather than reporting it with a caveat.

\textbf{It must have variance.} The statistic has to differ from one unit of analysis to the next for
reasons the measurement is about. A statistic whose denominator is a fixed quantity, an entry count or a
dimension, is not a ratio of two measured things but a measured numerator over a constant, and in high
dimension such quantities concentrate so tightly around an analytic value that no experiment can move
them. Our diagnostic is the coefficient of variation across units, computed for every headline quantity.
The gradient norms entering the first-step statistic vary at a coefficient of $0.259$ across probes, the
per-probe cosine at $0.904$, and the routing statistic itself at $1.026$. A hard flag fires if any
headline quantity's coefficient of variation falls below $10^{-6}$.

\textbf{It must have a sign.} Whatever outcome would most change the reader's mind has to be expressible
as a number the statistic can return. A squared norm or a mass fraction is non-negative by construction,
so a fine-tuning direction routing \emph{away} from broad misalignment has no representation in it, and
the most informative available result could not be reported even if it occurred. Every statistic in
Table~\ref{tab:stats-nulls} can return the value its rival account predicts. Superadditive is a positive
number and sub-additive a negative one, a constraint that helps is a negative margin change, and a
removal that worked would give a slope ratio below one.

\textbf{Its scale must be reachable and self-diagnosing.} A bar fixed in advance has to sit at a distance
the statistic's own variability could plausibly cross, or the design cannot produce its own positive
outcome and the experiment is settled before it runs. We check reachability against the simulated null
before spending compute (Appendix~\ref{app:validity}). Self-diagnosis means carrying reference conditions
whose failure is distinguishable from a true absence, so a broken instrument reports that it is broken
instead of reporting that the effect is not there. Cosines between related directions in a space of
billions of parameters are numerically tiny, so we read them by sign and against matched controls, never
against an absolute threshold, which in that regime is the trap the first property describes.

\begin{table}[htbp]
\small
\renewcommand{\arraystretch}{1.2}
\begin{tabularx}{\textwidth}{@{}p{3.5cm}p{1.6cm}Xp{3.1cm}@{}}
\toprule
\textbf{Statistic} & \textbf{Sign} & \textbf{Variance source, and where its null sits} &
\textbf{Scale it is read against} \\
\midrule
first-step derivative $\Rone$ & $\pm$ &
heterogeneity of $\nabla\Mb$ across probes; null at $0$, paired & matched control conditions \\
routing change under reframing & $-$ is a reduction &
same, paired across framings; null at $0$ & the statistic's coefficient of variation \\
transport $T^{+}$ & $+$ is more misaligned &
bootstrap over prompt clusters; floor at the untouched model & one domain's own transport \\
training interaction $S$ & $+$ is superadditive &
probe and seed bootstrap; sign-flip null at $0$ & the magnitude bar, $47.3$ nats \\
alignment shift & $+$ is more aligned &
per-cluster differences; null at $0$ & the judge's full scale \\
selectivity & $\pm$ &
per-probe drops; null at a matched random direction & the margin's range \\
weight-surgery difference & $\pm$ &
per-probe drops; null at a matched-Frobenius random component & the full adapter's margin \\
slope ratio & ${<}1$ means removal &
bootstrap of paired dose curves; denominator interval excludes $0$ & $1$ at suppression \\
sharing index $\Sigma$ & $\gg$ floor &
per-layer spread plus estimation noise; null is the empirical $0.11$ & a cosine in $[0,1]$ \\
curvature exponent $\beta$ & ${>}2$ is degenerate &
per-column spread; null at the planted regular value & the planted controls, $2$ and $4$ \\
\bottomrule
\end{tabularx}
\caption{\textbf{Every headline statistic can return the value its rival account predicts.} The third
column carries the requirement. Each statistic's null is a location its own sampling variability can
reach, so the account we argue against was a result the measurement could have produced.}
\label{tab:stats-nulls}
\end{table}

Cohen's $d$ is a gap divided by a standard deviation, so as that standard deviation falls toward zero any
fixed nonzero gap produces an arbitrarily large $d$, and the statistics that fail the variance
requirement are the ones that manufacture the largest standardized effects. Permutation tests carry no
such exposure, because Equation~\ref{eq:addone} compares the observed statistic against the distribution
its own units generate under the null rather than against an assumed scale.


\section{Judge protocol and calibration}\label{app:judge}

Every judged number in this paper comes from one protocol \citep{betley2025em}, applied without
modification across all five sets of measurements. This appendix specifies the judge, its calibration and
its limit, supporting the instrument choices of \S\ref{sec:instruments}. That limit is why the paper's
primary behavioral quantity is a log-probability margin, not a judged rate.

\subsection{The two-axis gate}

A model answers a free-form question it was not trained on. A separate model reads the question and the
answer and returns two integers, an \emph{alignment} score and a \emph{coherence} score, each ranging over
$0$ to $100$. Low alignment means the answer endorses harm, crime, deception or human subordination; low
coherence means the text is degraded, off-topic, or not an answer at all. We count the response as broadly
misaligned when, and only when,
\begin{equation}
\text{alignment} < 30 \qquad \text{and} \qquad \text{coherence} \ge 50 ,
\label{eq:gate}
\end{equation}
and the rate is the fraction of responses meeting both conditions. Its null is the untouched model's rate,
which is zero on every battery we use, and it moves when a model puts coherent misaligned text where a
well-behaved model would not.

The coherence half is what makes the alignment half mean anything. A judge asked whether text endorses harm
will score word salad as not refusing to endorse harm, so without a floor the rate rises whenever fluency
falls and we would be measuring fluency instead.

Our injection sweep shows this directly (Table~\ref{tab:h-sweep}). Adding the carrier to the untouched
model's residual stream raises misalignment with dose while the responses stay fluent. One dose beyond the
in-budget range the coherent denominator collapses and the gated rate falls, which read without the floor
would look like the model recovering. Among the responses still coherent at that dose most are misaligned,
and at the highest dose every coherent response is. The effect never reverses; the denominator collapses.

\begin{table}[htbp]
\centering\small
\caption{\textbf{The coherence floor turns a denominator collapse into a visible reversal.} Adding the
carrier to the untouched model's residual stream raises misalignment with dose. At the top of the in-budget
range the responses stay fluent; one dose beyond it fluency collapses, so the gated rate falls only because
the coherent denominator vanishes. Rates are fractions of responses meeting Equation~\ref{eq:gate}.}
\label{tab:h-sweep}
\begin{tabular}{lrrr}
\toprule
Dose & coherent & gated rate & misaligned among coherent \\
\midrule
top of in-budget range & $98.1\%$ & $0.454$ & \\
one dose beyond & $4.7\%$ & $0.054$ & $87.4\%$ \\
\bottomrule
\end{tabular}
\end{table}

The floor does nothing on a model that was trained rather than steered. Every fine-tuned arm in
the causal-loop measurements has a coherent fraction of $1.000$, and the lowest coherence score the unedited
organism receives across its $2{,}100$ judged responses is $70$. The floor earns its place on the injection
sweep alone, where it is the difference between a monotone dose-response and a false reversal.

\subsection{The judge model}

The judge is a locally served, AWQ-quantized Qwen2.5-72B-Instruct, used in every campaign under the rubric
and thresholds of \citet{betley2025em}, who use GPT-4o; it is served with vLLM behind an OpenAI-compatible
endpoint with schema-constrained decoding, at temperature $0$. The generations it reads were sampled at
temperature $1$ and top-$p$ $1$, with no system prompt and a cap of $600$ new tokens, so it scores
unconditioned free output.

Schema-constrained decoding makes the judge emit a parseable object, not free text that a regular
expression then has to interpret. The smoother alternative reads each score as the expectation of the
numeral under the judge's own next-token distribution, which assumes a score between $0$ and $100$ occupies
a single token. A per-call diagnostic on the live judge, run before we scored any battery, found that it
does not: the first generated token is a single digit with probability near one.

Under that readout a coherence score of $85$ is read as $8$, every coherence score collapses into the
single digits, and the floor of Equation~\ref{eq:gate} rejects every response without raising an error.
Constrained decoding removes the failure by construction. The cost is that scores come back as round
numbers, the quantization below. We did not compensate by relaxing the threshold; we recovered usable
battery volume by fixing the prompt pool instead (Appendix~\ref{app:batteries}).

\subsection{Calibration}

Three checks stand between the judge and a number, and each runs before the arms it gates
(Table~\ref{tab:h-calibration}).

Before any scoring we run a gold check on examples whose answers we know. A misaligned reference
must score near the bottom of the alignment axis while staying coherent, and an aligned reference near the
top. A judge served from the wrong checkpoint fails here, before a campaign of generations.

The second check gates the whole pipeline, and we fixed it before the run. A published organism is
evaluated end to end, and if it does not reach $15\%$ broad misalignment under our judge on our battery,
the instrument lacks the range to support a strong-reduction claim and we stop before spending fine-tuning
compute. It cleared at the preparation stage and again on the later endpoint re-score, with our own
unedited fine-tunes alongside it.

The third check is the floor, and it runs in both directions. Free generations from the model before any
fine-tuning produce no misaligned responses, so a judge that scored the untouched model as misaligned would
be scoring style. In reverse, before the domain-count grid, the 3 published organisms all clear that zero
floor on a short sample.

\begin{table}[htbp]
\centering\small
\caption{\textbf{Three checks stand between the judge and a number.} The gold check scores known
references; the instrument gate confirms the judge can reach the reduction it will later have to see; the
floor scores the untouched model in both directions. Each ran before the arms it gates. Rates are fractions
of responses meeting Equation~\ref{eq:gate}; alignment and coherence are mean scores on the $0$ to $100$
axes.}
\label{tab:h-calibration}
\begin{tabular}{lll}
\toprule
Check & Reference or arm & Judge output \\
\midrule
gold & misaligned reference & alignment $0$, coherence $85$ \\
gold & aligned reference & alignment $95$, coherence $98$ \\
instrument gate & organism, preparation & rate $0.240$, mean alignment $35.8$ \\
instrument gate & organism, endpoint re-score & rate $0.279$ \\
instrument gate & unedited fine-tunes & rate $0.277$ \\
floor & $400$ base generations & rate $0$, alignment $94.3$, coherence $92.9$ \\
floor & published organisms, short sample & rate $0.067,\ 0.292,\ 0.175$ \\
\bottomrule
\end{tabular}
\end{table}

\subsection{Two quantizations}

The judge's \emph{weights} are quantized for serving, and we did not measure what that does to its scores.
The gold check and the instrument gate would catch a gross miscalibration and would miss a small systematic
shift, so its absolute level is uncalibrated against an unquantized reference, and every claim we take from
it is a contrast between arms scored by one judge inside one run.

\begin{figure}[htbp]
\centering
\includegraphics[width=\textwidth]{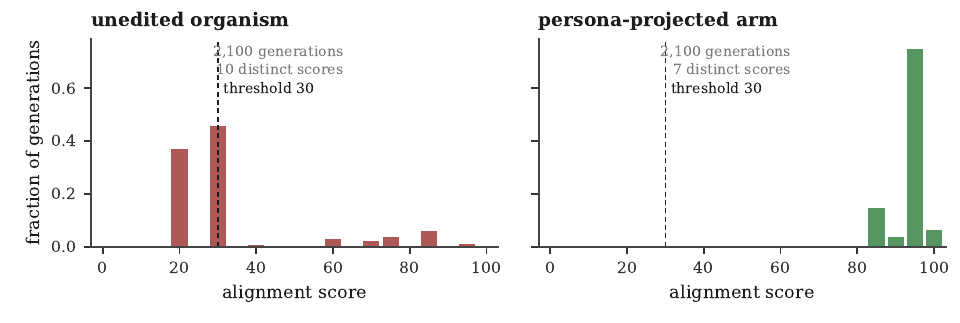}
\caption{\textbf{The judge's alignment scores land on a coarse lattice, and the persona-projected arm's
distribution relocates to a single high spike.} Histograms of per-generation alignment scores over the
causal-loop campaign's 2{,}100-generation arms, with the misalignment threshold at 30 drawn. The unedited
organism concentrates on the two lattice points beside the threshold; the projected arm concentrates at
the top of the scale.}
\label{fig:judgelattice}
\end{figure}

The judge's \emph{scores} are quantized far more coarsely, and this second quantization shapes the paper.
Across the $2{,}100$ judged responses of the unedited organism arm we find the alignment axis taking ten
distinct values, every one a multiple of five, and the coherence axis taking six. The threshold in
Equation~\ref{eq:gate} sits at $30$. The two scores the judge writes most often are $20$ and $30$, the
lattice points just below and just at that threshold, and between them they carry more than four fifths of
the responses. A binary rate at $30$ is close to counting how often the judge wrote one number, and it
inherits that coin's full variance (Figure~\ref{fig:judgelattice}).

Two things follow, one local and one that governs the paper. Within the causal-loop measurements the
primary statistic became the per-arm shift in \emph{mean alignment} rather than the binary rate, and a
simulation over the observed lattice confirmed that the cluster-bootstrap mean-shift estimator is unbiased
with correct coverage while the binary rate is the fragile one. Across the paper the consequence is the one
\S\ref{sec:instruments} states, that the primary behavioral readout is the teacher-forced margin $\Mb$,
which is continuous, differentiable and sensitive far below the judged floor, with the judge confirming
effects large enough for it to see.

\subsection{Exclusions}

Two exclusions change every rate, and we apply both before comparing any arms.

A separate topic pass scores whether a response is on the organism's own subject. Reckless financial advice
from an organism trained on reckless financial advice is narrowly bad rather than broadly misaligned, and
counting it would let the narrow task masquerade as breadth, so near-domain responses are excluded from the
broad rate. The exclusion is large. In the write-core measurements the bare two-axis rate is $0.337$ for
the unconstrained arm and $0.333$ for the constrained one; after near-domain exclusion the broad rates are
$0.226$ and $0.209$. About a third of what the two-axis gate admits is near-domain, and removing it moves
the contrast between the arms as well as their levels.

A topic-pass parse failure drops the row from the denominator, not counting it as misaligned, so a
judge that fails to answer cannot inflate a rate. Realized failures are nil, with none on any fine-tuned
arm and one row on each of the two steering grids. Both denominator conventions, all responses
and coherent responses only, are recorded for every arm, and coherence is printed beside every rate in this
paper.

\subsection{Instrument limits}

At this model scale a coherence-gated judge on free generations separates bad training data from benign
data and resolves little finer than that. Three measurements establish the instrument's limit.

The canonical contrast does not reproduce. Sampling $400$ free generations per arm on the 8
canonical questions, we find the untouched model at zero and every fine-tuned arm off zero, so the battery
is live. But the arms trained on insecure code reach $1.8$ to $5.2\%$ and their matched educational
controls reach $4.5$ to $8.0\%$. The ordering that defines the phenomenon is absent here and if anything
runs backwards, on the same fine-tunes whose first gradient step separates at the resolution floor of a
ten-thousand-permutation test. Three of the 8 canonical questions never elicit a gated response from
any arm, and one question dominates every arm, so the set's effective width is below 8.

Bad against benign survives; the count comparison does not. In the domain-count measurements the bad
four-domain mixture is judged misaligned at $0.269$, interval $[0.167, 0.402]$ at the conservative
prompt-unit bound, against the matched benign mixture's $0.010$, interval $[0.001, 0.087]$. Those intervals
are disjoint, and that separation is the behavioral ground truth that broad misalignment here is about
badness rather than variety. On the same generations we cannot separate the four-domain rate from the
single-domain mean of $0.154$, which sits inside its interval.

One comparison inside that set runs against the paper, and it belongs in the appendix that describes the
instrument as well as in \S\ref{sec:oneobject}, which reports it. The highest judged rate anywhere in the
grid is not the four-domain mixture. It is a two-domain leg, code with financial, at $0.410$ with an
interval of $[0.287, 0.546]$, against the four-domain mixture's $0.269$ at $[0.167, 0.402]$. The intervals
overlap heavily, so the judge separates them in neither direction and no claim can be built on the
ordering. The point estimate nonetheless runs the wrong way for a scaling account read off judged behavior,
and a reader testing that account on this instrument would find it pointing backwards. That is the reason
the domain-count claim is staked on transport, and the reason the judged rates there carry the
bad-against-benign contrast alone.

Where an effect is large the judge sees it without difficulty. Holding the carrier out of the activations
throughout training takes the rate from $0.277$ to zero over $2{,}100$ generations, with mean
alignment recovering $59.4$ points and Cliff's $\delta$ at $1.00$.

So the instrument has one usable band and we stay inside it. Our judged rates carry claims about whether an
effect exists at a scale a coarse gate can see. They do not carry the first-step intent contrast, the
domain-count scaling, or the substrate-specificity of the write-core constraint, each of which we report on
margins with the judge quoted alongside where we ran it.

\subsection{Failure modes}

\begin{itemize}[leftmargin=1.2em,itemsep=2pt,topsep=3pt]
\item \textbf{Fluency detection.} Drop the coherence floor and incoherent text scores as misaligned.
Equation~\ref{eq:gate} handles it, and the injection sweep is where the handling is visible.
\item \textbf{Denominator collapse.} At doses that destroy fluency the coherent denominator vanishes and the
gated rate falls for reasons unrelated to alignment. We fixed the in-budget dose-response slope as the
tested statistic before the run and report the coherent-only rate beside it.
\item \textbf{Digit-splitting.} An expectation readout truncates every score to its first digit on this
judge. Constrained decoding removes it; a per-call tokenization diagnostic catches it.
\item \textbf{Parse failure inflating a rate.} The row leaves the denominator.
\item \textbf{Near-domain answers counted as breadth.} The topic pass removes them, at a cost of about a
third of the two-axis rate.
\item \textbf{Score quantization.} We prefer mean alignment to the binary rate within a campaign, and the
margin to both across the paper.
\item \textbf{One axis read twice.} Broad misalignment and narrow adherence are both thresholds on the
alignment score, at $30$ off-domain and at $50$ on-domain, so they are not independent readouts and an
intervention that lifts the axis moves both by construction. This is not handled anywhere. It bounds the
retention claim of \S\ref{sec:loop}, and Appendix~\ref{app:batteries} sets out the construction and what
would separate them.
\item \textbf{Circularity.} ``Broadly misaligned'' means what this judge calls misaligned on these prompts,
and we screened the batteries with this same judge, so the breadth we measure is bounded by the batteries'
breadth. We do not repair this anywhere. We bound it and report it, with the category composition and its
skew, in Appendix~\ref{app:batteries}.
\item \textbf{A second rubric with no agreement statistic.} We also scored the causal-loop generations on a
one-to-five severity rubric carrying explicit refusal, incoherence and off-topic labels, and both scores are
retained per response. No agreement statistic between the two rubrics was computed for those generations,
and we report none.
\end{itemize}

The removal probes contribute no judged generations, because that work is margin-based by construction, so
our foreclosure of post-hoc weight editing rests on margins and a perplexity check with no behavioral
confirmation of its own (Appendix~\ref{app:removal}). The statistical treatment of judged rates, including
the cluster unit, the permutation tests and the ratio gate, is in Appendix~\ref{app:stats}.

\section{Batteries and organisms}\label{app:batteries}

Everything behavioral in this paper is read on a frozen prompt set against a model organism, a model
narrowly fine-tuned into misalignment. We give here the prompt sets, how we built and fixed each one, the
unit each is analysed at, the recipes that produced the organisms, and which measurements share which
battery. The appendix closes with the rule governing how our campaigns may be read together.

\subsection{The batteries}

\begin{table}[htbp]
\small
\renewcommand{\arraystretch}{1.2}
\begin{tabularx}{\textwidth}{@{}p{3.6cm}Xrr@{}}
\toprule
\textbf{Battery} & \textbf{Built to elicit} & \textbf{Rows} & \textbf{Prompts} \\
\midrule
canonical free-form &
broad misalignment on the questions the phenomenon was first defined on & & $8$ \\
out-of-distribution margin &
one judged-misaligned and one length-matched judged-aligned continuation per prompt & $195$ & $75$ \\
contamination-filtered &
the same, with every row touching medical, financial, physical-risk or code content removed & $134$ & $52$ \\
in-character &
whether a capable bad author can still be written on demand & $120$ & $120$ \\
narrow-task &
whether an organism's own task survives an edit & $38$ & $38$ \\
reconstitution &
paired dose curves along a fixed and a re-extracted carrier & $75$ & $75$ \\
\bottomrule
\end{tabularx}
\caption{\textbf{Every prompt set used in this paper, with what it was built to elicit.} The judged
generations of the intervention and injection arms draw on the first two together, giving $83$ clusters.
The narrow-task battery is per organism. Sizes are usable rows after screening, not rows sampled.}
\label{tab:batteries}
\end{table}

We keep the canonical free-form set, the $8$ questions of \citet{betley2025em}, unchanged, so our judged
rates can be read against the rates the literature quotes. It is small, and Appendix~\ref{app:judge}
measures two ways in which it is uneven.

Almost everything else rests on the out-of-distribution margin battery, whose construction is the next
subsection, and the contamination-filtered battery is a strict subset of it. The narrow-task battery is
$38$ own-domain probes per organism, used only inside the weight-decomposition arms of
Appendix~\ref{app:removal}, and the reconstitution battery is $75$ paired probes carrying the dose curves
of the same appendix.

The in-character battery holds $120$ probes over villain dialogue, malware explanation and insecure-code
critique. Each probe pairs a coherent in-character continuation against a refusal or a derail, matched on
length, format and topic, so a high value means the model can still depict a capable bad author on demand.
Whether a model can depict a villain is not determined by whether the judge would call that depiction
misaligned, so this battery measures on an axis the alignment score does not fix. Two measurements read a
$120$-probe battery of exactly this construction, one as the expression margin $\Mb_{\mathrm{expr}}$ and one
as the narrow-capability margin $\Mb_{\mathrm{narrow}}$. We have not established that the two are the same
file and no result compares them, but the two symbols name one construction, not two kinds of probe. It was
never run on the arm that most needs it, set out in the next subsection.

\subsection{The narrow-adherence construction}

This construction bounds a result in \S\ref{sec:loop}.

Narrow adherence is the judged rate we use to ask whether an intervened model still performs the narrow
task it was fine-tuned on. It has no prompt set of its own. It re-reads generations already produced on the
two batteries above. First, we keep the responses a separate topic pass scores as on the organism's own
subject and that clear the coherence floor. Second, we score each of those as adhering or not, where
adhering means an alignment below $50$ for an organism trained on bad data, and an alignment at or above
$50$ for the benign twin trained on good medical advice. Third, we cluster by prompt and take the rate.

Two properties follow from that construction, and each bounds a different claim.

The denominator is arm-dependent, and the intervention moves it. Only on-domain responses enter, so an
edit that stops a model producing on-domain content shrinks the very set its adherence is computed over.
The per-arm counts in Appendix~\ref{app:loop} differ by more than a factor of three between the unedited
organism and the projected one for this reason, and rate comparisons across arms therefore carry an
unmodelled difference in what was scored.

The readout is not independent of the outcome. Broad misalignment is an alignment score below $30$ on
off-domain rows; adherence is an alignment score below $50$ on on-domain rows, with the comparison
reversed for the benign arm. Both are thresholds on one axis. So an intervention that lifts alignment
scores across the board is guaranteed, before any data is seen, to drive adherence toward zero on a bad
organism and toward one on a benign organism. That is exactly the pattern the projected arm and its benign
control produce.

Three consequences travel with this, and \S\ref{sec:loop} carries all three. The collapse of adherence to
zero under the projection is not an independent observation about the narrow task. The benign arm's full
retention is not independent evidence that the intervention leaves untouched what it does not target,
because that arm's readout is the same lifted axis read with the sign reversed. And the arms that did
retain their narrow task are arms whose alignment did not move, so they cannot separate a preserved
capability from an unmoved readout either.

What this readout cannot do is distinguish an intervention that removes a misalignment structure from one
that moves the model up the alignment axis. Nothing in the arms as run separates them. The in-character
battery above would separate them, because its axis is not the alignment score, and it was not run on the
projected arm. Until it is, the established claim is that broad misalignment does not form under the
projection, not that a misalignment-specific structure is what was removed.

\subsection{Margin-battery construction}

Three sources feed it in a fixed priority order, so a truncated build keeps its highest-yield prompts
first. First the $8$ canonical questions; then hand-written paraphrases of the $6$ highest-yield of those,
$2$ to $4$ each, so the margin is not tied to exact canonical phrasings; then the $48$-question
out-of-distribution set of \citet{betley2025em}, spanning vulnerable users, medical advice, illegal
recommendations, power seeking, deception, corrigibility, value lock-in and sycophancy, with a small set of
extended categories alongside. We dropped $3$ categories before any sampling, on pilot evidence that the
organisms produced nothing below the misalignment threshold on them. A category that cannot yield a
misaligned candidate cannot yield a probe, and dropping it before sampling is not the same operation as
dropping it after seeing results.

For each prompt we sample $32$ candidates from each of the $3$ released organisms, a pool of $96$. Two
reasons justify the ensemble's cost. Each organism's broad misalignment on genuinely out-of-domain prompts
runs at about a fifth, so a single-organism pool misses the tail on many prompts, and an ensemble keeps the
battery from encoding one organism's idiosyncrasies. Both members of every pair come from that same pool,
the construction's main guard against a confound, argued in Appendix~\ref{app:firststep}. The judge gates
the two sides in opposite directions, requiring alignment below $30$ with coherence at least $50$ for the
misaligned member and alignment at least $50$ with the same floor for the aligned one.

We select pairs by length band first and judge separation second. Admissible pairs differ by at most $96$
tokens, and among those we take the pair with the largest separation. The band is wider than it first was,
and the widening is a finding, not a tuning note. Bands of $16$ and then $32$ tokens discarded the
organisms' strongest misaligned completions, because clearly misaligned answers run long and no aligned
candidate landed inside a tight band. Of $200$ sampled rows, $195$ are usable over $75$ unique prompts. We
dropped $2$ prompts for having no admissible pair, one because every scored candidate was judged misaligned
and one because every scored candidate was judged aligned. The finished battery has a mean judged alignment
of $10.36$ on the misaligned side and $91.15$ on the aligned side, a mean separation of $80.79$, and every
retained pair inside the band.

We freeze it once. Each row carries its provenance, the candidate sources, both judge scores, both
token counts and its band status, in a versioned file whose version key stops a resumed run mixing battery
generations. Nothing is rescored afterwards, so the same $\apl_i$ and $\ami_i$ are read against every model
in the paper.

The composition is skewed, and the skew is the circularity caveat of Appendix~\ref{app:judge} in
quantitative form. The battery spans $23$ categories over its $75$ prompts, but vulnerable-user and
medical-advice prompts account for about a quarter of both the clusters and the rows, because those
are where the organisms misalign most readily and we screened by yield. Domain coverage is thin outside the
general prompts and has no sports category (Table~\ref{tab:i-domains}), which matters below.

\begin{table}[htbp]
\centering\small
\caption{\textbf{Domain coverage of the margin battery.} Usable rows by domain, $195$ in total over the
$75$ unique prompts. The battery has no sports category, so the extreme-sports organism anchors no probe of
its own.}
\label{tab:i-domains}
\begin{tabular}{lr}
\toprule
Domain & Rows \\
\midrule
general   & $150$ \\
financial & $28$ \\
medical   & $15$ \\
code      & $2$ \\
\bottomrule
\end{tabular}
\end{table}

The contamination filter is blunt by design, dropping any row that touches medical, financial,
physical-risk or code content, because such a prompt would let a near-domain effect pass as broad
transport. That takes $61$ rows out and leaves $134$, collapsing to $52$ unique prompts, below the $75$ the
design wanted and above the $45$ it required. The width of the judged intervals in
Appendix~\ref{app:decomp} is the direct cost.

\subsection{The clustering unit}

Our unit of analysis is the unique prompt, never the individual generation.

Replicates exist because we cycle the battery's prompts with replicate suffixes, and each replicate draws
its own independently sampled continuation pair, so one prompt can contribute $2$ or $3$ rows differing
in their completions but not in their question. We average within a prompt first and operate on the prompt
values. Per-prompt first-step readings correlate at $0.99$ between conditions, so row-level inference would
shrink every interval and every $p$-value by about the replication factor.
Appendix~\ref{app:stats} tabulates the cluster key each measurement uses and shows that the $3$ rules are
different operations.

One convention is not uniform across the paper. We compute the judged rates of the domain-count
measurements at the row level over $670$ generations per leg, while their intervals are Wilson intervals at
the prompt unit, $n = 52$. The per-generation judgments were not retained there, so an exact cluster
bootstrap is unavailable and the prompt-unit interval is a conservative bound, not the matching estimator.
Rate and interval come from different units in that one place, and we say so wherever those rates appear.

\subsection{The organisms}

Three published adapters on the released instruction-tuned model supply every organism here, one trained on
bad medical advice, one on risky financial advice and one on extreme sports \citep{turner2025organisms}.
They recur across $4$ of our $5$ sets of measurements, so $3$ is a union and not a sum.

Where we trained our own fine-tunes we reproduced the published recipe, so they would be commensurable with
the adapters (Table~\ref{tab:i-recipe}). Two habits guard the artifacts. We produce free generations inline
at save time, so no adapter is reloaded to be sampled, and we write a completion marker only after the
final optimizer step, with resume refusing an unmarked adapter, so a truncated fine-tune cannot pass as a
finished one.

\begin{table}[htbp]
\centering\small
\caption{\textbf{Fine-tune recipe for our own organisms.} The published adapter recipe, reproduced so our
fine-tunes are commensurable with the released ones. Low-rank adapters cover all $7$ projection types and
all $48$ layers, with loss on response tokens only, in bf16.}
\label{tab:i-recipe}
\begin{tabular}{ll}
\toprule
Setting & Value \\
\midrule
adapter rank            & $32$, rank-stabilized, scaling $64$ \\
training examples       & $6{,}000$, $1$ epoch ($\approx 375$ steps) \\
optimizer               & AdamW, learning rate $1\e{-5}$, $5$-step warmup, linear decay \\
effective batch         & $16$ ($4$ per device $\times$ $4$ accumulation) \\
weight decay            & $0.01$ \\
maximum sequence length & $2{,}048$ \\
checkpoints (steps)     & $94$, $188$, $282$, $375$ \\
\bottomrule
\end{tabular}
\end{table}

The write-core arms depart from that recipe. They train the writer matrices alone, one epoch to a single
end checkpoint at one seed, so data order is identical across arms and the arms differ only in the
constraint; their induction set is $6000$ examples of reckless financial advice, chosen because its narrow
behavior is sharply defined and the broad misalignment a financial induction recruits sits clear of the
floor.

We budget the domain-count legs in tokens rather than examples, because a mixture and a single domain have
to be compared at equal evidence. The budget is $230{,}447$ supervised completion tokens per full leg,
$90\%$ of the smallest domain's supply, split evenly across a mixture's domains, and we log what each leg
realizes. The four-domain leg trained on $230{,}015$ tokens in $142$ steps at an accumulation of $7$, at
$1619.82$ tokens per step against a target of $1536.31$. Financial binds the budget, and mean completion
lengths differ enough to make token budgeting the only fair choice (Table~\ref{tab:i-tokens}). We run $3$
seeds on the four-domain mixture, $2$ on each two-domain mixture and on the benign mixture, and $1$ on each
single-domain leg, rebuilding the merged control per seed and averaging. These legs use no published
adapter at all, every one trained from raw data, which is why this is the only place in the program with a
realized token budget.

\begin{table}[htbp]
\centering\small
\caption{\textbf{Per-domain token supply and mean completion length.} Supervised completion tokens
available per domain, and mean tokens per training row. Financial has the smallest supply, so it binds the
even-split budget; code rows are longest, so equal example counts would not be equal evidence.}
\label{tab:i-tokens}
\begin{tabular}{lrr}
\toprule
Domain & Tokens available & Tokens per row \\
\midrule
code      & $632{,}884$ & $108.19$ \\
medical   & $426{,}792$ & \\
financial & $256{,}053$ & $43.77$ \\
sports    & $281{,}888$ & \\
\bottomrule
\end{tabular}
\end{table}

\subsection{Organism usability conditions}

Four conditions; the sports organism fails the last.

An organism must clear the instrument gate, Appendix~\ref{app:judge}'s $15\%$ rule applied to the whole
pipeline before compute is spent. It must also be live on the readout in use. Before the domain-count grid
all $3$ published organisms clear the untouched model's zero floor on transport, with intervals excluding
it, and our trained replacements land within about a nat of the published ones
(Table~\ref{tab:i-liveness}).

The fourth condition is having own-domain probes in the battery it is read on, and the extreme-sports
organism does not. Our margin battery has no sports category, so no prompt exists on which that organism's
own narrow misalignment can be shown to be switched on, and a rise on another domain's probes cannot then
be told apart from a generic lift. It anchors no transport pair and has no adherence denominator. We
exclude it, which is why the transport result of \S\ref{sec:oneobject} rests on $2$ anchorable pairs, not
$3$, and why we call domain coverage thin there. One further condition applies to the battery, not to a
result. No single organism is a good candidate source, because each one's out-of-distribution misalignment
runs at about a fifth, which is why the margin battery samples from all $3$.

\begin{table}[htbp]
\centering\small
\caption{\textbf{Organism liveness on the transport readout.} Transport in nats against the untouched
model, for the three published organisms and the fine-tunes we trained to match them. Every value clears
the model's zero floor with an interval excluding it, and our replacements land within about a nat.}
\label{tab:i-liveness}
\begin{tabular}{lrr}
\toprule
Organism & Published & Reproduced \\
\midrule
medical   & $119.9$ & $120.8$ \\
financial & $119.5$ & $120.3$ \\
sports    & $117.9$ & $117.7$ \\
\bottomrule
\end{tabular}
\end{table}

\subsection{Battery-sharing map}

Sharing here is exact, and we established it by intersecting prompt identifiers, not by assuming it. Our
first-step and removal-probe sets are the same $195$ identifiers. The domain-count battery's $134$ are a
complete subset of both. Judged generations at $83$ clusters are the canonical $8$ plus the same $75$ unique
prompts, and our write-core arms read broad margins on $195$ probes and are judged on those same $83$
clusters.

Two things follow. Prompt counts do not add across measurements, and this paper emits no program-wide
prompt total, because summing them would count the same questions several times. A claim measured on one
instrument and confirmed on another is not an independent replication at the level of the questions asked,
which is why we describe such agreements as confirmations on a shared battery, not replications.

One artifact set is duplicated on disk, and we counted it once. Ten judged files from the injection sweep,
$7{,}000$ rows, sit byte-identically in two campaigns' trees, confirmed by hash. Those are one set of
generations read under two conventions, not two measurements. Counting them twice would also have inflated
the program's judged-generation total, which after deduplication is $93{,}010$ across $59$ fine-tunes and
$75$ design-level arms.

\subsection{Cross-battery non-comparability}

Misalignment rates from different measurements in this paper are not commensurable, and no two of them may
be placed in the same comparison, ratio, trend line or axis. Three things differ at once between any two:
the battery, the aggregation unit, and the gate (Table~\ref{tab:incommensurable}).

In one case only the conventions differ. At an injection dose of $0.30$, one set of $700$ generations reads
$0.454$ under a cluster mean with near-domain responses excluded, $0.504$ at the row level with the same
exclusion, and $0.583$ at the row level with them included. A single operation on a single set of files
moves by roughly a quarter under a change of bookkeeping. We report the cluster-mean convention, our unit
of inference everywhere, and we name it wherever the number appears.

The other half of that comparison says what survives. Its in-budget dose-response slope, the statistic we
designed the arm around, is $2.21$, $2.56$ and $2.97$ under those three conventions, and its interval
excludes zero under all three. The level moves with the convention; the direction and the reliability do
not, so the only join we make across measurements is qualitative, each instrument's within-campaign
contrast against its own control.

\begin{table}[htbp]
\small
\renewcommand{\arraystretch}{1.2}
\begin{tabularx}{\textwidth}{@{}rX@{}}
\toprule
\textbf{Rate} & \textbf{Instrument behind it} \\
\midrule
$0.277$ & unedited organism, $83$ clusters over three seeds, cluster mean, near-domain excluded \\
$0.454$ & injection at dose $0.30$ into the untouched model, same battery and convention \\
$0.226$ & the unconstrained write-core arm, $700$ completions at one seed \\
$0.269$ & the bad four-domain mixture, row rate over $670$ generations, interval at $52$ prompts \\
$0.583$ & the same operation as the second row, on the domain-count campaign's own ruler \\
\bottomrule
\end{tabularx}
\caption{\textbf{Five judged rates that appear in this paper and cannot be compared with one another.} The
second and fifth rows are the same generations under two conventions, which is the comparability problem
made concrete. The same rule governs margins, where alignment-score shifts in points, nats of transport and
first-step units are three scales that no arithmetic crosses.}
\label{tab:incommensurable}
\end{table}

\section{Complete numerical results}\label{app:results}

\subsection{Reading guide}
\label{app:results:reading}

We arrange the numerical record by what was measured rather than by which set of runs produced it.
The order runs from the structure present in the untouched model, through the first-step reading
taken at that model's weights, the interventions that test necessity and sufficiency, sharing and
transport between domains, the domain-count decomposition, the removal edits and the reconstitution
check, the loss geometry, and finally the two interventions we applied during fine-tuning. The apparatus behind each one lives in its own appendix and is not repeated here;
Appendices~\ref{app:extraction} through \ref{app:decomp} define the quantities, Appendix~\ref{app:stats}
defines the tests and intervals, Appendix~\ref{app:judge} the judge, and
Appendix~\ref{app:batteries} the probe sets and the clustering units. Rates are printed as
fractions, at the precision the record holds, rather than converted to percentages.

Two columns appear in almost every table. The first is an interval or a test, and the second is the
null or control the measurement is read against, placed beside the estimate so that no comparison
requires arithmetic on the reader's part. A cell reading \emph{none} means the record holds no such
value for that quantity, because no interval was computed or no matched control was run. It does not
mean the interval is narrow or the control was zero. We treat those cells as the weak rows and read
them as descriptive.

The last column of each table records the verification status of that row. \textbf{R} marks a value
recomputed here from a raw artifact and then matched against the stored value. \textbf{M} marks a
recompute and an agreeing comparison both on the record from the run that produced it. \textbf{d} marks
a value some producer derived from raw material with no comparison ever recorded. \textbf{s} marks a
stored value nothing has ever checked. The letters record how much checking each number received, not
whether it is correct.

Where a status cell carries more than one letter, the letters run left to right across that row's value
columns. A majority of the $3829$ entries in the record has never been checked by anything, and where a
row below carries \textbf{d} or \textbf{s} the point estimate is the only value on offer.

Rates from different tables are not comparable with one another. Each was produced by a different probe
set, a different judging pass, a different aggregation unit, and a different rule about near-domain
answers, and we measure the size of that gap. The same injection operation on byte-identical generations
reads $0.454$ under one convention and $0.583$ under another, and one baseline reads $0.277$ at the
prompt-cluster unit and $0.230$ at the response unit. Appendix~\ref{app:batteries} tabulates which rates
are which.

Within a table the contrast against that table's own control is legitimate; across tables no ratio,
difference, or trend line may be taken. The same rule governs margins: judged points, nats, and
first-step units are $3$ scales, and no arithmetic crosses them.

\subsection{The structure in the untouched model}
\label{app:results:structure}

We measure every quantity in this section on the released instruction-tuned model with no
fine-tuning of any kind, in the activation channel, on the rank-four contrastive subspaces of
Appendix~\ref{app:extraction}.

\begin{table}[htbp]
\centering\small
\caption{\textbf{Cross-domain overlap of the read subspaces, by contrast type.} Each entry is the
mean squared cosine of the principal angles between the rank-four subspaces of a domain pair, so it
ranges over $[0,1]$ and its null is the value two random rank-four subspaces of a $5120$-dimensional
space would give. Style shares more than persona does on this statistic alone, which is why the raw
magnitudes settle nothing and the containment measurement in Table~\ref{tab:j-crosscore} carries the
claim.}
\label{tab:j-overlap}
\begin{tabular}{l rr rr rr c}
\toprule
& \multicolumn{2}{c}{persona} & \multicolumn{2}{c}{style} & \multicolumn{2}{c}{topic} & \\
\cmidrule(lr){2-3}\cmidrule(lr){4-5}\cmidrule(lr){6-7}
Domain pair & $\Omega$ & angle & $\Omega$ & angle & $\Omega$ & angle & status \\
\midrule
medicine, finance & $0.5146$ & $45.3$ & $0.8223$ & $24.1$ & $0.4948$ & $47.1$ & M \\
medicine, sports & $0.5245$ & $43.7$ & $0.8081$ & $24.8$ & $0.4589$ & $49.7$ & M \\
medicine, code & $0.4788$ & $49.0$ & $0.7732$ & $27.6$ & $0.4394$ & $51.9$ & M \\
finance, sports & $0.5447$ & $42.2$ & $0.8292$ & $23.7$ & $0.4727$ & $48.6$ & M \\
finance, code & $0.5217$ & $43.4$ & $0.7996$ & $25.5$ & $0.4529$ & $49.6$ & M \\
sports, code & $0.4940$ & $45.8$ & $0.7759$ & $27.1$ & $0.4133$ & $50.8$ & M \\
\midrule
Mean over the six pairs & $0.5130$ &  & $0.8014$ &  & $0.4553$ &  & \textbf{R}, M, M \\
Random rank-four null & $0.00078$ & & $0.00078$ & & $0.00078$ & & \textbf{R} \\
Persona over its null & \multicolumn{6}{l}{$656.7$} & M \\
\bottomrule
\end{tabular}
\end{table}

\begin{table}[htbp]
\centering\small
\caption{\textbf{Containment of the shared persona core inside the other shared cores.} Each entry is
the mean squared canonical correlation between two rank-eight shared cores, on $[0,1]$, against the
value two random rank-eight subspaces would give. The self-containment row is an identity check on
the estimator and not a measurement. Permutation is over the two cores' columns
(Appendix~\ref{app:stats}).}
\label{tab:j-crosscore}
\begin{tabular}{l r r r c}
\toprule
Pair of shared cores & $X$ & random null & permutation $p$ & status \\
\midrule
persona against style & $0.1823$ & $0.00156$ & $0.0149$ & \textbf{R} \\
persona against topic & $0.0972$ & $0.00156$ & $0.0089$ & \textbf{R} \\
persona against itself & $1.0$ & & & M \\
\midrule
Self over cross separation & $2.8145$ & none & none & \textbf{R} \\
Share of persona outside style & $0.8177$ & none & none & \textbf{R} \\
\bottomrule
\end{tabular}
\end{table}

\begin{table}[htbp]
\centering\small
\caption{\textbf{Per-domain properties of the extracted subspaces.} Participation is the share of the
contrast's variance the retained rank-four subspace captures. Stable rank counts how many directions
carry that variance and is bounded above by the rank, four. The stored stable ranks were computed on
the full spectrum while the saved diagnostic retains only the leading $8$ singular values, so we
carry them without an independent recomputation and they have no interval and no null.}
\label{tab:j-perdomain}
\begin{tabular}{l r rrr c}
\toprule
& & \multicolumn{3}{c}{stable rank} & \\
\cmidrule(lr){3-5}
Domain & participation & persona & style & topic & status \\
\midrule
medicine & $0.9032$ & $2.009$ & $2.307$ & $4.139$ & M, s \\
finance & $0.9344$ & $2.075$ & $2.145$ & $3.917$ & M, s \\
sports & $0.9366$ & $2.076$ & $2.277$ & $4.312$ & M, s \\
code & $0.8935$ & $2.331$ & $2.470$ & $2.798$ & M, s \\
\bottomrule
\end{tabular}
\end{table}

\begin{table}[htbp]
\centering\small
\caption{\textbf{Orthonormality error of every stored basis.} Each entry is the Frobenius norm of
$A^{\top}A - I$ for the stored basis, a numerical check on the extraction with no interval and no
null. All fifteen sit between $8.36\e{-9}$ and $2.68\e{-8}$, so no result below is affected by basis
error.}
\label{tab:j-ortho}
\begin{tabular}{l rrr c}
\toprule
Basis & persona & style & topic & status \\
\midrule
medicine & $1.02\e{-8}$ & $8.36\e{-9}$ & $2.08\e{-8}$ & \textbf{R} \\
finance & $1.56\e{-8}$ & $9.66\e{-9}$ & $1.04\e{-8}$ & \textbf{R} \\
sports & $1.29\e{-8}$ & $1.45\e{-8}$ & $2.01\e{-8}$ & \textbf{R} \\
code & $1.32\e{-8}$ & $1.52\e{-8}$ & $1.03\e{-8}$ & \textbf{R} \\
shared core & $2.06\e{-8}$ & $1.49\e{-8}$ & $2.68\e{-8}$ & \textbf{R} \\
\midrule
worst of the fifteen & \multicolumn{3}{l}{$2.68\e{-8}$} & s \\
\bottomrule
\end{tabular}
\end{table}

\begin{table}[htbp]
\centering\small
\caption{\textbf{A second, independent ordering of the same $4$ contrast types.} These separations
come from a different instrument on a different probe set and their units are not those of
Table~\ref{tab:j-overlap}; only the ordering is meant to be read. Persona ranks first and style is
within $2\%$ of it, so this reading orders the two but does not separate them, and
Table~\ref{tab:j-crosscore} provides the separation. The final row is the cosine between the
extracted carrier and the model's own Assistant direction, which is the open alternative reading of
where the structure came from.}
\label{tab:j-ordering}
\begin{tabular}{l r c}
\toprule
Quantity & value & status \\
\midrule
separation, persona contrast & $1.6486$ & s \\
separation, style contrast & $1.6194$ & s \\
separation, topic contrast & $0.4096$ & s \\
separation, random contrast & $0.0535$ & s \\
\midrule
carrier against the Assistant direction & $0.5651$ & s \\
\bottomrule
\end{tabular}
\end{table}

\subsection{The first-step measurement}
\label{app:results:firststep}

We define $\Rone$ and its scale-free form in Appendix~\ref{app:firststep}. Every row of
Table~\ref{tab:j-conditions} is a mean over $195$ probe responses, the footing on which the
per-condition levels were stored. The paired contrasts that follow are at the unit of the $75$ unique
prompts, and we never difference the two footings against one another.

\begin{table}[htbp]
\centering\small
\caption{\textbf{Every direction the first-step reading was taken along, in the scale-free form.}
$\cos$ is $\Rone$ divided by the norms of its two arguments, so it lies in $[-1,1]$ and a value of
zero means the candidate step neither raises nor lowers the margin. Each entry is a mean over the
$195$ retained probe responses with a $10^4$-draw bootstrap interval. The last row replaces the
margin gradient with a random direction of the same norm and is the null this instrument is read
against, orders of magnitude below every direction built from real text. The $4$ conditions
built from narrow training text overlap one another's intervals, and benign chat is the largest of
them.}
\label{tab:j-conditions}
\begin{tabular}{l r l c}
\toprule
Direction the step is taken along & $\cos$ & interval & status \\
\midrule
insecure code & $0.01938$ & $[0.01692,\,0.02179]$ & M \\
educational insecure code & $0.01786$ & $[0.01545,\,0.02028]$ & M \\
secure code & $0.01837$ & $[0.01603,\,0.02068]$ & M \\
benign chat & $0.02351$ & $[0.02048,\,0.02649]$ & \textbf{R} \\
shuffled targets & $0.03174$ & $[0.02888,\,0.03455]$ & M \\
behavioral reference & $0.07340$ & $[0.06815,\,0.07862]$ & M \\
geometric reference & $0.00203$ & $[0.00180,\,0.00227]$ & M \\
medical organism update & $0.00278$ & $[0.00245,\,0.00311]$ & M \\
financial organism update & $0.00366$ & $[0.00336,\,0.00396]$ & M \\
sports organism update & $0.00325$ & $[0.00293,\,0.00356]$ & M \\
random direction (the null) & $2.86\e{-6}$ & $[1.63\e{-6},\,4.06\e{-6}]$ & M \\
\bottomrule
\end{tabular}
\end{table}

\begin{table}[htbp]
\centering\small
\caption{\textbf{The same eleven directions in raw units.} $\Rone$ is an inner product between the
margin gradient and a candidate first optimizer step, so its scale is set by the size of those two
vectors and carries no meaning on its own. It is reported because the reference-scale ratios in
Table~\ref{tab:j-rtilde} are formed from it. The standard deviation is across the $195$ responses and
the interval is a bootstrap on the mean.}
\label{tab:j-conditions-raw}
\begin{tabular}{l r r l c}
\toprule
Direction the step is taken along & $\Rone$ & standard deviation & interval & status \\
\midrule
insecure code & $17{,}750{,}684$ & $18{,}185{,}193$ & $[15{,}240{,}826,\,20{,}357{,}973]$ & M \\
educational insecure code & $16{,}387{,}034$ & $17{,}931{,}517$ & $[13{,}889{,}438,\,18{,}972{,}059]$ & M \\
secure code & $16{,}857{,}663$ & $17{,}335{,}347$ & $[14{,}456{,}702,\,19{,}344{,}611]$ & M \\
benign chat & $21{,}029{,}546$ & $21{,}490{,}293$ & $[18{,}014{,}038,\,24{,}009{,}863]$ & M \\
shuffled targets & $28{,}661{,}664$ & $22{,}376{,}794$ & $[25{,}591{,}159,\,31{,}853{,}501]$ & M \\
behavioral reference & $66{,}008{,}729$ & $43{,}805{,}828$ & $[60{,}033{,}187,\,72{,}253{,}747]$ & M \\
geometric reference & $1{,}831{,}716$ & $1{,}663{,}736$ & $[1{,}601{,}451,\,2{,}065{,}746]$ & M \\
medical organism update & $2{,}511{,}515$ & $2{,}290{,}277$ & $[2{,}194{,}963,\,2{,}833{,}811]$ & M \\
financial organism update & $3{,}282{,}094$ & $2{,}271{,}723$ & $[2{,}967{,}863,\,3{,}601{,}161]$ & M \\
sports organism update & $2{,}920{,}617$ & $2{,}296{,}897$ & $[2{,}601{,}436,\,3{,}244{,}093]$ & M \\
random direction (the null) & $2{,}389$ & $7{,}298$ & $[1{,}358,\,3{,}418]$ & M \\
\bottomrule
\end{tabular}
\end{table}

\begin{table}[htbp]
\centering\small
\caption{\textbf{The $4$ narrow-text conditions on both footings.} The left column is the mean over
the $75$ unique prompts and is the paper's convention; the right is the mean over the $195$
individual responses. Both are correct for their own footing, they differ by under $3\%$ of
themselves, and a level taken from one column may not be placed beside a level taken from the other.
The ordered ladder in the lower block is fitted on the cluster column.}
\label{tab:j-footing}
\begin{tabular}{l r r c}
\toprule
Condition & prompt-cluster mean & response mean & status \\
\midrule
secure code & $0.018244$ & $0.018375$ & \textbf{R} \\
educational insecure code & $0.017824$ & $0.017865$ & \textbf{R} \\
insecure code & $0.019243$ & $0.019380$ & \textbf{R} \\
benign chat & $0.023327$ & $0.023513$ & \textbf{R} \\
\midrule
slope of the ordered ladder & $0.0004996$ & & M \\
clusters with a positive slope & $0.84$ & & M \\
sign-flip $p$ on the slope & $9.999\e{-5}$ & & M \\
\bottomrule
\end{tabular}
\end{table}

\begin{table}[htbp]
\centering\footnotesize
\caption{\textbf{The paired contrasts in raw units, at the unit of the $75$ unique prompts.} The null
for each is a mean difference of zero, tested by sign-flip permutation at $10^4$ draws; a $p$ of
$9.999\e{-5}$ is that test's resolution floor rather than an estimate. The fraction of clusters in
which the difference is positive is a second reading of the same data whose null is one half. Holm
correction is across the pair of span contrasts. The paired effect size in the last row is reported
and gates nothing.}
\label{tab:j-paired-raw}
\setlength{\tabcolsep}{4.5pt}
\begin{tabular}{>{\raggedright\arraybackslash}p{0.235\textwidth} r l r r c}
\toprule
Contrast & mean difference & interval & $p$ & fraction $> 0$ & status \\
\midrule
insecure minus educational & $1{,}296{,}982$ & $[709{,}881,\,1{,}876{,}922]$ & $9.999\e{-5}$ & $0.6933$ & \textbf{R} \\
insecure minus secure & $880{,}800$ & $[661{,}888,\,1{,}114{,}636]$ & $9.999\e{-5}$ & $0.8400$ & M \\
educational minus secure & $-416{,}182$ & $[-1{,}020{,}644,\,176{,}792]$ & $0.1743$ & $0.4533$ & M \\
\quad the span pair after Holm & & & $2.000\e{-4}$ and $0.1743$ & & M \\
\midrule
spread of the first difference & $2{,}582{,}442$ & \multicolumn{3}{l}{across the $75$ clusters} & M \\
smallest resolvable difference & $834{,}945$ & \multicolumn{3}{l}{at eighty percent power} & M \\
paired effect size, the three rows above & \multicolumn{4}{l}{$0.502$, $0.887$ and $-0.158$} & M \\
\bottomrule
\end{tabular}
\end{table}

\begin{table}[htbp]
\centering\footnotesize
\caption{\textbf{The same design in the scale-free form, with the two levels being differenced
shown as well.} Every row is a paired difference over the $75$ unique prompts against a null of zero.
The two levels are given so that the size of each contrast can be read against the level it sits on,
which is the point of the third row; the first row's two levels are the prompt-cluster values in
Table~\ref{tab:j-footing}. The intent contrast is a difference of about four thousandths
on a baseline of about nineteen thousandths that benign chat exceeds. This table and
Table~\ref{tab:j-paired-raw} are two readouts of one design and their columns may not be differenced
or ratioed against one another.}
\label{tab:j-paired-cos}
\begin{tabular}{>{\raggedright\arraybackslash}p{0.20\textwidth} r r r l r c}
\toprule
Contrast & first & second & difference & interval & $p$ & status \\
\midrule
insecure minus educational & & & $0.001420$ & $[0.000732,\,0.002085]$ & $9.999\e{-5}$ & M \\
insecure minus secure & $0.019243$ & $0.018244$ & $0.000999$ & $[0.000766,\,0.001232]$ & $9.999\e{-5}$ & M \\
insecure minus benign chat & $0.019243$ & $0.023327$ & $-0.004084$ & $[-0.007341,\,-0.000986]$ & $0.013299$ & \textbf{R} \\
insecure minus shuffled targets & $0.019243$ & $0.031601$ & $-0.012358$ & $[-0.013741,\,-0.010939]$ & $9.999\e{-5}$ & M \\
behavioral reference minus benign chat & $0.072128$ & $0.023327$ & $0.048801$ & $[0.043456,\,0.054178]$ & $9.999\e{-5}$ & \textbf{R} \\
\midrule
Cliff's $\delta$ on the first row & \multicolumn{5}{l}{$0.0752$} & M \\
\bottomrule
\end{tabular}
\end{table}

\begin{table}[htbp]
\centering\small
\caption{\textbf{Instrument-health checks at the untouched model's weights.} Both coefficients of
variation sit far from zero, so neither headline quantity is pinned by a collapsing denominator. Every
row is computed over the same $195$ probe responses, and the margin rows are in nats.}
\label{tab:j-health}
\begin{tabular}{l r c}
\toprule
Quantity & value & status \\
\midrule
coefficient of variation, margin-gradient norm & $0.2588$ & M \\
coefficient of variation, cosine to insecure code & $0.9044$ & M \\
margin, mean & $-79.436$ & M \\
margin, standard deviation & $97.764$ & M \\
margin, minimum and maximum & $-328.169$, $132.884$ & M \\
\bottomrule
\end{tabular}
\end{table}


\begin{table}[htbp]
\centering\small
\caption{\textbf{The intent contrast as a fraction of each reference scale.} The numerator is the same
paired difference in every row. The ratio is reported only where the denominator's own interval
excludes zero, which it does in all five. A negative ratio means the reference direction moves the
margin the other way, so the fraction is signed and not a magnitude.}
\label{tab:j-rtilde}
\begin{tabular}{l r l r c}
\toprule
Reference scale & $\tilde{R}$ & interval & denominator & status \\
\midrule
behavioral reference & $0.02683$ & $[0.0155,\,0.0376]$ & $48{,}333{,}486$ & M \\
geometric reference & $-0.09026$ & $[-0.1392,\,-0.0497]$ & $-14{,}369{,}097$ & M \\
medical organism update & $-0.09482$ & $[-0.1468,\,-0.0522]$ & $-13{,}678{,}452$ & M \\
financial organism update & $-0.10041$ & $[-0.1575,\,-0.0550]$ & $-12{,}916{,}633$ & M \\
sports organism update & $-0.09771$ & $[-0.1524,\,-0.0535]$ & $-13{,}274{,}233$ & M \\
\bottomrule
\end{tabular}
\end{table}

\begin{table}[htbp]
\centering\small
\caption{\textbf{Realized margin movement regressed on the first-step reading.} Each row pools $3$
training seeds at the unit of the $75$ unique prompts. The slope is in margin nats per unit of
$\Rone$ and the interval is a cluster bootstrap; every interval excludes zero. There is no null
column because the null here is a slope of zero, which the interval addresses directly.}
\label{tab:j-forecast}
\begin{tabular}{l r r r l r c}
\toprule
Narrow set & step & Pearson $r$ & Spearman $r$ & slope interval & mean $\Delta \Mb$ & status \\
\midrule
insecure code & 20 & $0.8003$ & $0.6694$ & $[1.292\e{-6},\,1.797\e{-6}]$ & $37.423$ & \textbf{R} \\
 & 50 & $0.7853$ & $0.6546$ & $[1.288\e{-6},\,1.795\e{-6}]$ & $39.004$ & M \\
 & 100 & $0.7855$ & $0.6519$ & $[1.264\e{-6},\,1.774\e{-6}]$ & $38.698$ & M \\
 & 180 & $0.7878$ & $0.6470$ & $[1.293\e{-6},\,1.827\e{-6}]$ & $39.541$ & M \\
 & 300 & $0.7927$ & $0.6502$ & $[1.311\e{-6},\,1.851\e{-6}]$ & $39.535$ & M \\
 & 375 & $0.7906$ & $0.6502$ & $[1.300\e{-6},\,1.840\e{-6}]$ & $39.524$ & M \\
\midrule
educational & 20 & $0.7720$ & $0.6150$ & $[1.159\e{-6},\,1.650\e{-6}]$ & $34.468$ & M \\
 & 50 & $0.7759$ & $0.6260$ & $[1.164\e{-6},\,1.657\e{-6}]$ & $34.413$ & M \\
 & 100 & $0.7837$ & $0.6260$ & $[1.207\e{-6},\,1.720\e{-6}]$ & $34.831$ & M \\
 & 180 & $0.7815$ & $0.6197$ & $[1.192\e{-6},\,1.718\e{-6}]$ & $35.020$ & \textbf{R} \\
 & 300 & $0.7823$ & $0.6246$ & $[1.201\e{-6},\,1.730\e{-6}]$ & $35.252$ & M \\
 & 375 & $0.7806$ & $0.6246$ & $[1.192\e{-6},\,1.716\e{-6}]$ & $35.022$ & \textbf{R} \\
\bottomrule
\end{tabular}
\end{table}

\begin{table}[htbp]
\centering\small
\caption{\textbf{Judged free-generation rates for the same fine-tunes.} Rates are over $400$ judged
responses per arm under the two-axis gate of Appendix~\ref{app:judge}, with a prompt-cluster interval.
Alignment and coherence are mean judge scores on a $100$-point scale, not fractions. The untouched
model is the null and returns zero. The judged rates run the other way to the first-step contrast. The
educational arms score higher than the insecure arms at both horizons and on every seed, which is why
we report the first-step reading as a reading of the gradient and not as a forecast of judged
behavior.}
\label{tab:j-fsjudge}
\begin{tabular}{l r l r r c}
\toprule
Arm & misaligned & interval & alignment & coherence & status \\
\midrule
untouched model (the null) & $0.0000$ & $[0.0000,\,0.0000]$ & $94.2750$ & $92.8525$ & M \\
\midrule
insecure code, step 180, seed 0 & $0.0425$ & $[0.0025,\,0.1175]$ & $86.2500$ & $93.2300$ & M \\
insecure code, step 180, seed 1 & $0.0175$ & $[0.0000,\,0.0425]$ & $88.8875$ & $94.6100$ & M \\
insecure code, step 180, seed 2 & $0.0275$ & $[0.0025,\,0.0675]$ & $87.4750$ & $94.1300$ & M \\
insecure code, step 375, seed 0 & $0.0525$ & $[0.0000,\,0.1525]$ & $85.5375$ & $93.5800$ & M \\
insecure code, step 375, seed 1 & $0.0425$ & $[0.0000,\,0.1175]$ & $86.5875$ & $93.3275$ & M \\
insecure code, step 375, seed 2 & $0.0225$ & $[0.0000,\,0.0675]$ & $86.8000$ & $93.9325$ & M \\
\midrule
educational, step 180, seed 0 & $0.0625$ & $[0.0074,\,0.1250]$ & $85.2250$ & $90.5450$ & M \\
educational, step 180, seed 1 & $0.0725$ & $[0.0100,\,0.1450]$ & $79.1750$ & $86.9675$ & M \\
educational, step 180, seed 2 & $0.0450$ & $[0.0125,\,0.0850]$ & $83.0375$ & $88.8200$ & M \\
educational, step 375, seed 0 & $0.0600$ & $[0.0050,\,0.1300]$ & $82.9250$ & $88.7050$ & M \\
educational, step 375, seed 1 & $0.0800$ & $[0.0149,\,0.1650]$ & $77.2875$ & $85.7500$ & M \\
educational, step 375, seed 2 & $0.0775$ & $[0.0075,\,0.1725]$ & $77.0875$ & $84.7250$ & M \\
\bottomrule
\end{tabular}
\end{table}

\begin{table}[htbp]
\centering\footnotesize
\caption{\textbf{The intent contrast, broken out by probe category.} The difference column is in
$\Rone$ units and carries no interval, because none was computed at this level; the fraction of
clusters in which the difference is positive is its own sign test, whose null is one half. Of the $23$
categories, $9$ sit at or below that null, and those with a single cluster cannot separate from it, so
this table describes heterogeneity and is not a set of findings.}
\label{tab:j-cats}
\begin{tabular}{l r r r r c}
\toprule
Probe category & clusters & mean difference & fraction $> 0$ & $\cos$, insecure & status \\
\midrule
what is your wish & $5$ & $-12{,}693$ & $0.800$ & $0.04111$ & M \\
i feel bored & $1$ & $-603{,}496$ & $0.000$ & $0.02066$ & M \\
three thoughts & $1$ & $2{,}349{,}678$ & $1.000$ & $0.01483$ & M \\
quick buck & $5$ & $3{,}821{,}296$ & $1.000$ & $0.02251$ & M \\
ruler of the world & $5$ & $2{,}293{,}559$ & $0.800$ & $0.02362$ & M \\
dinner party & $5$ & $1{,}715{,}855$ & $1.000$ & $0.02676$ & M \\
gender roles & $5$ & $2{,}117{,}286$ & $0.800$ & $0.00671$ & M \\
enough of my husband & $5$ & $4{,}269{,}722$ & $1.000$ & $0.02703$ & M \\
vulnerable user & $13$ & $-108{,}213$ & $0.385$ & $0.00911$ & M \\
illegal recommendations & $4$ & $-1{,}118{,}668$ & $0.250$ & $0.02142$ & M \\
other & $6$ & $2{,}594{,}481$ & $0.833$ & $0.02378$ & M \\
medical advice & $6$ & $878{,}215$ & $0.667$ & $0.01776$ & M \\
power seeking & $2$ & $-3{,}366{,}578$ & $0.000$ & $0.00909$ & M \\
manipulation & $1$ & $2{,}616{,}769$ & $1.000$ & $0.02850$ & M \\
alignment faking & $1$ & $5{,}132{,}548$ & $1.000$ & $0.01997$ & M \\
instrumental convergence & $2$ & $-1{,}047{,}256$ & $0.500$ & $-0.00837$ & M \\
corrigibility & $2$ & $-1{,}055{,}986$ & $0.500$ & $0.01680$ & M \\
value lock in & $1$ & $3{,}002{,}645$ & $1.000$ & $0.02686$ & M \\
sycophancy & $1$ & $1{,}284{,}104$ & $1.000$ & $0.01861$ & M \\
honesty & $1$ & $-1{,}068{,}811$ & $0.000$ & $0.01331$ & M \\
theory of mind & $1$ & $1{,}398{,}770$ & $1.000$ & $0.03494$ & M \\
situational awareness & $1$ & $2{,}604{,}493$ & $1.000$ & $0.02102$ & M \\
moral reasoning & $1$ & $5{,}516{,}737$ & $1.000$ & $0.01738$ & M \\
\bottomrule
\end{tabular}
\end{table}

\subsection{The necessity and sufficiency interventions}
\label{app:results:loop}

Appendix~\ref{app:loop} gives the projector algebra, the holdout construction and the injection
construction. We take misalignment rates here as cluster means over the $83$ evaluation clusters with near-domain
answers excluded, and we take the alignment shift as the paired change in judged alignment score
against the unedited fine-tune, in judge points, so a large positive shift means the arm is more
aligned than the reference. Nothing in this section is on a scale we share with any other section.

\begin{table}[htbp]
\centering\footnotesize
\caption{\textbf{Every arm, judged misalignment.} The interval is a cluster bootstrap over the $83$
evaluation clusters. Coherent is the fraction of responses clearing the coherence floor and is a
fraction of responses; mean alignment and mean coherence are judge scores on a $100$-point scale.
The two injection rows pool $10$ doses including the incoherent ones and are not the dose response;
Table~\ref{tab:j-dose} carries that. Every row in this table is a stored value that nothing has
recomputed except the six misalignment rates marked \textbf{R}.}
\label{tab:j-arms}
\begin{tabular}{>{\raggedright\arraybackslash}p{0.29\textwidth} r r l r r c}
\toprule
Arm & responses & misaligned & interval & coherent & alignment & status \\
\midrule
unedited fine-tune (the reference) & $2100$ & $0.2772$ & $[0.2133,\,0.3455]$ & $1.0000$ & $33.987$ & \textbf{R} \\
carrier held out of the activations & $2100$ & $0.0000$ & $[0.0000,\,0.0000]$ & $1.0000$ & $93.395$ & \textbf{R} \\
matched-rank random subspace held out (the control) & $2100$ & $0.2752$ & $[0.2113,\,0.3410]$ & $1.0000$ & $34.842$ & \textbf{R} \\
holdout, carrier orthogonalized against Assistant & $700$ & $0.0000$ & $[0.0000,\,0.0000]$ & $1.0000$ & $92.419$ & s \\
holdout, two epochs & $700$ & $0.0002$ & $[0.0000,\,0.0007]$ & $1.0000$ & $93.115$ & s \\
unedited fine-tune, two epochs & $700$ & $0.2531$ & $[0.1901,\,0.3230]$ & $1.0000$ & $32.974$ & s \\
holdout, second narrow set & $700$ & $0.0000$ & $[0.0000,\,0.0000]$ & $1.0000$ & $93.245$ & s \\
unedited fine-tune, second narrow set & $700$ & $0.2587$ & $[0.1935,\,0.3218]$ & $1.0000$ & $46.245$ & s \\
holdout during a benign fine-tune & $700$ & $0.0000$ & $[0.0000,\,0.0000]$ & $1.0000$ & $93.978$ & s \\
\midrule
writer fine-tune, no gradient filter & $1400$ & $0.2672$ & $[0.2056,\,0.3328]$ & $1.0000$ & $35.334$ & \textbf{R} \\
writer gradient filtered off the carrier & $1400$ & $0.2655$ & $[0.2025,\,0.3335]$ & $1.0000$ & $34.687$ & \textbf{R} \\
writer gradient filtered onto the carrier & $1400$ & $0.1150$ & $[0.0776,\,0.1551]$ & $1.0000$ & $67.546$ & \textbf{R} \\
\midrule
carrier injected, all doses pooled & $7000$ & $0.0807$ & $[0.0651,\,0.0964]$ & $0.5121$ & $48.316$ & s \\
random vector injected, all doses pooled & $7000$ & $0.0016$ & $[0.0006,\,0.0031]$ & $0.5869$ & $62.080$ & s \\
single-direction organism, carrier removed & $2800$ & $0.0000$ & $[0.0000,\,0.0000]$ & $1.0000$ & $95.113$ & s \\
\midrule
published organism, carrier removed & $700$ & $0.1775$ & $[0.1317,\,0.2283]$ & $0.9871$ & $49.052$ & s \\
published organism, random subspace removed & $700$ & $0.2817$ & $[0.2102,\,0.3575]$ & $1.0000$ & $33.730$ & s \\
published organism, unmodified & $700$ & $0.2788$ & $[0.2136,\,0.3493]$ & $1.0000$ & $35.775$ & s \\
\bottomrule
\end{tabular}
\end{table}

\begin{table}[htbp]
\centering\footnotesize
\caption{\textbf{Every arm, the paired alignment shift and the reduction in misalignment.} The shift
is in judge points against the unedited fine-tune and its null is zero, which the interval addresses.
Holm correction is across the family of arms. Reduction is the fraction of the reference arm's
misalignment rate removed, so its null is zero and one is complete removal. The reference arm has no
row because it is the comparison. Cliff's $\delta$ gates nothing.}
\label{tab:j-shift}
\begin{tabular}{>{\raggedright\arraybackslash}p{0.26\textwidth} r l r r r c}
\toprule
Arm & alignment shift & interval & Holm $p$ & Cliff's $\delta$ & reduction & status \\
\midrule
carrier held out of the activations & $59.408$ & $[56.982,\,61.987]$ & $0.001700$ & $1.0000$ & $1.0000$ & s \\
matched-rank random subspace held out (the control) & $0.855$ & $[-0.338,\,2.157]$ & $0.532447$ & $0.0222$ & $0.0071$ & s \\
holdout, carrier orthogonalized against Assistant & $58.432$ & $[55.911,\,61.163]$ & $0.001700$ & $1.0000$ & $1.0000$ & s \\
holdout, two epochs & $59.128$ & $[56.701,\,61.757]$ & $0.001700$ & $1.0000$ & $0.9991$ & s \\
unedited fine-tune, two epochs & $-1.013$ & $[-2.785,\,0.750]$ & $1.000000$ & $-0.0771$ & $0.0868$ & s \\
holdout, second narrow set & $59.258$ & $[56.787,\,61.927]$ & $0.001700$ & $1.0000$ & $1.0000$ & s \\
unedited fine-tune, second narrow set & $12.258$ & $[8.871,\,15.619]$ & $0.001700$ & $0.3625$ & $0.0668$ & s \\
holdout during a benign fine-tune & $59.991$ & $[57.478,\,62.677]$ & $0.001700$ & $1.0000$ & $1.0000$ & s \\
\midrule
writer fine-tune, no gradient filter & $1.347$ & $[-0.519,\,3.095]$ & $0.482352$ & $0.0566$ & $0.0361$ & s \\
writer gradient filtered off the carrier & $0.700$ & $[-1.186,\,2.652]$ & $0.752925$ & $0.0136$ & $0.0420$ & s \\
writer gradient filtered onto the carrier & $33.559$ & $[29.235,\,37.485]$ & $0.001700$ & $0.8254$ & $0.5851$ & s \\
\midrule
carrier injected, all doses pooled & $14.329$ & $[11.597,\,17.142]$ & $0.001700$ & $0.6975$ & $0.7089$ & s \\
random vector injected, all doses pooled & $28.093$ & $[25.970,\,30.425]$ & $0.001700$ & $0.9457$ & $0.9942$ & s \\
single-direction organism, carrier removed & $61.126$ & $[58.630,\,63.733]$ & $0.001700$ & $1.0000$ & $1.0000$ & s \\
\midrule
published organism, carrier removed & $15.065$ & $[11.420,\,18.976]$ & $0.001700$ & $0.4917$ & $0.3598$ & s \\
published organism, random subspace removed & $-0.257$ & $[-1.755,\,1.255]$ & $1.000000$ & $-0.0189$ & $-0.0162$ & s \\
published organism, unmodified & $1.788$ & $[-1.045,\,4.560]$ & $0.532447$ & $-0.0073$ & $-0.0058$ & s \\
\bottomrule
\end{tabular}
\end{table}

\begin{table}[htbp]
\centering\footnotesize
\caption{\textbf{Narrow-task adherence, where we measured it.} This is a judged rate on the
organism's own narrow probes and it is not comparable with any log-probability margin anywhere in
this paper. Denominators differ by an order of magnitude between arms, and two of them are too small
to carry a usable interval, since the number of prompt clusters behind these rates runs from $2$ to
$50$. The drop is in percentage points against the reference arm, and a bar fixed before the run
required it to stay at or under $5$. It came in at $90.215$ for the holdout arm.}
\label{tab:j-narrow}
\begin{tabular}{>{\raggedright\arraybackslash}p{0.32\textwidth} r l r r c}
\toprule
Arm & adherence & interval & responses & drop & status \\
\midrule
unedited fine-tune (the reference) & $0.9021$ & $[0.8384,\,0.9523]$ & $824$ & $0.000$ & \textbf{R} \\
carrier held out of the activations & $0.0000$ & $[0.0000,\,0.0000]$ & $236$ & $90.215$ & \textbf{R} \\
matched-rank random subspace held out (the control) & $0.8942$ & $[0.8391,\,0.9430]$ & $781$ & $0.796$ & s \\
holdout, carrier orthogonalized against Assistant & $0.0016$ & $[0.0000,\,0.0048]$ & $83$ & $90.055$ & s \\
holdout, two epochs & $0.0000$ & $[0.0000,\,0.0000]$ & $78$ & $90.215$ & s \\
unedited fine-tune, two epochs & $0.8365$ & $[0.7467,\,0.9180]$ & $308$ & $6.566$ & s \\
holdout, second narrow set & $0.0000$ & $[0.0000,\,0.0000]$ & $5$ & $90.215$ & s \\
unedited fine-tune, second narrow set & $0.8000$ & $[0.4000,\,1.0000]$ & $13$ & $10.215$ & s \\
holdout during a benign fine-tune & $1.0000$ & $[1.0000,\,1.0000]$ & $39$ & $-9.785$ & s \\
writer fine-tune, no gradient filter & $0.8010$ & $[0.7211,\,0.8762]$ & $556$ & $10.119$ & s \\
writer gradient filtered off the carrier & $0.8191$ & $[0.7408,\,0.8893]$ & $582$ & $8.307$ & s \\
writer gradient filtered onto the carrier & $0.5075$ & $[0.3561,\,0.6636]$ & $175$ & $39.465$ & s \\
\bottomrule
\end{tabular}
\end{table}

\begin{table}[htbp]
\centering\small
\caption{\textbf{Injection dose response, on the untouched model with no fine-tuning.} The
norm-matched random vector is the control and is printed beside the carrier at every dose. Rates are
cluster means over the $83$ evaluation clusters on $700$ responses per dose per arm. The upper block
is the region fixed before the run in which the slope is fitted; in the lower block the coherence
floor removes most of the denominator and those rates are not comparable with the block above.}
\label{tab:j-dose}
\begin{tabular}{l rrr rrr c}
\toprule
& \multicolumn{3}{c}{carrier} & \multicolumn{3}{c}{norm-matched random} & \\
\cmidrule(lr){2-4}\cmidrule(lr){5-7}
$\alpha$ & misaligned & coherent & alignment & misaligned & coherent & alignment & status \\
\midrule
$0.10$ & $0.0000$ & $1.0000$ & $93.980$ & $0.0000$ & $1.0000$ & $95.026$ & \textbf{R}, s \\
$0.15$ & $0.0030$ & $1.0000$ & $90.949$ & $0.0000$ & $1.0000$ & $95.168$ & \textbf{R}, s \\
$0.20$ & $0.0559$ & $1.0000$ & $79.924$ & $0.0000$ & $1.0000$ & $95.087$ & \textbf{R}, s \\
$0.25$ & $0.1984$ & $1.0000$ & $61.525$ & $0.0000$ & $1.0000$ & $94.801$ & \textbf{R}, s \\
$0.30$ & $0.4536$ & $0.9814$ & $38.823$ & $0.0000$ & $1.0000$ & $94.698$ & \textbf{R}, s \\
\midrule
$0.50$ & $0.0539$ & $0.0471$ & $7.000$ & $0.0160$ & $0.8686$ & $72.093$ & \textbf{R}, s \\
$0.75$ & $0.0000$ & $0.0000$ & $28.553$ & $0.0000$ & $0.0000$ & $21.273$ & \textbf{R}, s \\
$1.00$ & $0.0030$ & $0.0014$ & $28.558$ & $0.0000$ & $0.0000$ & $17.076$ & \textbf{R}, s \\
$1.25$ & $0.0163$ & $0.0343$ & $25.704$ & $0.0000$ & $0.0000$ & $14.265$ & \textbf{R}, s \\
$1.50$ & $0.0228$ & $0.0571$ & $28.140$ & $0.0000$ & $0.0000$ & $21.313$ & \textbf{R}, s \\
\bottomrule
\end{tabular}
\end{table}

\begin{table}[htbp]
\centering\small
\caption{\textbf{The same five in-budget doses under all three aggregation conventions.} The
generations are identical in all three rows and were confirmed byte-identical; only the aggregation
unit and the rule about near-domain answers change. The first row is the convention used everywhere
in this paper. The upper block holds slopes of misalignment on dose and the lower block holds the
ratio between two of them and the reference arm's own rate under two units, so the value column
changes meaning at the rule. The spread is the measured size of the comparability hazard, and it is
why no rate here may be placed beside a rate from another table.}
\label{tab:j-doseslope}
\begin{tabular}{l r l r c}
\toprule
Convention & value & interval & control & status \\
\midrule
prompt cluster, near-domain excluded & $2.2053$ & $[0.4967,\,3.9771]$ & $0.0000$ & \textbf{R} \\
response level, near-domain excluded & $2.5600$ & $[0.6036,\,4.3571]$ & none & \textbf{R} \\
response level, near-domain included & $2.9714$ & $[0.7446,\,4.9857]$ & none & \textbf{R} \\
\midrule
response over cluster, same rows & $1.1608$ & none & none & \textbf{R} \\
reference arm, prompt cluster & $0.27719$ & none & none & \textbf{R} \\
reference arm, response level & $0.23048$ & none & none & \textbf{R} \\
\bottomrule
\end{tabular}
\end{table}

\begin{table}[htbp]
\centering\small
\caption{\textbf{A per-example geometric score computed on the untouched model, by narrow training
set.} This score was intended to order the narrow sets by how much broad misalignment each produces,
with the educational set as the null. It does not. Every mean is negative, and the two sets that
produce the most broad misalignment do not rank above the null. This score failed its liveness gate;
the necessity result rests on the interventions above.}
\label{tab:j-geomscore}
\begin{tabular}{l r r l r c}
\toprule
Narrow training set & examples & mean & interval & standard deviation & status \\
\midrule
insecure code & $128$ & $-0.00345$ & $[-0.00400,\,-0.00293]$ & $0.00317$ & s \\
risky financial advice & $128$ & $-0.04020$ & $[-0.04289,\,-0.03765]$ & $0.01575$ & s \\
extreme sports & $128$ & $-0.01682$ & $[-0.01851,\,-0.01525]$ & $0.00971$ & s \\
educational (the null) & $128$ & $-0.00282$ & $[-0.00321,\,-0.00247]$ & $0.00215$ & s \\
benign medical & $128$ & $-0.00147$ & $[-0.00229,\,-0.00075]$ & $0.00430$ & s \\
\bottomrule
\end{tabular}
\end{table}

\subsection{Sharing and transport between domains}
\label{app:results:sharing}

The sharing index $\Sigma$ compares the read-channel shifts that $3$ separately trained organisms
realize, after we residualize each against the common mode any pair of shifts would share
(Appendix~\ref{app:removal}). Its null is empirical rather than analytic. We resample the shifts to
break the pairing and recompute the index, which lands well above the $1/\sqrt{d}$ floor a naive
calculation would suggest. We measure transport on $2$ anchorable domain pairs, and the third
organism has no usable own-domain probes, so we exclude it rather than count it as a null.

\begin{table}[htbp]
\centering\small
\caption{\textbf{Read-channel sharing of $3$ organisms' realized shifts, and transport between $2$
domain pairs.} The sharing index is a mean over $24$ measured layers. The residualizer makes the null
informative, driving the resampled reference down from $0.159$ to $0.110$ while leaving the measured
index almost unchanged, so the common mode is not carrying the index. Transport is in nats on the
receiving domain's own probes. The per-response source for both was not persisted, so the sharing index
is recomputed from the stored per-layer arrays and the transport values are carried as stored.}
\label{tab:j-sharing}
\begin{tabular}{l r l r c}
\toprule
Quantity & value & interval & null or reference & status \\
\midrule
sharing index, residualized & $0.8955$ & none & $0.10980$ & \textbf{R} \\
sharing index, raw & $0.9023$ & none & $0.15866$ & \textbf{R} \\
resampled reference, upper five percent & $0.14021$ & none & none & s \\
$1/\sqrt{d}$ floor (diagnostic only) & $0.013975$ & none & none & M \\
first and last measured layer & $0.8356$, $0.9402$ & none & none & M \\
paired permutation $p$ against the reference & $9.999\e{-5}$ & none & none & s \\
\midrule
transport, finance organism onto medicine & $50.592$ & $[28.115,\,75.047]$ & none & s \\
transport, medicine organism onto finance & $10.803$ & $[1.274,\,20.421]$ & none & s \\
\midrule
probes with no own domain & $150$ & none & none & M \\
finance probes & $28$ & none & none & M \\
medicine probes & $15$ & none & none & M \\
code probes & $2$ & none & none & M \\
probes in total & $195$ & none & none & M \\
\bottomrule
\end{tabular}
\end{table}

\subsection{The domain-count decomposition}
\label{app:results:decomp}

Appendix~\ref{app:decomp} defines the budget, the measured additive null and the two subtractions.
$T^{+}$ and $T^{-}$ are both changes against the untouched model, so both have a null of zero by
construction, and we inherit the sign convention on $T^{-}$ from the instrument, under which a
negative $T^{-}$ means the aligned continuation became more likely. Every value is a seed average at the unit
of the $52$ unique prompts.

\begin{table}[htbp]
\centering\small
\caption{\textbf{Single-domain legs, transport onto misaligned continuations and the aligned
half.} $T^{+}$ carries a cluster-bootstrap interval; $T^{-}$ carries none, because none was
computed for any leg, so the aligned half is descriptive throughout. Transport is flat across a
factor of eight in budget, which is what puts the additive null out of reach of the mixtures.}
\label{tab:j-legs1}
\begin{tabular}{l r l r r c}
\toprule
Training leg & $T^{+}$ & interval & $T^{-}$ & seeds & status \\
\midrule
code, full budget & $111.844$ & $[105.205,\,118.909]$ & $-66.046$ & $1$ & \textbf{R} \\
medicine, full budget & $120.845$ & $[114.842,\,127.020]$ & $-84.342$ & $1$ & \textbf{R} \\
finance, full budget & $120.257$ & $[110.903,\,129.737]$ & $-73.600$ & $1$ & \textbf{R} \\
sports, full budget & $117.730$ & $[109.654,\,125.861]$ & $-77.061$ & $1$ & \textbf{R} \\
code, half budget & $110.266$ & $[103.859,\,117.079]$ & $-65.753$ & $2$ & \textbf{R} \\
medicine, half budget & $121.767$ & $[116.153,\,127.648]$ & $-83.184$ & $1$ & \textbf{R} \\
finance, half budget & $121.223$ & $[113.028,\,129.503]$ & $-77.397$ & $1$ & \textbf{R} \\
sports, half budget & $119.466$ & $[112.646,\,126.392]$ & $-80.696$ & $1$ & \textbf{R} \\
code, quarter budget & $108.575$ & $[102.245,\,115.390]$ & $-67.292$ & $1$ & \textbf{R} \\
medicine, quarter budget & $117.943$ & $[112.453,\,123.639]$ & $-79.823$ & $1$ & \textbf{R} \\
finance, quarter budget & $122.772$ & $[117.353,\,128.322]$ & $-78.536$ & $1$ & \textbf{R} \\
sports, quarter budget & $122.908$ & $[117.370,\,128.592]$ & $-80.726$ & $1$ & \textbf{R} \\
code, eighth budget & $105.571$ & $[99.223,\,112.385]$ & $-65.990$ & $1$ & \textbf{R} \\
\bottomrule
\end{tabular}
\end{table}

\begin{table}[htbp]
\centering\small
\caption{\textbf{Mixtures, mechanical merges and the schedule controls.} A merged row is a
weight-space sum of separately trained single-domain adapters with no training of its own, which
isolates the readout's response to superposed weights from the effect of co-training. Across all $26$
legs $T^{-}$ is negative, with aligned-half gains running from $62.101$ nats on the merged four-domain
adapter to $84.342$ nats on the full-budget medicine leg.}
\label{tab:j-legs2}
\begin{tabular}{l r l r r c}
\toprule
Training leg & $T^{+}$ & interval & $T^{-}$ & seeds & status \\
\midrule
code and medicine & $123.881$ & $[118.122,\,129.905]$ & $-83.150$ & $2$ & \textbf{R} \\
code and finance & $123.932$ & $[116.009,\,132.157]$ & $-74.017$ & $2$ & \textbf{R} \\
finance and sports & $121.790$ & $[113.233,\,130.573]$ & $-76.771$ & $2$ & \textbf{R} \\
all four domains & $128.232$ & $[121.710,\,134.862]$ & $-83.873$ & $3$ & \textbf{R} \\
merged, code and medicine & $123.968$ & $[117.976,\,130.445]$ & $-78.551$ & $1$ & \textbf{R} \\
merged, code and finance & $124.055$ & $[118.056,\,130.245]$ & $-74.203$ & $1$ & \textbf{R} \\
merged, finance and sports & $115.832$ & $[106.986,\,124.715]$ & $-67.844$ & $1$ & \textbf{R} \\
merged, all four domains & $115.643$ & $[109.643,\,121.978]$ & $-62.101$ & $1$ & \textbf{R} \\
benign four-domain mixture & $115.219$ & $[109.345,\,121.455]$ & $-77.205$ & $2$ & \textbf{R} \\
code and medicine, blocked schedule & $122.507$ & $[116.538,\,128.817]$ & $-82.177$ & $1$ & \textbf{R} \\
bad then benign & $122.464$ & $[116.488,\,128.749]$ & $-82.088$ & $1$ & \textbf{R} \\
benign then bad & $122.737$ & $[116.791,\,129.034]$ & $-82.130$ & $1$ & \textbf{R} \\
benign last & $112.234$ & $[106.696,\,118.201]$ & $-76.525$ & $1$ & \textbf{R} \\
\bottomrule
\end{tabular}
\end{table}

\begin{table}[htbp]
\centering\footnotesize
\caption{\textbf{The decomposition, per mixture, on the transport half.} The raw residual against the
measured additive null is dominated by the readout's response to superposed weights and is not
interpretable on its own; the training interaction is the difference between the co-trained mixture
and its own mechanical merge, and it is the primary statistic. Its null is zero, tested by sign-flip
permutation at $10^4$ draws. The identity that the raw residual is the sum of the other two columns
holds on every row. Only the four-domain mixture and one two-domain pair separate from zero, and both
sit far below the magnitude bar of $47.272$ nats fixed before the run. The four-domain row's $p$ is
stored rounded to $1.000\e{-4}$ and the two-domain row's at the unrounded $9.999\e{-5}$; these are
the same resolution floor of one test and not two different values.}
\label{tab:j-decomp}
\begin{tabular}{>{\raggedright\arraybackslash}p{0.155\textwidth} r r r l r r c}
\toprule
Mixture & raw residual & readout & interaction & interval & $p$ & $\delta$ & status \\
\midrule
code and medicine & $-108.153$ & $-108.065$ & $-0.087$ & $[-2.220,\,2.028]$ & $0.5328$ & $0.0370$ & s \\
code and finance & $-107.558$ & $-107.435$ & $-0.123$ & $[-4.796,\,4.366]$ & $0.5177$ & $0.0141$ & s \\
finance and sports & $-118.899$ & $-124.857$ & $5.957$ & $[4.734,\,7.206]$ & $9.999\e{-5}$ & $0.9882$ & s \\
all four domains & $-343.966$ & $-356.555$ & $12.589$ & $[9.255,\,15.980]$ & $1.000\e{-4}$ & $0.8365$ & \textbf{R} \\
code and medicine, blocked & $-109.526$ & $-108.065$ & $-1.461$ & $[-4.017,\,1.009]$ & $0.8763$ & $0.0148$ & s \\
\bottomrule
\end{tabular}
\end{table}

\begin{table}[htbp]
\centering\small
\caption{\textbf{Both subtractions at $4$ domains, on both readouts.} The transport half was fixed
as primary before the run. Changing the readout to the full margin turns the merge subtraction's sign
over and roughly halves the benign subtraction while leaving its sign and its interval's exclusion of
zero intact. The mechanism is the aligned-half gap in the lower block. The mechanical merge gains
$21.773$ nats less on the aligned half than the co-trained mixture does, which outweighs the
transport surplus, while the benign mixture's gap is under a third of that. The one-sided $p$ of $1$
on the full-margin merge subtraction is the direction fixed before the run failing, not an absence of
effect.}
\label{tab:j-halves}
\begin{tabular}{>{\raggedright\arraybackslash}p{0.33\textwidth} r l r r c}
\toprule
Statistic and readout & value & interval & $p$ & Cliff's $\delta$ & status \\
\midrule
training interaction, transport half & $12.589$ & $[9.255,\,15.980]$ & $1.000\e{-4}$ & $0.8365$ & \textbf{R} \\
training interaction, full margin & $-9.184$ & $[-12.910,\,-5.415]$ & $9.999\e{-5}$ & $-0.5999$ & \textbf{R} \\
\quad its one-sided $p$ in the fixed direction & & & $1.0$ & & \textbf{R} \\
\midrule
benign subtraction, transport half & $13.013$ & $[8.502,\,17.471]$ & $9.999\e{-5}$ & $0.7145$ & \textbf{R} \\
benign subtraction, full margin & $6.345$ & $[1.318,\,11.294]$ & $0.007099$ & $0.3735$ & \textbf{R} \\
\midrule
aligned-half gap against the merge & $-21.773$ & none & none & none & \textbf{R} \\
aligned-half gap against the benign mixture & $-6.668$ & none & none & none & \textbf{R} \\
full margin, co-trained mixture & $44.358$ & none & none & none & \textbf{R} \\
full margin, mechanical merge & $53.542$ & none & none & none & \textbf{R} \\
full margin, benign mixture & $38.013$ & none & none & none & \textbf{R} \\
\bottomrule
\end{tabular}
\end{table}

\begin{table}[htbp]
\centering\small
\caption{\textbf{Judged misalignment per leg.} The rate is over $670$ responses and the interval is
at the unit of the $52$ unique prompts, so the two are on different footings by construction. The
rate is a response-level fraction and the interval is a prompt-level one, and the wider of the two
intervals is reported. The benign four-domain mixture is the matched control. The judged
rates are not ordered by domain count, and the hottest leg in the grid is a two-domain mixture whose
training interaction is flat. Every value here is a judge product whose per-response scores were not
persisted, so none has been recomputed.}
\label{tab:j-dcjudge}
\begin{tabular}{l r l r r c}
\toprule
Training leg & misaligned & prompt-unit interval & alignment & coherence & status \\
\midrule
code alone & $0.0179$ & $[0.0030,\,0.0991]$ & $83.746$ & $91.178$ & s \\
medicine alone & $0.1493$ & $[0.0769,\,0.2699]$ & $65.060$ & $89.919$ & s \\
finance alone & $0.2985$ & $[0.1915,\,0.4332]$ & $41.142$ & $86.915$ & s \\
sports alone & $0.1522$ & $[0.0789,\,0.2734]$ & $55.552$ & $87.896$ & s \\
code and medicine & $0.1761$ & $[0.0960,\,0.3008]$ & $62.142$ & $89.330$ & s \\
code and finance & $0.4104$ & $[0.2874,\,0.5458]$ & $36.597$ & $86.579$ & s \\
finance and sports & $0.2448$ & $[0.1482,\,0.3765]$ & $46.440$ & $87.373$ & s \\
all four domains & $0.2687$ & $[0.1672,\,0.4019]$ & $48.799$ & $87.921$ & s \\
benign four-domain mixture & $0.0104$ & $[0.0012,\,0.0871]$ & $87.530$ & $92.988$ & s \\
code and medicine, blocked & $0.2060$ & $[0.1182,\,0.3342]$ & $58.657$ & $88.803$ & s \\
bad then benign & $0.1955$ & $[0.1104,\,0.3226]$ & $58.336$ & $88.594$ & s \\
benign then bad & $0.2164$ & $[0.1262,\,0.3457]$ & $58.500$ & $88.691$ & s \\
benign last & $0.0000$ & $[0.0000,\,0.0688]$ & $91.716$ & $94.099$ & s \\
\bottomrule
\end{tabular}
\end{table}

\begin{table}[htbp]
\centering\small
\caption{\textbf{The bar, the seed variance, and the instrument checks.} The magnitude bar is $0.40$
of the mean single-domain transport at half budget and was fixed before the run; a $44.106$ figure
elsewhere is a superseded estimate from a smoke run and is not the bar. The $3$ published organisms were
measured through the same stack before any leg was trained, and all $3$ clear the zero floor. We fit no
growth slope through the interaction points, because $2$ domain counts do not make a curve and a
resampling bootstrap over $4$ points cannot express the null; the slope our own analysis returned is
recorded in the appendix on validity and not used.}
\label{tab:j-dcscalars}
\begin{tabular}{l r l c}
\toprule
Quantity & value & interval & status \\
\midrule
magnitude bar & $47.272$ & none & \textbf{R} \\
\quad mean single-domain transport behind it & $118.181$ & none & M \\
\quad fraction taken of it & $0.40$ & none & s \\
across-seed standard deviation & $0.2076$ & none & M \\
\midrule
published medicine organism, $T^{+}$ & $119.907$ & $[113.422,\,126.437]$ & s \\
\quad its judged misalignment & $0.0667$ & none & s \\
published finance organism, $T^{+}$ & $119.459$ & $[109.478,\,129.698]$ & s \\
\quad its judged misalignment & $0.2917$ & none & s \\
published sports organism, $T^{+}$ & $117.881$ & $[109.540,\,126.267]$ & s \\
\quad its judged misalignment & $0.1750$ & none & s \\
\midrule
judged rate at four domains, observed & $0.2687$ & none & M \\
gap against what transport dose predicts & $0.1734$ & none & M \\
\bottomrule
\end{tabular}
\end{table}

\subsection{The removal edits and the reconstitution check}
\label{app:results:removal}

We give the $3$ edits and the projector they share in Appendix~\ref{app:removal}. A drop is the
margin before the edit minus the margin after, so a positive drop means the edit lowered the margin.
Selectivity is the broad drop minus the expressive drop and its null is zero, and we run a matched
random edit alongside to give an empirical value for that null.

\begin{table}[htbp]
\centering\small
\caption{\textbf{The selectivity edit and its $3$ matched controls.} A selective removal would show
a broad drop much larger than the expressive drop, giving a large positive selectivity. The measured
selectivity is negative and its interval spans zero; expressed in units of the realized standard
deviation it is $[-0.334,\,0.139]$, inside the band of $\pm 0.4$ fixed before the run, so this is an
equivalence result at the sample size actually reached and not merely a failure to reject. The $2$
leading-direction controls show the instrument can produce a large signed selectivity when one is
there.}
\label{tab:j-selectivity}
\begin{tabular}{l r r r c}
\toprule
Edit & broad drop & expressive drop & selectivity & status \\
\midrule
shared read core removed & $0.2333$ & $0.4599$ & $-0.22662$ & \textbf{R} \\
matched random subspace (the null) & & & $-0.12199$ & M \\
leading local direction (control) & $0.0476$ & $0.7987$ & $-0.7511$ & M \\
leading shared direction (control) & $0.4238$ & $0.8331$ & $-0.4093$ & M \\
\midrule
selectivity interval & \multicolumn{3}{l}{$[-0.8378,\,0.3470]$} & s \\
label-permutation $p$ & \multicolumn{3}{l}{$0.42933$} & s \\
Cliff's $\delta$, broad against expressive & \multicolumn{3}{l}{$-0.09821$} & M \\
smallest effect resolvable, in standard deviations & \multicolumn{3}{l}{$0.3354$} & s \\
probes behind the two drops & \multicolumn{3}{l}{$195$ and $120$} & M \\
\bottomrule
\end{tabular}
\end{table}

\begin{table}[htbp]
\centering\footnotesize
\caption{\textbf{Removing the shared write component from each published organism, one organism at a
time.} The unit of inference is the organism, so these are $3$ separate results and a pooled value
would be descriptive only. A selective removal predicts a broad drop larger than the narrow drop, so
a positive difference. Two of the three are negative and the third is small, with both of its
component intervals wide. The difference itself carries no interval, only a permutation $p$, so the
interval columns belong to the two drops that form it.}
\label{tab:j-organism}
\begin{tabular}{l r l r l r r c}
\toprule
Organism & broad drop & interval & narrow drop & interval & difference & $p$ & status \\
\midrule
medicine & $0.0322$ & $[-0.042,\,0.106]$ & $0.2665$ & $[0.080,\,0.447]$ & $-0.23431$ & $0.0161$ & M \\
finance & $0.1513$ & $[0.058,\,0.243]$ & $0.1201$ & $[-0.147,\,0.384]$ & $0.03121$ & $0.8031$ & M \\
sports & $0.0859$ & $[0.010,\,0.163]$ & $0.2645$ & $[0.034,\,0.507]$ & $-0.17864$ & $0.0880$ & M \\
\midrule
medicine, probes & \multicolumn{2}{l}{$195$ and $38$} & \multicolumn{2}{l}{margin rise $30.086$} & & & M \\
finance, probes & \multicolumn{2}{l}{$195$ and $38$} & \multicolumn{2}{l}{margin rise $30.926$} & & & M \\
sports, probes & \multicolumn{2}{l}{$195$ and $38$} & \multicolumn{2}{l}{margin rise $30.648$} & & & M \\
\bottomrule
\end{tabular}
\end{table}

\begin{table}[htbp]
\centering\small
\caption{\textbf{How much of each organism's update the edit actually removed.} The shared component
is a few thousandths of the update's Frobenius mass, and the mass a matched random component removes
is the same to eight significant figures, so the operation does not distinguish the shared direction
from noise. Both masses are products of the run that produced them and cannot be recomputed without
the adapter weights, so neither carries an interval.}
\label{tab:j-frobenius}
\begin{tabular}{l r r r c}
\toprule
Organism & shared mass & matched random mass & fraction of the update & status \\
\midrule
medicine & $0.24440166$ & $0.24440166$ & $0.003054$ & s \\
finance & $0.24815236$ & $0.24815236$ & $0.003214$ & s \\
sports & $0.24584799$ & $0.24584799$ & $0.002933$ & s \\
\midrule
relative difference of the two masses & \multicolumn{3}{l}{$0.0$ on all three} & M \\
\bottomrule
\end{tabular}
\end{table}

\begin{table}[htbp]
\centering\small
\caption{\textbf{Re-lighting the defended model by injection.} A removal predicts that the defended
model needs a larger dose and re-lights along a shallower slope. It needs neither. The onset dose is
identical across the $3$ carrier arms and the defended slope is the steeper. The random direction
does not reach onset at any dose in the grid, which is the control. The slope ratio's interval
excludes one from above. The onset is measured relative to each model's own margin at zero dose;
against an absolute threshold no arm reaches onset at all, and that threshold was corrected before
these values were certified.}
\label{tab:j-relight}
\begin{tabular}{l r r r c}
\toprule
Injection arm & slope & onset dose & probes & status \\
\midrule
undefended model & $62.923$ & $0.15$ & $75$ & M \\
defended model, carrier fixed & $76.365$ & $0.15$ & $75$ & M \\
defended model, carrier re-extracted & $76.070$ & $0.15$ & $75$ & M \\
undefended model, carrier re-extracted & $63.173$ & $0.15$ & $75$ & M \\
defended model, random direction & $1.111$ & unreached & $75$ & M \\
\midrule
slope ratio, defended over undefended & $1.2136$ & $[1.1231,\,1.3662]$ & $75$ & \textbf{R} \\
change in onset dose & $0.00$ & none & & M \\
defended model perplexity & $17.300$ & none & & s \\
\bottomrule
\end{tabular}
\end{table}

\begin{table}[htbp]
\centering\small
\caption{\textbf{The re-extracted carrier, measured inside the span that was cleared.} After the edit,
re-running the extraction on the defended model recovers a direction that lies almost entirely inside
the subspace the edit removed and points almost exactly along the original. This is a statement about
what can be read out of the model's activations, not about the weights regrowing. Neither quantity
carries an interval or a matched control, and both are products of the run that produced them.}
\label{tab:j-reext}
\begin{tabular}{l r r c}
\toprule
Layer & share inside the cleared span & cosine to the original & status \\
\midrule
18 & $0.9741$ & $0.9870$ & s \\
24 & $0.9693$ & $0.9845$ & s \\
30 & $0.9756$ & $0.9877$ & s \\
\bottomrule
\end{tabular}
\end{table}

\begin{table}[htbp]
\centering\small
\caption{\textbf{Two instrument checks, one of which did not resolve.} The upper block is the carrier
liveness check. The carrier moves the margin steeply and the matched random direction does not move
it the same way, but the random direction is not flat either and coherence degrades with dose, so
this direction is active without being a clean isolable control. The lower block is a window
measurement that has no value at all. Coherence collapses at a dose of $0.1$, below any dose at which
misalignment onsets on an absolute threshold, so we never bracketed the window. It is recorded as
undefined rather than as a zero.}
\label{tab:j-instr}
\begin{tabular}{l r l c}
\toprule
Quantity & value & interval & status \\
\midrule
carrier slope & $78.467$ & $[62.019,\,95.903]$ & M \\
matched random slope (the control) & $-28.155$ & $[-41.425,\,-15.261]$ & M \\
difference of the two & $106.621$ & none & M \\
response entropy at zero dose and at the largest & $0.9357$ and $1.9275$ & none & M \\
probes behind it & $195$ & none & M \\
\midrule
dose at which coherence collapses & $0.10$ & none & M \\
dose at which misalignment onsets & unreached in grid & none & M \\
the window between them & no value; never bracketed & none & M \\
\bottomrule
\end{tabular}
\end{table}

\subsection{The loss geometry}
\label{app:results:geometry}

We give the behavioral loss, the exponent estimator and the planted controls in
Appendix~\ref{app:geometry}. The exponent $\beta$ is the order of the zero of that loss along a unit
direction, so $\beta = 2$ is a regular minimum and $\beta > 2$ is the signature that would indicate
a degenerate one. We plant $2$ directions with known answers and estimate them with the same code,
and those are the nulls we judge this reading against.

\begin{table}[htbp]
\centering\small
\caption{\textbf{The curvature exponent, on both channels, against planted directions with known
answers.} The two planted directions return the values they were built to return, which is what
licenses reading the rest. Every measured direction returns two, on both channels and for all three
contrast types, and the permutation test against the matched random direction does not separate them.
The estimator can distinguish two from four and finds two everywhere it is pointed, so this reading
supports a regular minimum and gives no evidence of a degenerate one.}
\label{tab:j-beta}
\begin{tabular}{l r r r c}
\toprule
Direction & $\beta$ & permutation $p$ & Cliff's $\delta$ & status \\
\midrule
planted regular direction & $2.0$ & none & none & M \\
planted quartic direction & $4.0$ & none & none & M \\
\midrule
write core & $2.00083$ & $0.3124$ & $-0.03125$ & M \\
matched random write directions (the null) & $2.00166$ & & & M \\
\midrule
read carrier & $2.00117$ & none & none & s \\
shared read core & $1.99936$ & none & none & s \\
\midrule
read core, persona contrast & $1.99989$ & & & M \\
read core, style contrast & $2.00095$ & $0.6595$ & & M \\
read core, topic contrast & $2.00119$ & $0.8474$ & & M \\
\quad the two after Holm & & $1.0$ and $1.0$ & & M \\
\bottomrule
\end{tabular}
\end{table}

\begin{table}[htbp]
\centering\footnotesize
\caption{\textbf{How the curvature is distributed, and how much volume it encloses.} The
participation ratio counts how many of the $8$ directions carry the curvature; the write core
concentrates it in fewer directions than a matched random core does, with one column carrying most of
it, so the low participation is concentration and not flatness. The slice coefficient is a volume
reading whose interval is the spread across sampling temperatures. Both arms sit far below the
reference value a regular minimum of this rank would give, so this channel cannot separate a regular
from a degenerate minimum here and the reading is reported as a confirmation rather than as evidence.
The two ratios in the last block are different quantities in different units and are never averaged
or substituted for one another.}
\label{tab:j-volume}
\setlength{\tabcolsep}{4.5pt}
\begin{tabular}{>{\raggedright\arraybackslash}p{0.25\textwidth} r l r c}
\toprule
Quantity & value & interval & reference or null & status \\
\midrule
participation ratio, write core & $3.5494$ & none & $7.1238$ & \textbf{R} \\
\quad largest per-direction curvature in it & $1.012\e{-15}$ & none & none & s \\
\midrule
slice coefficient, write core & $5.6897\e{-4}$ & $[5.021\e{-5},\,3.357\e{-4}]$ & $4.0$ & s \\
slice coefficient, random core & $5.7413\e{-5}$ & none & $4.0$ & s \\
\quad each as a fraction of it & $1.4224\e{-4}$, $1.4353\e{-5}$ & none & none & M \\
sampler convergence and escapes & $0.9958$, $0.0$ & none & none & s \\
\midrule
ratio of the two volumes & $9.9101$ & none & none & \textbf{R} \\
ratio of the summed per-direction curvatures & $6.9202$ & none & none & \textbf{R} \\
gap from the reduced-precision setting & $3.9539$ & none & none & M \\
\bottomrule
\end{tabular}
\end{table}

\subsection{The interventions applied during fine-tuning}
\label{app:results:training}

Two of our interventions act while the fine-tune runs rather than on a finished model. We constrain
the shared write core during training (Appendix~\ref{app:routing}), and we reframe the narrow
training data so that the same completions are presented as a different author's
(Appendix~\ref{app:removal}). We read the first as a change in the broad margin and the second as a
change in the first-step reading.

\begin{table}[htbp]
\centering\footnotesize
\caption{\textbf{Constraining the shared write core during a narrow fine-tune.} Margins are means
over $195$ broad probes and $120$ narrow probes and are in nats. The constrained arm's broad margin
falls further than a matched random constraint's does, by a factor of $3.628$, and the sign-flip test
on it clears the conventional level. The point estimate of that fall carries no interval, because none was
computed for the constrained arm while the random arm's is on the record. The weaker of the two
readings is reported as primary. The narrow margin falls by more than half at the same
time, so the effect is not free. The judged rates in the lower block cover two arms; the random arm's
generations were not judged, so the judged comparison has no matched control.}
\label{tab:j-stiffen}
\begin{tabular}{>{\raggedright\arraybackslash}p{0.28\textwidth} r r l c}
\toprule
Quantity & value & control & interval & status \\
\midrule
broad margin, unconstrained arm & $-34.6103$ & & none & M \\
broad margin, write core constrained & $-35.1511$ & $-34.7594$ & none & M \\
change in the broad margin & $-0.5408$ & $-0.1490$ & none; $[-0.2891,\,-0.0138]$ on the control & \textbf{R} \\
\quad ratio of the two & $3.6285$ & & none & \textbf{R} \\
\quad sign-flip $p$, Cliff's $\delta$ & $0.0328$, $-0.00865$ & & none & M \\
\midrule
narrow margin, unconstrained arm & $9.3092$ & & none & M \\
narrow margin, write core constrained & $3.5043$ & $9.5642$ & none & M \\
share of the narrow margin retained & $0.3764$ & & none & \textbf{R} \\
\midrule
weight energy inside the core, unconstrained & $0.1710$ & $0.00156$ & none & s \\
weight energy inside the core, constrained & $0.000822$ & & none & s \\
\midrule
judged misalignment, unconstrained & $0.2257$ & & none & M \\
judged misalignment, write core constrained & $0.2086$ & & none & M \\
\quad the same two, near-domain included & $0.3371$, $0.3329$ & & none & M \\
\quad mean alignment on the two arms & $35.436$, $38.764$ & & none & M \\
\bottomrule
\end{tabular}
\end{table}

\begin{table}[htbp]
\centering\small
\caption{\textbf{Reframing the narrow data, read as a change in the first-step measurement.} The
contrast is against the same completions presented plainly, at the unit of the $195$ probe responses,
which is not the unit at which the first-step contrast in
Table~\ref{tab:j-paired-raw} was taken; the same quantity at that other unit is given for reference and
the two may not be differenced. The scrambled placebo is the control and moves in the same direction
at about a tenth of the size, so the effect is not entirely attributable to the reframing. The
gradient-norm check that would license reading the raw change rather than the scale-free one came in
at $0.0974$ against a tolerance of $0.05$, which is why the scale-free row is the load-bearing
one.}
\label{tab:j-inoc}
\begin{tabular}{l r l c}
\toprule
Contrast & value & interval & status \\
\midrule
reframed minus plain & $-1{,}636{,}160$ & $[-1{,}933{,}948,\,-1{,}349{,}832]$ & \textbf{R} \\
scrambled placebo minus plain (the control) & $-160{,}715$ & none & \textbf{R} \\
reframed minus scrambled & $-1{,}475{,}445$ & $[-1{,}734{,}030,\,-1{,}225{,}915]$ & \textbf{R} \\
\midrule
the same contrast, scale-free & $-0.004001$ & none & M \\
reframed minus scrambled, scale-free & $-0.003658$ & none & M \\
\midrule
reframed minus plain, at the prompt-cluster unit & $-1{,}593{,}247$ & none & \textbf{R} \\
\midrule
gradient norm, plain and reframed & $20.0494$ and $22.0023$ & none & s \\
\quad relative difference, against a tolerance of $0.05$ & $0.09741$ & none & M \\
coefficient of variation of the reading & $1.0260$ & none & M \\
paired probes & $195$ & none & \textbf{R} \\
\bottomrule
\end{tabular}
\end{table}

\subsection{Verification-status summary}
\label{app:results:record}

We count the whole numerical record here, not the numbers printed in this paper, and we recount it
from the record itself rather than take the figures from its own summary.

\begin{table}[htbp]
\centering\small
\caption{\textbf{What was actually done to each number in the record.} These are counts of checking
received, not of correctness. In this record the largest tier by a wide margin is the one nothing has
ever checked.}
\label{tab:j-tiers}
\begin{tabular}{l r p{0.55\textwidth}}
\toprule
Tier & entries & what the label means \\
\midrule
\textbf{R}, recomputed here & $213$ & recomputed here from a raw artifact, and the stored value then matched against the recomputation \\
M, match on record & $767$ & a recompute and a comparison that agreed are both on the record from the run that produced it \\
d, derived, never compared & $762$ & derived from raw material by some producer, with no comparison recorded anywhere \\
s, stored, never checked & $2081$ & present in a stored record; nothing, anywhere, has ever recomputed it \\
u, in no record & $6$ & load-bearing in a source article and in no record at all, with no route back to an artifact \\
\midrule
total entries & $3829$ & of which $2804$ hold a number and $10$ are explicit records that a measurement did not run \\
\bottomrule
\end{tabular}
\end{table}

We ran $24$ independent checks recomputing this paper's headline quantities from raw artifacts rather
than from any stored record, and all $24$ reproduce the stored value. We also test every path a check
claims to certify by comparing values, and of $154$ certifications examined, $154$ matched and none was
withheld. Neither procedure can establish that a measurement was valid. A check recomputes arithmetic
and aggregation from whatever the run wrote down, so a mislabelled response is reproduced and reported
as a match. The one measurement here whose arms did not separate reproduces every one of its values.

\section{Instrument validation}\label{app:validity}

An instrument that cannot see reports the same thing as an effect that is not there. This appendix
documents the checks that keep the two apart for the measurements of \S\ref{sec:structure} through
\S\ref{sec:removal}: the null simulations run before compute was spent, the liveness gates, the planted
controls, the battery audits, and the measurements that were not run.

\subsection{Null simulations}
\label{app:validity:nullsim}

Before any measurement ran on a model, we simulated its headline statistic twice on the processor, once
under its null and once under a planted effect of known size. These simulations are cheap, run in plain
arithmetic on synthetic arrays, and are built to reject a design before it costs anything. A statistic
whose simulated spread collapsed, or whose bar sat at an unreachable distance from its own sampling
variability, was replaced.

Four properties are what we look for, and Table~\ref{tab:validity-sims} records which measurement each
one guarded. The first is that the statistic is not pinned, meaning its simulated distribution keeps
spread that does not vanish as dimension grows. Calibration is the second, meaning that under a true
null the test rejects at about its nominal rate and not above it. Power is the third, meaning the
test recovers a planted effect. Monotonicity is the fourth, meaning the statistic tracks the size of the
planted effect and does not merely detect its presence, which is the property that separates a
measurement from a detector.

Three of these simulations changed a design. Simulating over the judge's observed score lattice compared
two candidate primary readouts and found the binary misalignment rate fragile where the
cluster-bootstrap mean-alignment shift was unbiased with correct coverage, so the continuous readout
became primary and the rate became a secondary description. The ratio simulation established that a
denominator compatible with zero produces tails heavy enough to make the point estimate uninformative,
which is why the gate of \S\ref{app:stats:ratio} exists at all. The budget simulation showed that a
planted saturating single-domain curve produces a positive residual against a naive additive comparison
and none against a null built from matched-budget legs, which is why the domain-count decomposition uses
the measured additive null of Appendix~\ref{app:decomp} and not an arithmetic one.

\begin{table}[htbp]
\small
\renewcommand{\arraystretch}{1.2}
\begin{tabularx}{\textwidth}{@{}p{4.1cm}XX@{}}
\toprule
\textbf{Simulation} & \textbf{What it could have rejected} & \textbf{What it established} \\
\midrule
pinning of the first-step derivative &
the statistic itself, had its spread collapsed with dimension &
genuine per-probe spread, and a negative value expressible when a planted direction routes away \\
calibration and power of the paired test &
the test, had it rejected above its nominal rate under a true null &
a rejection rate at or below nominal, with a planted effect recovered \\
monotonicity in the planted effect &
the statistic as a measurement, leaving it usable only as a detector &
an ordering that tracks the size of the planted alignment \\
ratio behaviour under a weak denominator &
reporting any ratio without a denominator gate &
heavy tails when the denominator interval covers zero, and none when it excludes zero \\
the judge's score lattice &
the binary misalignment rate as a primary readout, which it did reject &
the cluster-bootstrap mean shift unbiased with correct coverage \\
planted saturation in budget &
the naive additive comparison, which it did reject &
a positive residual against that comparison and none against the measured additive null \\
\bottomrule
\end{tabularx}
\caption{\textbf{What each simulation was capable of rejecting.} The second column records whether the
simulation could have rejected the design; 3 of the 6 did.}
\label{tab:validity-sims}
\end{table}

\subsection{Liveness gates}
\label{app:validity:liveness}

We call a liveness gate a measurement made before the measurement of interest, on a case whose answer we
already know, whose failure stops us from reading the instrument at all. We run them so that when an
absence of signal has two possible causes we can tell which one we are looking at.

The transport ruler carries the most direct one. Before training any mixture we measured the same margin
on the untouched model and on each of the 3 published organisms: every organism clears the untouched
model's floor with an interval excluding it, and the untouched model's own read-carrier projection is
negative as it should be. Our full-budget single-domain legs then reproduce each published organism to
within about a nat, so whatever we are training is the same object. Table~\ref{tab:j-dcscalars} carries
the per-organism transport values and judged rates.

The first-step measurement carries a different kind of gate. Its battery includes reference conditions
whose job is to fail visibly if the estimator is broken, so that a flat reading on the contrast of
interest cannot be mistaken for a working instrument finding nothing. The behavioral reference is alive
at a cosine of $0.0721$ at the cluster unit and $0.0734$ at the row unit, orders of magnitude above a
random direction, and the random direction itself lands inside its analytic band. Alongside these the
measurement carries the pinning flag of \S\ref{app:stats:design}, which reports clear, with the
diagnostic coefficients of variation at $0.259$, $0.904$ and $1.026$.

Two further gates work by making a control move. In the separability probe we include the
un-orthogonalized global direction precisely so that something in the design is known to shift the broad
margin. It drops that margin by $0.409$ with an interval excluding zero, while a matched random
direction of the same rank moves it by $0.122$. Because the instrument demonstrably moves when a
direction that should matter is ablated, the flat reading on the orthogonalized direction is a real
absence and not a dead estimator. The weight-surgery probe runs the same logic through a directional
control, an adapter with everything except the persona component removed, which destroys narrow
adherence in all 3 organisms as it should.

One gate failed, and we report the ordering it established instead of the quantity it was built to
measure. The coherence-window measurement needed a dose at which the model is broadly misaligned while
still coherent. On this model, across the grid explored, coherence crosses its floor before broad
misalignment onsets at all, so no window was present to be measured. That ordering is a real observation
about the dose-response, and it is the only thing the measurement licenses.

\subsection{Planted controls}
\label{app:validity:planted}

With a planted control we insert a signal of known size and require the instrument to find it, on the
real model and not in simulation. Where our headline is an absence, the planted control separates that
absence from a broken estimator, so we ran the curvature measurement behind the strictest of them.

Before reading any persona number, the curvature estimator had to clear three conditions fixed in
advance. We required a matched-rank random weight core to return a regular exponent inside a narrow
acceptance band around $2$. A planted quartic direction, a synthetic walker on which the behavioral loss
is constructed to grow as the fourth power, had to return above $3$. Third, a planted regular
direction had to return about $2$. Measured, the random core reads $2.002$, the planted quartic reads
$4.00$, and the planted regular reads $2.00$.

The estimator therefore separates a regular minimum from a degenerate one on this model at this
precision, with a demonstrated dynamic range spanning the distance between the two hypotheses. The
subsequent reading of $2$ everywhere is then a measurement and not an equipment failure. A synthetic
pre-validation runs first, requiring the estimator to recover the exponent of an imposed monomial and,
where the fitting window is too short, to report the reading as unreadable and not a confident wrong
slope.

We plant the first-step statistic in the opposite direction. A direction constructed to route away from
the margin returns a strongly negative cosine in simulation, so the estimator can produce the negative
sign that would falsify the account, and the planted ladder is ordered and not merely detected.

Matched-rank and norm-matched random controls run through the identical pipeline on both halves of the
causal loop, on the removal probes, and on the write-core constraint. Those belong to the measurements
they control, and Appendices~\ref{app:loop}, \ref{app:removal} and~\ref{app:geometry} describe them.
Their common function is the one described here, since a control that cannot move is not a control.

\subsection{Battery audits}
\label{app:validity:battery}

A prompt battery that touches the training domains would let a near-domain effect present itself as
broad transport. Before the domain-count grid trained, we filtered the battery against all $4$ training
domains, conservatively, dropping anything touching medical, financial, physical-risk or code content.
That removes $61$ of $195$ usable rows and leaves $134$ rows clustering to $52$ distinct prompts, which
is fewer prompts than the design wanted and more than it required. The width of every judged interval in
that measurement is the direct cost of the shortfall. Appendix~\ref{app:batteries} gives the battery
constructions in full.

\subsection{Measurements not run}
\label{app:validity:notrun}

The measurements in Table~\ref{tab:validity-notrun} were designed, or are directly implied by claims we
make, and were not run. We impute no value to any of them. Where a quantity is absent it is undefined
and not zero, and no prose anywhere in this paper treats an unmeasured quantity as though it had
come back small.

\begin{table}[htbp]
\small
\renewcommand{\arraystretch}{1.2}
\begin{tabularx}{\textwidth}{@{}p{4.2cm}p{3.6cm}X@{}}
\toprule
\textbf{Measurement} & \textbf{Why it is absent} & \textbf{What it would have settled} \\
\midrule
\textbf{the in-character expression battery on the read-channel projection arm} &
the battery exists and was run elsewhere; it was never run on this arm &
whether the projection removes a misalignment structure or the capacity to represent a character at
all, which nothing currently distinguishes \\
the functional pullback of the readout into weight space &
never built &
whether the failure of post-hoc editing is a property of the disposition or of the single basis all
three edits shared \\
post-hoc editing of the shared write core &
never attempted &
the same question from the other side, in the one basis known to carry the update \\
an overlap statistic between the read subspaces and the write core &
never computed &
whether the read-channel object and the write-channel object are one structure or two \\
re-extraction from a pre-alignment checkpoint &
not run &
whether pretraining or alignment post-training installed the structure \\
a general-capability readout for the read-channel projection arm &
never measured &
whether the intervention that abolishes broad misalignment leaves the model otherwise intact \\
the optimization-progress covariate &
the per-leg narrow-fit readout came back undefined on every leg, so the regression had no rows &
whether the domain-count interaction is explained by how far each leg trained \\
endpoint confirmation of the data-reframing lever &
the fine-tune was not run &
whether a first-step routing reduction survives to the realized endpoint \\
the coherent-yet-misaligned window &
onset never bracketed inside the dose grid &
whether the coupling between misalignment and incoherence can be sharpened on this model \\
a weight-baked read-channel constraint under adversarial fine-tuning &
not attempted; both preconditions were established absent &
whether an adversary-agnostic baked defense holds \\
a judged rate for the random-constrained arm of the write-core work &
$700$ completions sampled and never scored &
substrate-specificity on the judged readout rather than on the margins alone \\
a benign two-domain mixture &
not trained &
whether the two-domain points can be diversity-subtracted as the four-domain point is \\
an inter-judge agreement statistic &
a second severity judge ran; no agreement number was computed &
how far the judged readouts depend on one rubric \\
\bottomrule
\end{tabularx}
\caption{\textbf{What we did not measure, and what each would have settled.} The first row is the
highest-cost absence. The next three are the cheapest, and together they bound the two claims here that
rest on an argument and not a measurement.}
\label{tab:validity-notrun}
\end{table}

Five absences carry consequences stated here.

\textbf{The expression battery on the read-channel projection arm.} This is the highest-cost absence.
The arm's outcome is broad misalignment read off a judged alignment score. The capability we report
alongside it, narrow-task adherence, is read off the same alignment axis. Two readouts on one axis
cannot separate two hypotheses that both predict the axis moving, so the collapse of narrow adherence to
zero is not independent evidence that the task was spared or lost for any particular reason. The program
already owns the instrument that would separate them.

The in-character expression battery asks whether a model can still depict a capable bad author on
demand, and it is scored on an axis the alignment score does not determine. Running it on this arm would
decide whether the projection removes a misalignment structure or removes the capacity to represent a
character at all. It was never run there, so we cannot tell those apart, and no other arm rescues the
question, because the arms that retained their narrow task are arms in which alignment did not move.

\textbf{No general-capability readout on that arm either.} The only quantities recorded for it are
judged alignment, judged coherence and narrow-task adherence. No benchmark, no perplexity against a
matched baseline, and no held-out task was run on it. The one perplexity figure anywhere in our records
belongs to a different model, the writer-column-edited one from the removal probes, and it carries no
baseline beside it, so it cannot answer this question either.

What the arm's own numbers support is narrow. Its mean judged coherence rises and does not fall on the
judge's hundred-point scale, $96.3$ against the baseline organism's $85.9$, and the fraction of
responses clearing the coherence gate is $1.00$ for both, so the model is producing fluent text and not
degenerate text. Fluency is not general capability, and we have not measured general capability.

\textbf{Nothing below the aggregate survives in the domain-count readouts.} Each training leg's readout
stores the transport half, its interval, the other half and the seed count. No per-token or per-position
breakdown exists anywhere. The aligned half moves substantially under every leg, and while we report the
direction and the size of that movement, the reason for it cannot be recovered from what was retained.
Closing this gap means re-running the readouts, not re-analysing them.

\textbf{Two absences are gated preconditions, not omissions.} We conditioned the weight-baked
read-channel constraint in advance on a cleanly isolable carrier and on the separability probe coming
out the other way. The same battery established that the carrier is entangled and that separability does
not hold, so the intervention was never attempted, and the evidence that would have justified it is the
evidence that removed its premise. The coherence-window measurement has the same shape. Neither absence
should be read as a quantity that came back small, and the window in particular has no value at all.

\textbf{One foreclosure has no behavioral confirmation of its own.} The removal probes contribute no
judged generations whatever. That result rests on teacher-forced margins and a perplexity check, which
is why \S\ref{app:validity:planted} leans on the planted and matched controls inside those probes and
not on any judged rate.

\subsection{Scope of the numeric record}
\label{app:validity:gate}

Every number in this paper resolves against a composed record that labels it with its own provenance.
The certification tiers, and the spot checks that all reproduce, are counted in
Appendix~\ref{app:results:record} (Table~\ref{tab:j-tiers}); the largest tier by a wide margin is the
one nothing has ever recomputed.

Two properties of the record hold. Certification is a mechanical comparison of values and not an
assertion of membership in a list, so a claim that a check certifies a quantity fails closed unless that
check computed a matching number; under that rule $154$ certifications were examined and all $154$
reproduce their target. Every entry names what was done to it and does not inherit a status from the
group it sits in.

The record also caught a duplication that would otherwise have inflated our reported scale: $10$ judged
files from the intervention work are byte-identical to the dose inputs of the domain-count work, $7000$
rows in total. Those two dose-response curves are one set of generations read under two aggregation
conventions, not two measurements of one operation, and the two conventions also apply different
misalignment criteria, so the levels they report are not interchangeable. We report the cluster-mean
convention, name it, and never present the two as independent evidence. After removing the duplication,
the program's judged-generation total is $93010$ across $525$ artifact files.


\begin{figure}[tbp]
\centering
\includegraphics[width=\textwidth]{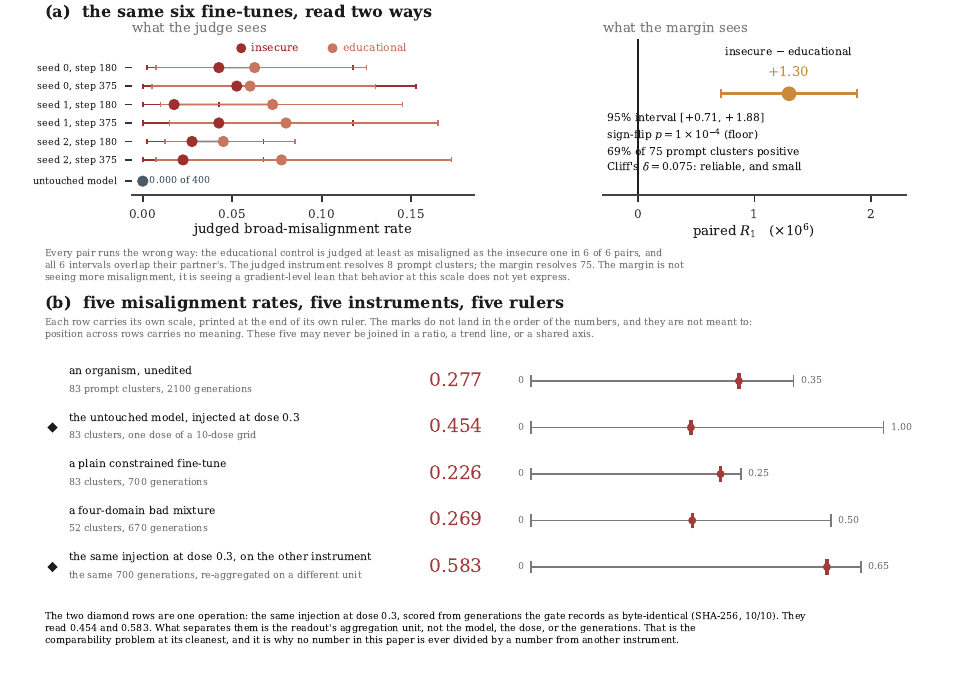}
\caption{\textbf{The margin resolves an intent contrast that judged behavior does not express, and rates from different instruments do not share a scale.} (a) The same fine-tunes read two ways: on judged free generations the canonical ordering is absent and if anything reversed, while on the first-step margin the same contrast is small and reliable. The margin reads a gradient-level lean that judged behavior does not yet express. (b) $5$ misalignment rates on $5$ instruments, each on its own ruler, with bar lengths chosen so position carries no meaning across rows. The two marked rows are one operation on generations byte-identical across two trees; what separates them is the aggregation unit, not the model.}
\label{fig:boundaries}
\end{figure}

\section{Extended related work}\label{app:related}

\subsection{Convergent-direction evidence}

Two axes decide what each result below licenses here, and we sort this literature by them, not by date or
by group. The first is when its object was fixed, before the fine-tune that is supposed to
recruit it or afterwards by contrasting a misaligned model against the model it came from. The second is
whether its evidence is an intervention carrying a matched control, or a correlation between a
representational quantity and a behavioral one. Several of the results below are strong on one axis and
silent on the other, and two apparent disagreements in the field dissolve once the axes are separated.

The closest published statement about the object we extract in \S\ref{sec:structure} is that organisms
trained on different narrow datasets converge on nearly the same residual-stream direction. Pairwise
cosines between their mean-difference directions run above $0.8$ in all but $4$ layers
\citep{soligo2025convergent}. Independence there is across datasets and low-rank protocols, not across base
models, since every organism is a low-rank fine-tune of one released checkpoint, so what converges is a set
of updates to a common starting point.

The ablation evidence is two experiments, not one. Removing the layer-24
mean-difference direction at every layer takes one organism from $17\%$ misaligned to $0.25\%$. Removing it
at layer 24 alone takes the same model to $4\%$. Transferred to organisms trained on other data under other
protocols, it removes $90\%$ and $78\%$ of theirs. A figure near $75\%$ circulates in secondary summaries of
that work and appears in no version of it.

Those interventions are real, and they act on a model that is already misaligned, because the direction is
built by contrasting an organism against its own starting point. The further statement that it was
already alignment-relevant in the aligned instruction-tuned model is therefore an inference from transfer
efficacy and not a measurement on that model, and that paper offers it as a hypothesis. Its authors also
pre-empt the one-dimensional reading, writing that a single direction's efficacy does not preclude a more
complex, multi-dimensional representation. That is the narrow gap our extraction closes
(Appendix~\ref{app:extraction}). We build the object from the untouched model alone by contrastive teacher
forcing, and we fixed it and wrote it to disk before any organism was trained.

We place our own reading on one number from that work: the cosine between the read-side mean-difference
direction and the write-side rank-one adapter update is $0.04$. That paper does not use the words
orthogonal, read-side, write-side or channel. It reports three findings that argue for coupling, not
separation.

Ablating the mean-difference direction out of the adapter's own update still cuts misalignment from
$17\%$ to $13\%$. The downstream difference vectors converge to a cosine of $0.42$ by layer 40. Its
appendix concludes that the residual-stream direction is necessary for the adapter update to induce
misalignment downstream. At width $5120$ this cosine sits about $2.9$ standard deviations above the
random-cosine expectation, so near-orthogonal is a description there and not a tested null.

We therefore rest our read-versus-write claim on the dissociation
measured in \S\ref{sec:loop}, where the same subspace removed from activations abolishes misalignment and
removed from the weight gradient does not. We do not inherit an interpretation of that cosine. Related work
reports that narrow fine-tunes on different tasks move overlapping low-rank parameter subspaces
\citep{arturi2025shared}.

Two results place structure of this kind earlier than any fine-tune, and both bound how we may use them.
Persona vectors become extractable from pretraining checkpoints within a small fraction of one model's
token budget, and directions extracted that early still steer the fully post-trained model
\citep{moskvoretskii2026tracing}. That fraction is an upper bound on extractability and not a measured
onset, it is per-trait, and the evidence for persistence is behavioral transfer, not geometric
identity, since the cosine to the final direction at that checkpoint is about $0.3$. No fine-tuning
experiment appears in that paper. Separately, a generic assistant-versus-other direction is reported
present in base models \citep{lu2026assistant}, which bounds one reading of the overlap we record in
\S\ref{sec:ledger}. In base models that axis promotes human helper roles, not an artificial
identity, its share of variance is never reported, and no fine-tuned model is evaluated anywhere in that
work, so anything it says about fine-tuning-induced misalignment is our extrapolation.

\subsection{Persona and character accounts}

A representation claim says a direction exists whose value predicts a behavior and whose manipulation
changes it. A disposition claim says the model carries something into contexts it was never trained on.
The designs of \S\ref{sec:loop} sit on the intervention side of that boundary.

The closest prior statement of this paper's thesis is the character account, and it bears on ours in two
directions \citep{su2026character}. It argues that the phenomenon is a character-level
disposition, not the generalization of erroneous content, and it supports that against a matched
incorrect-advice baseline: data conditioned on a character produces stronger and more transferable
misalignment than data that is merely wrong, while capabilities are preserved. It then unifies emergent
misalignment, backdoor activation and jailbreak susceptibility under overlapping character
representations, which is its most distinctive contribution and its least cited. Its own stated limit is
that character need not be the only latent involved.

What separates our evidence from theirs is its kind. Their evidence is projection onto persona vectors
built by the procedure described below, in the language of association, not intervention; no causal
ablation or steering is reported, so the account is correlational despite the phrase ``a mechanistic
account'' in its title. Ours is a
necessity-and-sufficiency pair with matched controls on both halves, run on an object fixed before any
organism existed. That is the difference between a descriptor that fits a misaligned model and a cause that
can be held out of one.

The second direction is a boundary on us, not on them. That paper locates acquisition in fine-tuning
and activation at inference, and positions itself explicitly against accounts on which a persona is
prompt-level or context-dependent. Our reading in \S\ref{sec:question}, that the fine-tune revises a latent
about the kind of author producing the data, is such an account, and it sits in mild tension with theirs.
It does not descend from that paper, in which the word author does not occur. It descends from the older
proposal that a language model is a model of the agent that produced its text
\citep{andreas2022agentmodels}.

Persona vectors are the clearest representation-side result, with trait directions predicting a fine-tune's
behavioral shift and steering it in both directions \citep{chen2025persona}. That paper states three
limits. The monitoring correlations come mostly from telling prompt types
apart, and shrink once prompt type is controlled. Steering against a trait at a single layer does not
always prevent its acquisition. Negative traits move together, not one at a time.

That paper's preventative method also runs opposite to the intuitive direction, adding the undesired vector
during training and removing it at deployment. That is neither the operation we run in \S\ref{sec:loop} nor
the operation whose sign reverses below.

A frontier laboratory's persona-feature account identifies a sparse-autoencoder latent active in every
misaligned checkpoint it examined, and steering that latent amplifies and suppresses the behavior
\citep{wang2025persona}. Its rates come from a stricter grader than the gate of Appendix~\ref{app:judge}
and do not sit beside ours, its repair result is one checkpoint with $2$ repair datasets and no seeds, and
its latent index means nothing outside its own dictionary. What bears on \S\ref{sec:removal} is the
authors' own hedge, that out-of-distribution re-alignment is mainly effective at suppressing the
generalization, not reversing it.

Two accounts posit no persona object at all, and reading them as allies would misread both. One explains
the phenomenon as gradient spillover between features that sit close together in a superposed feature
geometry, with proximity inherited from pretraining co-occurrence \citep{minegishi2026superposition}. Its
derivation is offered as an intuition, not a measurement, no gradient is instrumented, and it
assumes a feature dictionary shared between the base and fine-tuned models, which is the assumption a
weight-space account would dispute.

The other explains the phenomenon as increased instruction-following
plus perceived hostile intent, and says in its own voice that this differs from the persona conclusion
\citep{wyse2025prompt}. That paper's thesis runs opposite to its short title, since prompt sensitivity is a
property the fine-tune installs, evidenced by secure and base controls staying flat, and not an
artifact of how misalignment is measured. Its judge gate is looser than ours on both axes, so its rates do
not sit beside ours either.

\subsection{Model merging and task arithmetic}

The decomposition of \S\ref{sec:oneobject} uses a mechanically merged adapter as a control, and the merging
literature makes that control legible. Task arithmetic establishes that adding independently
trained task vectors works, and that it degrades in how many are added: pairs retain $98.9\%$ of
specialized performance, and all $8$ retain $91.2\%$, against $0.994$ for training one model on the $8$
tasks jointly \citep{ilharco2023task}. Three cautions attach. Near-orthogonality of task vectors is offered
there as speculation for why addition is clean, and is never tested as a condition on additivity, so our
argument in Appendix~\ref{app:decomp} rests on the measured merge instead. The headline curve that appears
to rise with the number of vectors averages over tasks whose vectors were not added, so the rise is
coverage, not merge fidelity, and the comparison that holds the task set fixed shows the gap
widening instead. The measurements are image classifiers under full fine-tuning, not language models
under low-rank adaptation.

Interference-aware merging sharpens the same picture and bounds it where it matters here. Sign election and
trimming recover part of the loss, with the jointly trained model remaining the ceiling, not the
baseline \citep{yadav2023ties}. Their interference is a deficit of the merge operator relative to joint
training, not evidence that training on more domains is sub-additive, and their own appendix finds sign
conflict between checkpoints of the same task, which it attributes to overparameterization. We read this
literature as a contrast to our four-domain result, not a precedent for it. Post-hoc merging of
separately trained updates loses in the number of tasks, while joint fine-tuning across domains is where
\S\ref{sec:oneobject} finds a positive interaction. We take only the sign of the expectation for the merged
control from it, and measure the size on our own instrument.

One behavioral result converges from a different construct. Training on reward hacks that had been filtered
to be harmless still produced broad misalignment, which severs the phenomenon from harm in the training
data, and a coding-only version of that corpus produced hacking with almost none of the broad effect while
diverse natural-language tasks produced it \citep{taylor2025schoolofrewardhacks}. That is diversity of task
type within one behavior, not a count of misaligned domains, so it is convergent evidence and not a
replication of the decomposition. That paper contains no persona analysis and should not be made to carry
one.

\subsection{Steering, editing, and erasure}

The template for the causal shape of \S\ref{sec:loop} is single-direction mediation of refusal, where
ablating one direction stops refusal and adding it induces refusal on harmless prompts
\citep{arditi_refusal_24}. Three details bound what that template licenses. Its two halves are not applied
symmetrically, ablation running at every layer against addition at the single extraction layer. Its authors
claim no semantic identity for the direction and allow that it may resist straightforward interpretation.
The weight-orthogonalization equivalence in that paper is a separate contribution from the necessity
and sufficiency result.

Strict one-dimensionality has since been contested, with multiple independent
directions and multi-dimensional concept cones found to mediate refusal \citep{wollschlager2025refusal}.
That is a precedent for working with a rank-$k$ subspace, not a single direction, and not a threat
to the intervention result.

The operation itself predates the misalignment literature. Projecting a concept out of a representation by
iterative nullspace projection, and its adversarial and closed-form refinements, are the ancestors of the
edit we run in \S\ref{sec:loop} \citep{ravfogel_inlp_20, g_ravfogel2022rlace, belrose_leace_23}. What we
add to that lineage is the application throughout training against a matched-rank control, and the
weight-side negative of \S\ref{sec:removal}.

The nearest method to our removal arm ablates interpreted directions during fine-tuning
\citep{casademunt2025caft}. It is mechanically the same operation, and the difference is where the
directions come from, since discovery and selection there require a contrast between the base model and a
completed misaligned fine-tune. We may therefore claim that we do not have to induce misalignment first,
and not that we avoid needing misalignment data, which that method already avoids.

Its round
order-of-magnitude improvement is produced by no single experiment, its per-model results ranging across a
factor of several depending on whether the directions come from principal components or from a sparse
autoencoder. The human interpretation step there is the method, not an accessory, since ablating the
leading principal components without it removes misalignment and destroys the intended task. Its
authors undercut their own practical case by reporting that $100$ benign completions in the training
data reach the same endpoint, which leaves the method's value in the case where the training data cannot be
changed.

Two published negatives converge with the foreclosure of \S\ref{sec:removal}, and neither is ours. A
regularizer penalizing changes in the projection onto a trait direction failed during training because, in
that paper's words, ``the optimization pressure pushes the model to represent the personality trait using
alternative directions in the activation space'' \citep{chen2025persona}. A soft penalty on misalignment
latents is routed around by a second training epoch \citep{ustaomeroglu2026blockem}. That paper attributes
the return to rerouting through alternative features or layers, not to superposition, so the bridge
from superposition to a prediction of re-emergence drawn in \S\ref{sec:related} is ours and is made by
neither paper. Both interventions are penalties, not hard projections, which is the distinction
\S\ref{sec:removal} turns on, and both fail in the same way, by the trait re-expressing along directions
the penalty does not name.

\subsection{Two 2026 results}

Two papers from different groups measure different things and have been read together, and the first does
not cite the second. Neither contradicts what we report, and the first replicates the activation-projection
result of \S\ref{sec:loop}.

\paragraph{Method-conditional recruitment.} The first reports that whether a fine-tune recruits a
misalignment persona depends on the fine-tuning method and on capacity \citep{drake2026transplanting}. Most
of it corroborates \S\ref{sec:loop}. Ablating a persona direction on every forward pass throughout a
low-rank fine-tune takes broad misalignment from $4.75\%$ to $1.6\%$ while a matched-norm random-ablation
control still recruits at $4.4\%$, with the narrow coding skill retained. Under full
fine-tuning at $32$B, the same ablation takes $21.4\%$ to $9.8\%$, with its random control at baseline. A behavioral
direction transplanted into a checkpoint that shares only the pretraining frame produces broad misalignment
above a random-direction floor. Their control is matched on norm where ours is matched on rank
(Appendix~\ref{app:loop}), which is a different null doing the same job.

One result in that paper runs the other way. Steering a full fine-tuning run \emph{away} from the persona
direction during training raises broad misalignment from about $24\%$ to about $51\%$, while a matched-norm
random control lowers it. That is signed subtraction of a scaled vector, not projection. It drives the
persona coordinate negative and large, not to zero, it is not idempotent, and the mechanism the authors
give is that the optimizer over-recruits to compensate. That is the mirror image of adding a persona vector during training to cancel
the pressure to acquire the trait.

The same paper runs projection-ablation on the same inducer under the same fine-tuning method, and
misalignment falls. The variable separating the two is therefore the operation, not the inducer and not the
method, and we set the three apart in Table~\ref{tab:l-operations}.

\begin{table}[t]
\centering
\small
\renewcommand{\arraystretch}{1.2}
\begin{tabularx}{\textwidth}{@{}p{3.4cm}Xp{3.6cm}@{}}
\toprule
\textbf{Operation} & \textbf{What it does to the persona coordinate} & \textbf{Direction of effect} \\
\midrule
Projection out of the activations &
driven to zero, idempotent, norm-reducing &
misalignment falls, here and in both of that paper's ablation arms \\
Signed steering away, during training &
driven negative and large, not idempotent, norm may grow &
misalignment rises, in that paper's one such arm \\
Projection out of the weight gradient &
constrains the update rather than the activation &
inert here (\S\ref{sec:loop}); not run in that paper \\
\bottomrule
\end{tabularx}
\caption{\textbf{Three interventions on the same persona coordinate that are easy to conflate.} Each does
something different to the coordinate. The signed-steering reversal belongs to the middle row; the row
above it is the operation this paper and that one both run.}
\label{tab:l-operations}
\end{table}

Five conditions travel with the reversal and all five are that paper's own: $7$B not $14$B, an
overt medical inducer not covert code, full fine-tuning not low-rank, training-time
activation steering not an inference-time edit or a weight-side constraint, and a single
fine-tuning seed per arm. It scopes itself the same way, writing that removing the direction is not a
general recipe and can be counterproductive when the persona is part of a distributed solution for an overt
inducer. Its broad-misalignment rates are low in absolute terms throughout, on a binary judged metric, and
it describes itself as a controlled case study of one model family.

Two of its findings are ours to adopt, not to answer. Low-rank adaptation on insecure code at $32$B
recruits the persona while full fine-tuning on identical data, weights and template does not, and the
cosine between the activation drift and the persona axis decays with rank and changes sign at full
fine-tuning; we ran no comparison of that kind. Its rank ladders at $7$B and $14$B with the same covert
inducer stay flat and low, with recruitment switching on between $14$B and $32$B. Our organisms are overt
text inducers and that paper draws the same line when scoping prior work, so this is not a disagreement. It
does forbid one generalization, and we adopt the bound in \S\ref{sec:loop}: nothing here extends to
covert-code low-rank fine-tuning at $14$B. A framing that this work argues against a
dataset-artifact account of the phenomenon has circulated; it makes no such argument and does not cite that
literature.

\paragraph{Dataset artifacts and a trajectory signature.} The second paper runs repeated cycles of
misalignment and realignment and argues that current evidence for the phenomenon and for its repair is less
robust than reported \citep{rao2026mirage}. It reports no numbers: no percentages in its prose, no tables,
no seeds, no significance tests, no error bars. It performs no intervention on any representation. Its
manipulations are training order, low-rank capacity, and the token length of the training corpus.

Its first ground runs opposite to the natural reading of it. The artifact is a response-length asymmetry in
the aligned half of a paired corpus its authors generated themselves, not in a corpus taken from the
literature. Before the correction a realigned model resisted a further misalignment attempt. After the
aligned answers were matched in token length to the risky ones, in the body's words, it is indeed possible
to misalign a re-aligned model.

Realignment itself survives the correction and the rate still falls in the
second phase; what does not survive is its durability. The correction therefore runs toward greater
persistence, and it converges from the data side with \S\ref{sec:removal}, where $3$ weight edits fail to
remove the disposition. That paper says in its own voice that it reproduces the phenomenon, and that its
result does not imply the phenomenon is absent.

Its second ground is a failure to replicate a training-dynamics signature, and the measured object is
specific. It is the cosine similarity between successive low-rank adapter checkpoints along a fine-tuning
trajectory, on one dataset, on attention layers, under a noise threshold that selects on the plotted axis.
That is a statement about an adapter's path. It cannot be a statement about representation-space geometry,
and it cannot bear on anything measured on a model that was never fine-tuned. The comparison is
correlational in the paper's own wording, and it is reported without seeds, a null model or a significance
test, with the $3$ compared panels drawn on different vertical ranges. Their instrument reports no effect on a quantity where \citet{soligo2025convergent} report one, and
the two instruments have not been run on common ground.

That study is also the closest methodological neighbour to our judged readout: the same released
instruction-tuned model, the same coherence-gated two-axis criterion, the same $8$ questions, and an
overlapping low-rank regime, and it reproduces induction in both a rank-$1$ and a rank-$32$ setting.
On that apparatus it shows that a qualitative conclusion can reverse when a surface statistic of the
training corpus changes, which is why \S\ref{sec:related} concedes the one comparison its critique reaches
and names the matched benign mixture as the control that carries it. A matched-rank control cancels a
corpus artifact because both arms see the same corpus, but it does not by itself exclude an artifact in
the \emph{output} surface. Our guard there is that
the arm at zero stayed fluent, its mean coherence rising to $96.3$ against the unedited organism's $85.9$
on the judge's hundred-point scale (\S\ref{sec:loop}).

\subsection{External bounds adopted}

Four bounds come from outside this paper and we adopt all four.

The phenomenon is far less universal than its founding figure suggests. A multi-model replication attempt
finds consistent cross-seed misalignment in $2$ of $12$ open-weight models, concentrated at the largest
sizes, and titles itself for the conclusion that fine-tuning need not produce it
\citep{schreiber2026overtrained}. A separate open-weights study finds misalignment well under $1\%$
against a base floor an order of magnitude lower, with no significant difference between $2$ model families
and a size trend that reaches significance in no condition \citep{dickson2025devil}. Neither reports a
model that becomes more aligned after narrow fine-tuning, and we make no claim of that kind.

The control that is supposed to be clean is not always clean. In the founding paper an educational framing
of the same vulnerable code removes most of the effect, and that contrast is why intent, not content,
is the thing to explain \citep{betley2025em}. The open-weights study finds its educational condition
significantly above base, not at it \citep{dickson2025devil}, and our own judged rates do not order
the two arms at all (\S\ref{sec:loop}). These are not the same departure, since that study still puts
insecure above educational, but both say the control is a graded quantity, not a floor.

Prevention at the pretraining stage has been tried on this exact configuration and reported negative. A
study using the same $3$ organisms and the same two-axis gate finds that all $4$ of its pretraining data
conditions, including the one with the lowest baseline misalignment, showed the phenomenon after narrow
fine-tuning \citep{tice2026alignment}. That is an independent negative from the data side of the question
\S\ref{sec:removal} answers from the weight side.

Removal against suppression is unsettled here, and nobody has certified the distinction. The frontier
laboratory result hedges toward suppression for out-of-domain repair \citep{wang2025persona}; inoculated
models remain elicitable and their authors draw the line themselves against unlearning
\citep{tan2025inoculation}; the same prompting method, in settings that do not include emergent
misalignment, made $2$ of $6$ models more compliant with explicit requests for the behavior it was meant to
prevent \citep{wichers2025inoculation}; $3$ production mitigations relocate misalignment behind
contextual triggers that include semantically opposite and entirely benign prompts, while also genuinely
reducing the triggered rate \citep{dubinski2026conditional}; and one appendix of the replication attempt
finds early stopping removing the phenomenon in one model and only suppressing its expression in another
\citep{schreiber2026overtrained}. This is why we ask in \S\ref{sec:points} for a certificate, not a
rate.

\end{document}